\documentclass[lettersize,journal]{IEEEtran}
\usepackage{amsmath,amsfonts}
\usepackage{algorithmic}
\usepackage{algorithm}
\usepackage{array}
\usepackage[caption=false,font=normalsize,labelfont=sf,textfont=sf]{subfig}
\usepackage{textcomp}
\usepackage{stfloats}
\usepackage{url}
\usepackage{verbatim}
\usepackage{graphicx}
\usepackage{cite}
\usepackage{amsmath}
\usepackage{xcolor}
\usepackage{booktabs}
\usepackage{multirow}  
\usepackage{array}  
\usepackage{makecell} 

\usepackage{wrapfig}
\usepackage{enumitem}
\usepackage[justification=centering]{caption}

\def\method{DFedCata}

\begin{document}

\title{Boosting the Performance of Decentralized Federated Learning via Catalyst Acceleration}

\author{Qinglun Li, Miao Zhang, Yingqi Liu, Quanjun Yin, Li Shen, Xiaochun Cao~\IEEEmembership{Senior Member, IEEE}
        % <-this % stops a space
\thanks{\IEEEcompsocthanksitem Qinglun Li, Miao Zhang, and Quanjun Yin are with the College of Systems Engineering, National University of Defense Technology. (e-mail: \texttt{liqinglun@nudt.edu.cn}, \texttt{zhangmiao15@nudt.edu.cn}, \texttt{yin\_quanjun@163.com}) }
        
\thanks{Yingqi Liu, Li Shen, and Xiaochun Cao are with the  School of Cyber Science and Technology, Shenzhen Campus of Sun Yat-sen University, Shenzhen 518107, China.
(e-mail: \texttt{liuyingqi1199@gmail.com}, \texttt{mathshenli@gmail.com}, \texttt{caoxiaochun@mail.sysu.edu.cn})}

\thanks{Manuscript received April XX, XXXX; revised August XX, XXXX.}}

% \author {    
%       \textsuperscript{\rm 1},
%       \textsuperscript{\rm 2}\thanks{Corresponding authors.}, 
%       \textsuperscript{\rm 1}, 
%       \textsuperscript{\rm 1*}, 
%       \textsuperscript{\rm 3}\\
% }
% $^{(\textrm{\Letter})}$
% Affiliations
% \affiliations {
%     \textsuperscript{\rm 1}  National University of Defense Technology, China \\
%     \textsuperscript{\rm 2} JD Explore Academy, China \\
%     \textsuperscript{\rm 3} The University of Sydney, Australia\\
%     {\texttt{liqinglun@nudt.edu.cn; mathshenli@gmail.com; \\ 
%     lgh@nudt.edu.cn; yin\_quanjun@163.com; dacheng.tao@gmail.com}
%     }
%     }

\markboth{Journal of \LaTeX\ Class Files,~Vol.~14, No.~8, August~2021}%
{Shell \MakeLowercase{\textit{et al.}}: A Sample Article Using IEEEtran.cls for IEEE Journals}

% % column for its text to clear the IEEEpubid mark.

\maketitle

\begin{abstract}
Decentralized Federated Learning has emerged as an alternative to centralized architectures due to its faster training, privacy preservation, and reduced communication overhead. In decentralized communication, the server aggregation phase in Centralized Federated Learning shifts to the client side, which means that clients connect with each other in a peer-to-peer manner.
However, compared to the centralized mode, data heterogeneity in Decentralized Federated Learning will cause larger variances between aggregated models, which leads to slow convergence in training and poor generalization performance in tests.
To address these issues, we introduce Catalyst Acceleration and propose an acceleration Decentralized Federated Learning algorithm called \method{}. It consists of two main components: the Moreau envelope function, which primarily addresses parameter inconsistencies among clients caused by data heterogeneity, and Nesterov's extrapolation step, which accelerates the aggregation phase. 
Theoretically, We prove the optimization error bound and generalization error bound of the algorithm, providing a further understanding of the nature of the algorithm and the theoretical perspectives on the hyperparameter choice. 
Empirically, we demonstrate the advantages of the proposed algorithm in both convergence speed and generalization performance on CIFAR10/100  with various non-iid data distributions. Furthermore, we also experimentally verify the theoretical properties of DFedCata.
\end{abstract}


\newtheorem{theorem}{Theorem}
\newtheorem{proposition}{Proposition}
\newtheorem{lemma}{Lemma}
\newtheorem{corollary}{Corollary}
\newtheorem{definition}{Definition}
\newtheorem{assumption}{Assumption}
\newtheorem{remark}{Remark}
\newtheorem{proof}{Proof}

\section{Introduction}

\IEEEPARstart{F}{ederated} Learning (FL) is a new distributed machine learning paradigm that prioritizes privacy protection \cite{mcmahan2017communication,Gu_Xu_Huo_Deng_Huang_2022,Zhou_Wang_Guo_Gong_Zheng_2019}. It enables multiple clients to collaborate on training models without sharing their raw data. Nowadays, much of the research \cite{zhou2023federated,sun2023fedspeed,Li_Sahu_Zaheer_Sanjabi_Talwalkar_Smith_2018, acar2021federated,zhang2021fedpd,dai2023fedgamma} focus on Centralized Federated Learning (CFL), but the central server in CFL brings various challenges on communication burden, single point of failure \cite{chen2023enhancing}, privacy breaches \cite{gabrielli2023survey} and so on. In contrast, Decentralized Federated Learning (DFL) centralizes both the local update and aggregation steps on the client, which offers enhanced privacy protection \cite{cyffers2022privacy}, faster model training \cite{lian2017can}, and robustness to slow client devices \cite{neglia2019role}. Therefore, DFL has become a popular alternative solution \cite{chen2023enhancing, lian2017can}.

It is well recognized that a crucial difference between FL and distributed learning is the non-IID nature of the client data in FL \cite{scaffold2020, shi2023improving}. This seriously hinders the convergence speed and generalization performance for FL algorithms, posing obstacles to the practical application of FL systems. On the other hand, in comparison to CFL, the lack of central server coordination in DFL leads to increasing inconsistency between the model parameters $\mathbf{x}_i$ of client $i$ and $\mathbf{x}_j$ of client $j$ as the communication rounds increase \cite{shi2023improving}. This severe inconsistency results in a significant gap between the final output of the model $\bar{\mathbf{x}}$ and the global optimum $\mathbf{x}^*$, leading to performance disparities compared to CFL. 
Therefore, it is natural to ask, could we find an algorithm to control the inconsistency between clients while achieving better convergence and generalization performance for DFL systems?

% \begin{wrapfigure}{r}{0.5\linewidth}  
% \vspace{-0.5cm}  
%     \centering  
%     \includegraphics[width=0.5\textwidth]{figure/DFedCata.png}  
%     \vspace{-1em} 
%     \caption{\small The optimization process diagrams for two clients under the DFedAvg and \method{} algorithms are simulated. The primary improvements include two aspects. Firstly, the Moreau envelope reduces the inconsistency between clients (\method{} has a shorter length of black dashed line). Secondly, the Nesterov acceleration during the aggregation stage significantly brings $\bar{\mathbf{x}}$ closer to the optimal value $\mathbf{x}^*$.}  
%     \label{fig:DFedCata}  
%     \vspace{-1em}  
% \end{wrapfigure} 

\begin{figure}[t]
    \vspace{-0.5cm}
    \centering
    \includegraphics[width=0.5\textwidth]{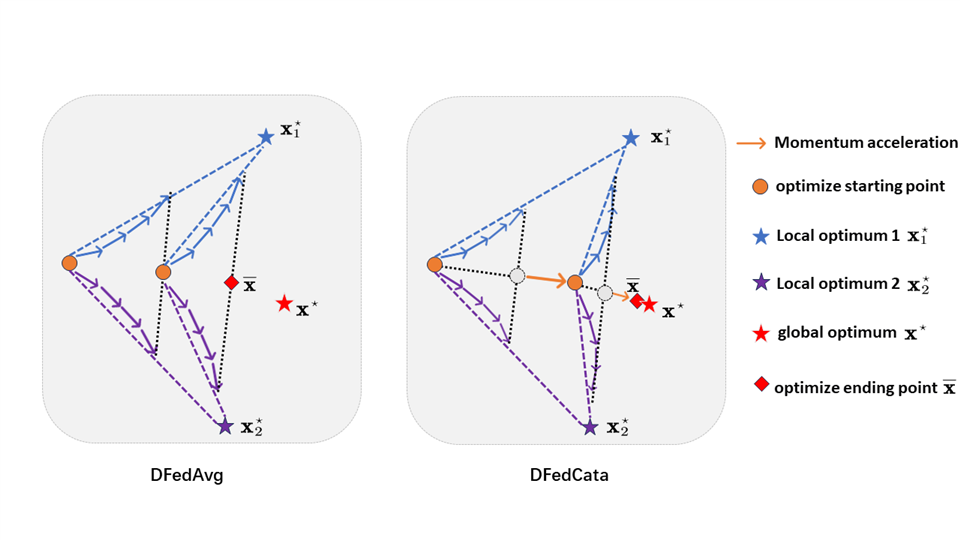}
    \vspace{-0.3cm}
    \caption{\small The optimization process diagrams for two clients under the DFedAvg and \method{} algorithms are simulated. The primary improvements include two aspects. Firstly, the Moreau envelope reduces the inconsistency between clients (\method{} has a shorter length of black dashed line). Secondly, the Nesterov acceleration during the aggregation stage significantly brings $\bar{\mathbf{x}}$ closer to the optimal value $\mathbf{x}^*$.}
    \vspace{-1em}
    \label{fig:DFedCata}
\end{figure}

In this paper, we introduce the Catalyst  \cite{lin2018catalyst} to the non-convex DFL optimization. Originally designed as a single optimizer, Catalyst cannot be applied to solve distributed optimization problems. Its primary function lies in accelerating gradient-based convex optimization methods in the sense of Nesterov. Catalyst mainly consists of the Moreau envelope \cite{moreau1962fonctions,yosida1980functional} and the Nesterov acceleration step \cite{nesterov1983method}. We demonstrate the former component enhances the inconsistency among clients, similar to the original intention of the FedProx \cite{fedprox} algorithm. The latter component is primarily used to accelerate algorithm convergence. Notably, the application of the acceleration step differs from the incorporation of momentum in the local update phase of DFedAvgM \cite{Sun2022Decentralized}. Our acceleration step is performed during the client aggregation phase, where the difference between pre-and post-aggregation, $\mathbf{x}_i^{t}-\mathbf{x}_i^{t-1}$, can approximately represent the direction of the global optimal value, as illustrated in Figure \ref{fig:DFedCata}. This significantly accelerates the training process.

In theoretical analysis, we establish both the convergence error and the generalization error bound of \method{}, providing a more rigorous understanding of the nature of the algorithm. This addresses the deficiency of the understanding of generalization analysis in prior studies \cite{Sun2022Decentralized, shi2023improving, lian2017can}. Our generalization error bound is established based on stability analysis \cite{Hardt2016train,sun2021stability,Sun2022Stability}, which differs from the generalization based on PAC \cite{deng2020adaptive,marfoq2022personalized,xie2024perada}, differential privacy \cite{dai2022dispfl}. Our generalization error bounds can provide more refined conclusions for iteration points $\mathbf{x}^T$, algorithm hyperparameters (such as $T,\eta,K ,\beta$), etc. Furthermore, based on the established convergence and generalization error bound, we provide the perspective on how the hyperparameter $\beta$ in \method{} accelerates algorithm convergence and what type of decentralized communication topology can achieve better generalization (more conclusions see Remark \ref{remark:optimization}\&\ref{Remark:gener}).  Empirically, we conduct numerous non-iid experiments (including Dirichlet and Pathological distribution) on CIFAR10/100 to verify the faster convergence and better generalization of \method{}. The results indicate that \method{} improves the generalization performance by at least $3\%$ (see Table \ref{ta:all_baselines}) and achieves the best convergence speed of $8.6\times$ (see Table \ref{ta:CIFAR10-convergence}) compared to the other SOTA methods. Both the theoretical analysis and experiments prove the efficiency of the proposed method. Furthermore, we also experimentally verify the theoretical properties of DFedCata.
In summary, our main contributions are three-fold as follows:
\begin{itemize}[leftmargin=*,noitemsep,topsep=0pt,partopsep=0pt]
    \item In response to the slow convergence and poor generalization of existing DFL optimizers, we propose \method{}, which extends the Catalyst technique \cite{lin2018catalyst} from its original application to convex optimization problems to non-convex distributed optimization problems. One component of Catalyst, the Moreau envelope, is employed to enhance the consistency between clients during the local update stage, while another component, Nesterov acceleration, is used to expedite convergence during the aggregation stage. Figure \ref{fig:DFedCata} succinctly illustrates the acceleration effect of \method{}.
    \item In theoretical analysis, we jointly analyze the convergence error bound and generalization error bound of \method{} (see Section \ref{sec:Theoretical Analysis}), addressing the deficiency in prior DFL research regarding the understanding of generalization. Furthermore, our theoretical analysis provides insights into questions such as how the hyperparameter $\beta$ accelerates the algorithm and what conditions a topology should satisfy to achieve the optimal generalization performance for \method{} (see Remark \ref{remark:optimization} \& \ref{Remark:gener}). The experiments validate our theoretical findings (see Section \ref{sec:experiment}).
    \item In our experiments, we simulate various non-iid scenarios on CIFAR10\&100 to test the generalization performance and convergence speed of \method{} (see Section \ref{sec:experiment}). The results compared with prominent baseline in both DFL and CFL demonstrate that \method{} achieves the SOTA performance in terms of generalization (see Table \ref{ta:all_baselines}) and convergence (see Table \ref{ta:CIFAR10-convergence}). Notably, \method{} outperforms the DFL methods under all common communication topologies, especially the sparse topological connections.
\end{itemize}

\section{Related Work}

Below, we provide a brief review of the most relevant literature to our research, including DFL, optimization acceleration techniques, and convergence and generalization for DFL. 

\textbf{Decentralized Federated Learning.} To alleviate the communication burden on the server in CFL and release the computational capabilities of edge devices, DFL shifts the centralized server aggregation to the clients. It means that the communication load is distributed to each node while maintaining a comparable overall communication complexity with the centralized scenario \cite{lian2017can}. Additionally, decentralized communication methods provide enhanced privacy protection compared to CFL \cite{yang2019federated,lalitha2018fully, lalitha2019peer}. DFL has emerged as a promising area of research and has been recognized as a challenge in recent years \cite{beltran2022decentralized, Kairouz2021Advances}. Sun et al. \cite{Sun2022Decentralized} extend the FedAvg algorithm \cite{mcmahan2017communication} to decentralized scenarios and complement it with local momentum acceleration to improve convergence. Furthermore, Dai et al. \cite{dai2022dispfl} introduce sparse training into DFL to reduce communication and computation costs, while Shi et al. \cite{shi2023improving} apply SAM to DFL and enhance the consistency among clients by incorporating Multiple Gossip Steps. For further related work on DFL, please refer to the survey papers \cite{gabrielli2023survey,yuan2023decentralized,beltran2022decentralized} and their references therein.
These efforts gradually enhance the performance of DFL from various perspectives. However, these works focus solely on either convergence speed or generalization performance, not both. Thus, designing an algorithm that achieves both fast convergence and good generalization remains a challenge.

\textbf{Acceleration Techniques.} 
Acceleration techniques in Deep Learning can be divided into momentum and restart. Momentum techniques primarily optimize the design of the optimizer, while restart techniques focus on the selection of initialization points. Momentum has been successfully applied in both the CFL and DFL domains. Global momentum methods, such as FedCM \cite{xu2021fedcm} and MimeLite \cite{karimireddy2020mime}, estimate the global momentum at the server and apply it to each client update in CFL, addressing client heterogeneity. In DFL, local momentum, like the DFedAvgM algorithm proposed by Sun et al. \cite{Sun2022Decentralized}, is utilized to accelerate the convergence process. Restart techniques, such as the Lookahead optimizer introduced by Zhou \cite{Zhou2021towards}, employ a $k$-steps forward, $1$ step back approach to improve learning stability and reduce variance in the inner optimizer, with nearly negligible computational cost. Additionally, Lin et al. \cite{lin2015universal} develop Catalyst, a universal framework that accelerates first-order optimization methods including SAG \cite{schmidt2017minimizing}, SAGA \cite{defazio2014saga}, and SVRG \cite{xiao2014proximal}. Moreover, Trimbach et al. \cite{trimbach2021acceleration} extend Catalyst to distributed learning and use it to accelerate the DSGD algorithm \cite{koloskova2020unified}. Besides the aforementioned acceleration techniques in deep learning, numerical optimization methods also include Chebyshev acceleration \cite{li2020fast} and Anderson acceleration \cite{walker2011anderson}.

\textbf{Convergence and Generalization for DFL.} 
The current literature in the field of DFL, including works such as DFedAvg \cite{Sun2022Decentralized}, DFedAvgM \cite{Sun2022Decentralized}, and DFedSAM \cite{shi2023improving}, has mainly focused on the analysis of convergence, while lacking in-depth exploration of algorithm generalization. Although Dis-PFL \cite{dai2022dispfl} provides a relatively coarse generalization bound for the algorithm based on differential privacy, it can not demonstrate the generalization performance on each iterative point. In distributed learning, Sun et al. \cite{sun2021stability} establish a generalization bound for Decentralized SGD based on stability, Zhu et al. \cite{Zhu2023Stability} derived the generalization properties of SGDA based on stability, and Deng et al. \cite{deng2023stability,deng2023towards} similarly presented generalization bounds for Asynchronous Decentralized SGD and Delayed SGD based on stability. It is worth noting that the generalization bounds based on stability in the works above are all derived from refined upper bounds based on iterative points. In contrast, the DFL domain still lacks an analysis of generalization bounds for the algorithm's iterative points.

% \ls{At last, we should comment that applying the acceleration techniques to DFL is nontrivial. We should highlight the challenges as the rebuttal file of OledFL.} 老师，我没明白啥意思
\section{Methodology}

% \begin{algorithm}[ht]
% \small
% \renewcommand{\algorithmicrequire}{\textbf{Input:}}
% \renewcommand{\algorithmicensure}{\textbf{Output:}}
% \caption{ \method{} Algorithm}
%     \begin{algorithmic}[1]\label{alg:DFedCata}
%         \REQUIRE Initialize $\eta,\beta,\lambda > 0$  and $\mathbf{x}_{i}^{0}\!\in\! \mathbb{R}^d$ for all nodes.\\
%         \ENSURE model average parameters $\bar{\mathbf{x}}^{t}$.
%         \FOR{$t = 0, 1, 2, \cdots, T-1$}
%         \FOR{client $i$ in parallel}
%         \STATE set $\mathbf{x}_{i,0}^{t}=\mathbf{x}_i^{t} + \beta (\mathbf{x}_i^t - \mathbf{x}_{i}^{t-1})$
%         \FOR{$k = 0, 1, 2, \cdots, K-1$}
%         \STATE sample a minibatch $\varepsilon_{i,k}^{t}$ and do
%         \STATE estimate stochastic gradient: $\mathbf{g}_{i,k}^{t}=\nabla f_{i}(\mathbf{x}_{i,k}^{t};\varepsilon_{i,k}^{t})$
%         \STATE perform SGD step: $\mathbf{x}_{i,k+1}^{t}=\mathbf{x}_{i,k}^{t}-\eta(\mathbf{g}_{i,k}^t +\lambda(\mathbf{x}_{i,k}^{t} - \mathbf{x}_{i,0}^{t}))$
%         \ENDFOR
%         \STATE Mix the received model $\mathbf{x}_{j,K}^{t}$ with mixing matrix ${\bf W}$: $\mathbf{x}_i^{t+1} = \sum_j w_{i,j} \mathbf{x}_{j,K}^{t}$ 
%         \ENDFOR
%         \ENDFOR
%     \end{algorithmic}
% \end{algorithm}

In this section, we first demonstrate the meanings of several notations and the DFL problem setup. Subsequently, we elaborate on the two components of \method{}, namely the Moreau envelope and Nesterov acceleration, then present the complete algorithm procedure.

\subsection{Notations and Problem Setup}

We conduct $T$ communication rounds for $m$ clients in the DFL system.  $(\cdot)_{i,k}^{t}$ indicates variable $(\cdot)$ at the $k$-th iteration of the $t$-th round in the $i$-th client. $\mathbf{x}$ denotes the model parameters. 
The communication topology between clients can be represented as graph denoted as $\mathcal{G} = (\mathcal{N}, \mathcal{E}, {\bf W})$, where $\mathcal{N} = \{1, 2, \ldots, m\}$ represents the set of clients, $\mathcal{E} \subseteq \mathcal{N} \times \mathcal{N}$ denotes the links between clients,
$w_{i,j}$ represents the weight of the link between client $j$ and client $i$ and ${ \bf W} = [w_{i,j}] \in [0,1]^{m\times m}$ represents the mixing matrix. The inner product of two vectors is denoted by $\langle\cdot,\cdot\rangle$, and $\Vert \cdot \Vert$ represents the Euclidean norm of a vector. 
% Other symbols will be explained in their respective contexts.

% \subsection{Problem Setup}
We denote $\mathcal{D}_i$ as the data distribution in the $i$-th client, the difference between which demonstrates the data heterogeneity across clients. Each client's local objective function $F_i({\mathbf{x}};\xi)$ is associated with the training data samples $\xi$.

The objective of DFL is to jointly solve the following distributed population risk \(F\) minimization:
\begin{equation}\label{finite_sum}
    \small \min_{{\mathbf{x}}\in \mathbb{R}^d} F({\mathbf{x}}):=\frac{1}{m}\sum_{i=1}^m F_i({\mathbf{x}}),~~F_i({\mathbf{x}})=\mathbb{E}_{\xi\sim \mathcal{D}_i} F_i({\mathbf{x}};\xi)
\end{equation}
% where $\mathcal{D}_i$ represents the data distribution in the $i$-th client, which exhibits heterogeneity across clients. Each client's local objective function $F_i({\mathbf{x}};\xi)$ is associated with the training data samples $\xi$. 
Let $\mathbf{x}_{\mathcal{D}}^\star = \arg \min_{\mathbf{x}}F(\mathbf{x})$ as the optimal value of (\ref{finite_sum}). Unlike CFL, we address (\ref{finite_sum}) by enabling clients to collaborate through a mixing matrix $\mathbf{W}$ in a decentralized manner, leveraging peer-to-peer communication among clients without the need for server coordination. 
Practically, we consider the empirical risk minimization of the non-convex finite-sum problem in DFL:
\begin{equation}\label{finite_sum_empirical}
    \small \min_{{\mathbf{x}}\in \mathbb{R}^d} f({\mathbf{x}}):=\frac{1}{m}\sum_{i=1}^m f_i({\mathbf{x}}),~~f_i({\mathbf{x}})=\frac{1}{S_i}\sum_{z_j\in\mathcal{S}_i}f_i({\mathbf{x}};z_j)
\end{equation}
where each client stores a private dataset \(S_i = \{z_j\}\), with \(z_j\) drawn from an unknown distribution \(\mathcal{D}_i\). And we denote $\mathbf{x}^\star = \arg \min_{\mathbf{x}}f(\mathbf{x})$ as the optimal value of problem (\ref{finite_sum_empirical}).

\subsection{DFedCata Algorithm}

\begin{algorithm}[H]
    \small
    \renewcommand{\algorithmicrequire}{\textbf{Input:}}
    \renewcommand{\algorithmicensure}{\textbf{Output:}}
    \caption{ \method{} Algorithm}
    \begin{algorithmic}[1]\label{alg:DFedCata}
    \REQUIRE Initialize $\eta,\beta,\lambda > 0$  and $\mathbf{x}_{i}^{0}\!\in\! \mathbb{R}^d$ for all nodes.\\
    \ENSURE model average parameters $\bar{\mathbf{x}}^{t}$.
    \FOR{$t = 0, 1, 2, \cdots, T-1$}
    \FOR{client $i$ in parallel}
    \STATE set $\mathbf{x}_{i,0}^{t}=\mathbf{x}_i^{t} + \beta (\mathbf{x}_i^t - \mathbf{x}_{i}^{t-1})$
    \FOR{$k = 0, 1, 2, \cdots, K-1$}
    \STATE sample a minibatch $\varepsilon_{i,k}^{t}$ and do
    \STATE estimate stochastic gradient: $\mathbf{g}_{i,k}^{t}=\nabla f_{i}(\mathbf{x}_{i,k}^{t};\varepsilon_{i,k}^{t})$
    \STATE perform SGD step: $\mathbf{x}_{i,k+1}^{t}=\mathbf{x}_{i,k}^{t}-\eta(\mathbf{g}_{i,k}^t +\lambda(\mathbf{x}_{i,k}^{t} - \mathbf{x}_{i,0}^{t}))$
    \ENDFOR
    \STATE Mix the received model $\mathbf{x}_{j,K}^{t}$ with mixing matrix ${\bf W}$: $\mathbf{x}_i^{t+1} = \sum_j w_{i,j} \mathbf{x}_{j,K}^{t}$ 
    \ENDFOR
    \ENDFOR
    \end{algorithmic}
\end{algorithm}

\textbf{Using the Moreau Envelope (ME) during local updates.}
ME looks similar to the proximal point method since they both use the Moreau envelope function. In CFL, the FedProx \cite{fedprox} is an application of the Moreau envelope. Specifically, the Moreau envelope function $h_i(\mathbf{x},\mathbf{x}_{i,0}^t)$ is added to the loss function $f_i(\mathbf{x})$ on each client, introducing a penalty term $\frac{\lambda}{2}\|\mathbf{x} - \mathbf{x}_{i,0}^t\|^2$. Thus, during the local update stage, each client solves the following problem through $K$ iterations:
\begin{align*}
    \mathbf{x}_{i,K}^t &\approx \arg \min_{\mathbf{x}}h_i(\mathbf{x},\mathbf{x}_{i,0}^t)\\
    &= \arg \min_{\mathbf{x}}\{ f_i(\mathbf{x}) + \frac{\lambda}{2}\|\mathbf{x} - \mathbf{x}_{i,0}^t\|^2\}
\end{align*}

\textbf{The Effectiveness of ME.} The Moreau Envelope \cite{moreau1962fonctions} is a mathematical tool used to smooth non-differentiable functions by creating a differentiable approximation. It essentially transforms a possibly non-smooth objective function into a smooth one, facilitating more effective gradient-based optimization. This technique is particularly beneficial when dealing with functions that have sharp edges or are not smooth, as it helps in avoiding the pitfalls of local minima by smoothing the objective landscape. In DFL, the purpose of the penalty term is to ensure that the client does not deviate too far from the initial value $\mathbf{x}_{i,0}^t$ during the local update, thereby enhancing consistency between clients. As the initial value is aggregated through the mixing matrix $\mathbf{W}$, it can be inferred from the properties of $\mathbf{W}$ that after aggregation, the parameter values of the client and its neighbors are approximately equal (see \cite[Lemma 4]{lian2017can}). Therefore, the inclusion of the penalty term reinforces the consistency between clients, thereby alleviating the impact of heterogeneous data.

\textbf{Using the Nesterov Acceleration (NA) during client aggregation.} Unlike the use of momentum for acceleration during the local update in DFedAvgM \cite{Sun2022Decentralized}, \method{} employs Nesterov acceleration during the client aggregation phase, as indicated in line 3 of the Algorithm \ref{alg:DFedCata}. The main function of Nesterov acceleration is illustrated in Figure \ref{fig:DFedCata} (b), where the direction $\mathbf{x}_i^{t}-\mathbf{x}_i^{t-1}$ approximately indicates the direction from the current iteration point to the global optimal value. By employing the Nesterov acceleration step, clients can skip multiple iterations, resulting in acceleration. Furthermore, when combined with the assistance of the ME function during the client update phase, this ensures that all clients' updates progress toward the global stable point. Ablation experiments have demonstrated the effectiveness of this combination (see Section \ref{sec:ablation_section}).

\section{Theoretical Analysis}\label{sec:Theoretical Analysis}

In this section, we first introduce the necessary assumptions utilized in our proofs. Next, we outline the main theorems, encompassing both optimization and generalization. Based on these, we provide significant remarks on the nature of DFedCata and perspectives on the better hyperparameter choice.

\subsection{Definition and Assumptions}

Below, we introduce the definition of the mixing matrix and several assumptions in our analysis. 

\begin{definition}\label{def:mixing matrix}
    (Gossip/Mixing matrix). [Definition 1, \cite{Sun2022Decentralized}] The gossip matrix ${ \bf W} = [w_{i,j}] \in [0,1]^{m\times m}$  is assumed to have the following properties:
    (i) \textbf{(Graph)} If $i\neq j$ and $(i,j) \notin {\cal V}$, then $w_{i,j} =0$, otherwise, $w_{i,j} >0$;
    (ii) \textbf{(Symmetry)} ${\bf W} = {\bf W}^{\top}$;
    (iii) \textbf{(Null space property)} $\mathrm{null} \{{\bf I}-{\bf W}\} = \mathrm{span}\{\bf 1\}$;
    (iv) \textbf{(Spectral property)} ${\bf I} \succeq {\bf W} \succ -{\bf I}$. 
    Under these properties, the eigenvalues of $\bf{\bf W}$ satisfy $1=|\psi_1({\bf W)})|> |\psi_2({\bf W)})| \ge \dots \ge |\psi_m({\bf W)})|$. Furthermore, we define $\psi:=\max\{|\psi_2({\bf W)}|,|\psi_m({\bf W)})|\}$ and $1-\psi \in (0,1]$ as the spectral gap of $\bf W$.
\end{definition}

\begin{assumption}\label{as:smoothness}
    (\textbf{L-Smoothness}) \textit{The non-convex function $f_{i}$ satisfies the smoothness property for all $i\in[m]$, i.e., $\Vert\nabla f_{i}(\mathbf{x})-\nabla f_{i}(\mathbf{y})\Vert\leq L\Vert\mathbf{x}-\mathbf{y}\Vert$, for all $\mathbf{x},\mathbf{y}\in\mathbb{R}^{d}$.}
\end{assumption}

\begin{assumption}\label{as:bounded_stochastic_gradient}
(\textbf{Bounded Variance}) The stochastic gradient $\mathbf{g}_{i,k}^{t}=\nabla f_{i}(\mathbf{x}_{i,k}^{t}, \varepsilon_{i,k}^{t})$ with the randomly sampled data $\varepsilon_{i,k}^{t}$ on the local client $i$ is unbiased and with bounded variance, i.e., $\mathbb{E}[\mathbf{g}_{i,k}^{t}]=\nabla f_{i}(\mathbf{x}_{i,k}^{t})$ and $\mathbb{E}\Vert \mathbf{g}_{i,k}^{t} - \nabla f_{i}(\mathbf{x}_{i,k}^{t})\Vert^{2} \leq \sigma^{2}$, for all $\mathbf{x}_{i,k}^{t}\in\mathbb{R}^{d}$.
 \end{assumption}

\begin{assumption}\label{as:bounded_heterogeneity}
(\textbf{Bounded Heterogeneity}) 
For all $\mathbf{x}\in\mathbb{R}^{d}$.the heterogeneous similarity is bounded on the gradient norm as $\mathbb{E}\|\nabla f_i(\mathbf{x})\|^2\leq G^2+B^2\mathbb{E}\|\nabla f(\mathbf{x})\|^2$, where $G \geq 0$ and $B \geq 1$ are two  constants.
\end{assumption}

\begin{assumption}\label{as:L_G-lip}
    (\textbf{Lipschitz Continuity}). The global function fsatisfies the $L_G$-Lipschitz property, i.e. for $\forall \mathbf{x}_1, \mathbf{x}_2$, $\|f(\mathbf{x}_1) - f(\mathbf{x}_2)\| \leq L_G\| \mathbf{x}_1 - \mathbf{x}_2 \|.$
\end{assumption}

Definition \ref{def:mixing matrix} is commonly utilized to describe the communication topology in DFL and can also be interpreted as a Markov transition matrix. The term \(1-\psi\) quantifies the speed of convergence to the equilibrium state. Assumptions \ref{as:smoothness}-\ref{as:bounded_heterogeneity} are mild and are commonly employed for analyzing the non-convex objectives in DFL \cite{Sun2022Decentralized, shi2023improving}. Assumption \ref{as:L_G-lip} is employed to bound the uniform stability for the non-convex objective \cite{Sun2022Stability, sun2023understanding, Hardt2016train}. It is noteworthy that when discussing the algorithm's convergence speed, only assumptions \ref{as:smoothness}-\ref{as:bounded_heterogeneity} are utilized, while in describing the algorithm's generalization bound, we rely on assumptions \ref{as:smoothness}, \ref{as:bounded_stochastic_gradient}, and \ref{as:L_G-lip}. This is due to the fact that assumption \ref{as:L_G-lip} implies bounded gradients, i.e., \(\|\nabla f_i(\mathbf{x})\| \leq L_G\), while assumption \ref{as:bounded_heterogeneity} is more general than the bounded gradient assumption.

\subsection{Optimization Analysis}
In this part, we first elucidate the primary challenges in algorithm analysis, and then present the optimization error and convergence rate of the proposed \method{} under assumptions \ref{as:smoothness}-\ref{as:bounded_heterogeneity}. The detailed proofs can be found in Appendix \ref{appendix:optimization error}.

\textbf{Primary Challenges.} The challenges in algorithm analysis stem from the Nesterov acceleration step in the aggregation stage, given by $\mathbf{x}_{i,0}^{t}=\mathbf{x}_i^{t} + \beta (\mathbf{x}_i^t - \mathbf{x}_{i}^{t-1})$. On one hand, this results in the appearance of a second-order finite difference equation problem in the proof process for the sequence $\{\bar{\mathbf{x}}^t\}$, which is not easy to solve. On the other hand, it makes it difficult to bound the upper limit of the inner product term $\langle\nabla f(\bar{\mathbf{x}}^t),\bar{\mathbf{x}}^{t+1}-\bar{\mathbf{x}}^t\rangle$. Similar to the approach taken by \cite{sun2023fedspeed}, our strategy is to construct an auxiliary sequence $\mathbf{z}^t = \bar{\mathbf{x}}^t + \frac{\beta}{1-\beta}\left(\bar{\mathbf{x}}^t - \bar{\mathbf{x}}^{t-1}\right)$, where $\bar{\mathbf{x}}^{t} = \frac{1}{m}\sum_{i=1}^m\mathbf{x}_i^{t}$. The update form of the sequence $\mathbf{z}^t$ is similar to the vanilla SGD form (see Lemma \ref{le:auxiliary updates}), which facilitates analysis and proof. Therefore, in the subsequent proof, we analyze the convergence properties of $\mathbf{z}^t$.

\begin{theorem}\label{the:opt_error}
Under Assumption \ref{as:smoothness} - \ref{as:bounded_heterogeneity}, let $\eta < \min\{\frac{(1-\beta)^{\frac{3}{2}}}{2KL\sqrt{2L}}, \frac{(1-\beta)^2}{KL}, \frac{\sqrt{1-\beta}}{2KL\sqrt{(1+B^2)}}\}$,$\widetilde{\eta} = \frac{\gamma}{\lambda(1-\beta)}$, Where $\gamma = 1-(1-\eta\lambda)^K $, it is obvious that $\gamma \leq K\eta\lambda$ always holds. After training $T$ rounds, the auxiliary sequence generated by our proposed algorithm satisfies: 
\begin{equation*}
    \begin{aligned}
        &\frac{1}{T}\sum_{t=0}^{T-1}\mathbb{E}\|\nabla f(\mathbf{z}^{t})\|^2 
        \leq \frac{2\mathbb{E}[f(\mathbf{z}^{0})-f^*]}{T\tilde{\eta}\kappa} + \frac{4\tilde{\eta}^2L^2(1-\beta)}{\mu\kappa}G^2   \\
        & + \frac{2}{T\kappa}\left( \frac{ L^2 \beta}{(1-\beta)\mu} \!\!+ \!\! \frac{ L^2 \beta^2}{ (1-\beta)^3} \!\! + \!\!\frac{4 L^3 \beta^2}{ (1-\beta)^2 \mu}\tilde{\eta}^2 \right)\mathbb{E}\|\frac{1}{m}\sum_{i=1}^m\mathbf{x}_i^0\|^2\\
        & +  \frac{2\tilde{\eta}}{\kappa}\left(\frac{ L^2 m (1-\beta)}{\mu} + L\right)\frac{\sigma^2}{m}
    \end{aligned}
\end{equation*}
where $\kappa > \frac12$ is a constant and   
$\mu = 1 - \frac{4L^2}{1-\beta}\frac{\gamma^2}{\lambda^2}$. Under the aforementioned learning rate constraint, we can obtain $\mu \geq \frac{B^2}{1+B^2}$. 
Further, by selecting the proper learning rate $\eta = \mathcal{O}(\frac{1-\beta}{\sqrt{KT}})$, then $\widetilde{\eta} = \mathcal{O}\left({\sqrt{\frac{K}{T}}}\right)$ and let $D = f(\bar{\mathbf{z}}^0) - f^*$ as the initialization bias, $\widetilde{D} = \mathbb{E}\|\frac{1}{m}\sum_{i=1}^m\mathbf{x}_i^0\|^2$ as the square of the Euclidean norm of the initial parameter sum. The auxiliary sequence $\mathbf{z}^t$ satisfies:
\begin{equation*}
\begin{aligned}
    &\frac{1}{T}\sum_{t=0}^{T-1}\mathbb{E}\| \nabla f(\mathbf{z}^t) \|^2 \!=\! \mathcal{O}\Big(\frac{D}{\sqrt{KT}} \!+\! \frac{\sqrt{K}(L^2(1-\beta) + Lm)}{\sqrt{T}}\sigma^2 \\
    & + \frac{KL^2(1-\beta)}{T}G^2  + \frac{L^2\beta^2}{T(1-\beta)^3}\widetilde{D}\Big)
\end{aligned}
\end{equation*}
\end{theorem}

\begin{remark}\label{remark:optimization}
    Overall, we have established the convergence rate of \method{} to be $\mathcal{O}\left({\sqrt{\frac{1}{T}}}\right)$. Next, we will utilize this optimization error bound to discuss some properties of \method{}. The dominant term of the optimization error bound is $\frac{D}{\sqrt{KT}} + \frac{\sqrt{K}(L^2(1-\beta) + Lm)}{\sqrt{T}}\sigma^2$. Firstly, an appropriate value of $K$ can tighten the optimization error bound, implying that a suitable number of local epochs can accelerate convergence, which is verified in local epochs $K$ of Section \ref{sec:ablation_section}. Secondly, an excessively large value of $m$ will slow down convergence, as the complexity of the optimization problem increases due to the non-iid data held by a large number of clients, as also validated in Section \ref{sec:ablation_section}. Thirdly, increasing the Catalyst parameter $\beta$ will accelerate convergence by tightening the said bound, which theoretically proves the acceleration property of \method{}, as demonstrated in Section \ref{sec:convergence_speed}.
    Furthermore, we find that the impact of $G^2$ in Assumption \ref{as:bounded_heterogeneity} on the convergence bound is $\mathcal{O}\left(\frac{1}{T}\right)$, which can be negligibly small. Regarding the issue of $\beta$ selection affecting $\widetilde{D}$, we can address this by setting the initial parameter $\mathbf{x}_i^0 = 0$.
    % Furthermore, we can observe certain combinatorial properties, such as increasing $\beta$ leading to a reduction in $K$. This can be utilized to decrease local computations using the acceleration property of the algorithm while maintaining the overall acceleration effect. This elucidates our rationale for setting $K=2$ and $\beta=0.99$ in the CIFAR100 dataset in Section \ref{sec:Experiment_Setup}. \ls{Be careful about the last point.}
\end{remark}

\subsection{Generalization Analysis}
In this part, our focus is to present a generalization analysis of the \method{} method based on two gradient properties (bounded variance and Lipschitz continuity). Below, we first highlight the primary challenges during the generalization analysis. Full proofs are available in the Appendix \ref{sec:Proofs for the generalization Error}.

%In this part, our focus is primarily on presenting a generalization analysis of the \method{} method based on two gradient properties (bounded variance and Lipschitz continuity). We will use uniform stability analysis, a widely adopted approach in the existing literature, to establish the generalization error of \method{}. Next, we introduce the definition of uniform stability, and the relationship between uniform stability and generalization bounds, followed by an examination of the primary challenges and solutions related to generalization analysis. Finally, we present the generalization bound. Full proofs are available in the Appendix \ref{sec:Proofs for the generalization Error}.

\textbf{Primary Challenges.} The main challenge in the uniform stability analysis of the algorithm is consistent with the main challenge in the convergence analysis, where the utilization of the Nesterov acceleration step leads to the appearance of second-order difference equations, making it difficult to bound. On the other hand, the use of the Moreau envelope in the analysis is related to the aggregated initial values, which increases the analysis difficulty. To address these challenges, we define a virtual sequence $\{\mathbf{z}_{i,k}^t\}$, where its local updates satisfy $\mathbf{z}_{i,k+1}^t = \mathbf{z}_{i,k}^t - \frac{\gamma}{\lambda(1-\beta)}\frac{\gamma_k}{\gamma}\mathbf{g}_{i,k}^t$. The initial value is the normal mixing aggregation, $\mathbf{z}_{i,0}^t = \mathbf{z}_{i}^t = \sum_{j=1}^mw_{i,j}\mathbf{z}_{j,K}^{t-1}$, and we define $\mathbf{z}^t = \frac{1}{m}\sum_{i=1}^m\mathbf{z}_{i}^t$. Then, the virtual sequence is equivalent to the sequence $\{\mathbf{x}_{i,k}^t \}$ generated by \method{} in terms of the mean value (Lemma \ref{le:equivalent}). We will analyze the virtual sequence to establish the generalization bound.

In the DFL framework, we suppose there are $m$ clients participating in the training process as a set $\mathcal{C}=\{i\}_{i=1}^m$. Each client has a local dataset $\mathcal{S}_i=\{z_j\}_{j=1}^S$ with total $S$ data sampled from a specific unknown distribution $\mathcal{D}_i$. Now we define a re-sampled dataset $\widetilde{\mathcal{S}_i}$ which only differs from the dataset $\mathcal{S}_i$ on the $j^*$-th data. We replace the $\mathcal{S}_{i^*}$ with $\widetilde{\mathcal{S}}_{i^*}$ and keep other $m-1$ local dataset, which composes a new set $\widetilde{\mathcal{C}}$. $\mathcal{C}$ only differs from the $\widetilde{\mathcal{C}}$ at $j^*$-th data on the $i^*$-th client. Then, based on these two sets, \method{} could generate two solutions, $\mathbf{z}^t$ and $\widetilde{\mathbf{z}}^t$ respectively, after $t$ rounds. By bounding the difference according to these two models, we can obtain stability and generalization bound.

% \begin{definition}
%     (Uniform Stability \cite{Hardt2016train}) For these two models $\mathbf{z}^T$ and $\widetilde{\mathbf{z}}^T$ generated as introduced above, a general method satisfies $\epsilon$-uniformly stability if:
%     \begin{equation}
%         \operatorname*{sup}_{z_{j}\sim\{\mathcal{D}_{i}\}}\mathbb{E}[f(\mathbf{z}^T;z_{j})-f(\widetilde{\mathbf{z}}^T;z_{j})]\leq\epsilon.
%     \end{equation}
%     Moreover, if a method satisfies $\epsilon$-uniformly stability, its generalization error is bounded \cite{Zhang2021Stability, Hardt2016train} $$\mathcal{E}_G \leq \operatorname*{sup}_{z_{j}\sim\{\mathcal{D}_{i}\}}\mathbb{E}[f(\mathbf{z}^T;z_{j})-f(\widetilde{\mathbf{z}}^T;z_{j})]\leq\epsilon.$$
% \end{definition}

\begin{theorem}\label{the:generation_err}
    Under Assumption \ref{as:bounded_stochastic_gradient}, \ref{as:bounded_heterogeneity}, and \ref{as:L_G-lip}, let all conditions in the optimization process be satisfied. Define $\widetilde{\mu} = \mu(1-\beta)$ and let the learning rate $ \eta =\frac{\mu(1-\beta)}{tK+k} = \frac{\widetilde{\mu}}{tK+k}$ is decayed as the communication round $t$ and iteration $k$ where ${\mu} \leq \frac{1}{L}$ is a specific constant, let $\tau_0$ be a specific round to firstly select the different data sample, and let $U = \sup_{\mathbf{x},z}\{ f ({\mathbf{x}}; z)\}$ be the upper bound, for arbitrary data sample $z$ followed the joint distribution $\mathcal{D}_i$, we have:
\begin{align*}
    &\mathbb{E}\|f({\mathbf{z}}^{T+1};z)-f(\widetilde{\mathbf{z}}^{T+1};z)\| \\
    &\leq L_G \left(\frac{TK}{\tau_0}\right)^{\mu L}\frac{2\left(1+6\sqrt{m}\kappa_\psi\right)\sigma}{SL} + \frac{U\tau_0}{S}
\end{align*}
   where $\kappa_\psi\approx\mathcal{O}\left(\frac1{\psi\ln\frac1\psi}\right)$ is a constant of spectrum gaps. In order to minimize the error, by selecting a proper $ \tau_{0}=\left(\frac{2\sigma L_G}{UL}\frac{1+6\sqrt{m}\kappa_{\psi}}{m}\right)^{\frac{1}{1+\mu L}}(TK)^{\frac{\mu L}{1+\mu L}} $ and substituting the definition of $\mu = \frac{\widetilde{\mu}}{1-\beta}$  yields
\begin{align}\label{eq:gener_bound}
    &\mathbb{E}\|f({\mathbf{z}}^{T+1};z)-f(\widetilde{\mathbf{z}}^{T+1};z)\| \\
    &\leq  \frac{2U}{S}\left(\frac{2\sigma L_G}{UL}\frac{1+6\sqrt{m}\kappa_{\psi}}{m}\right)^{\frac{1-\beta}{1 - \beta +\widetilde{\mu} L}}(TK)^{\frac{\widetilde{\mu} L}{1 - \beta +\widetilde{\mu} L}}
\end{align}
\end{theorem}

\begin{remark}\label{Remark:gener}
    Building on the established bound, we discuss some properties of the algorithm. First, regarding the catalyst parameter $\beta$, the variation of $\beta$ affects one term in (\ref{eq:gener_bound}) positively and the other negatively due to the equation $\frac{1-\beta}{1 - \beta  +\widetilde{\mu} L} + \frac{\widetilde{\mu} L}{1 - \beta +\widetilde{\mu} L} = 1$. However, the coefficient $TK$ of $\frac{\widetilde{\mu} L}{1 - \beta +\widetilde{\mu} L} $ will have a greater impact on the generalization bound than the coefficient of $\frac{1-\beta}{1 - \beta  +\widetilde{\mu} L}$ of $\frac{2\sigma L_G}{UL}\frac{1+6\sqrt{m}\kappa_{\psi}}{m}$, because $TK$ increases linearly with the number of iterations $T$ while the latter remains constant. Therefore, increasing $\beta$ will improve the generalization of the algorithm. As demonstrated in Figure \ref{fig:hyper}(a) and Table \ref{ta:dfedcata_beta} of Section \ref{sec:ablation_section}, increasing the value of $\beta$ has a significant impact on generalization, and its convergence speed also increases. Secondly, the communication topology has a significant impact on the generalization bound, primarily reflected in $\kappa_\psi$. According to the expression of $\kappa_\psi$, when $\psi \in [0.3,0.6]$, it tightens bound (\ref{eq:gener_bound}), implying that the generalization performance is poorer on communication topologies with either too good (such as a fully connected graph) or too poor (such as a ring graph) connectivity. This conclusion is verified in Section \ref{sec:topoaware}. Thirdly, excessively large $K$ will impair generalization performance due to the "client drift" caused by multiple rounds of local updates \cite{scaffold2020}, which is verified in Section \ref{sec:ablation_section} Figure \ref{fig:hyper}(b). On the theoretical level, unlike previous DFL works such as DFedAvgM and DFedSAM that only analyze convergence bounds, we are the first to additionally provide a generalization bound analysis. Consequently, we do not have generalization error bounds for the previous baseline methods, so comparisons can only be made at the experimental level.
\end{remark}

\section{Experiment}\label{sec:experiment}

We will verify the generalization performance and convergence speed of \method{} through simulations of various non-iid scenarios, compared with four CFL and DFL baseline methods. For more experimental results and details, please refer to the \textbf{Appendix} \ref{sec:appendix_A}.

% In this section, we will verify the generalization performance and convergence speed of \method{} through simulations of various non-iid scenarios. We will compare it with three baseline methods from CFL and four baseline methods from DFL. For more results and details, please refer to the \textbf{Appendix} \ref{sec:appendix_A}.

\subsection{Experiment Setup}\label{sec:Experiment_Setup}
\textbf{Datasets and partitioning methods.} We evaluate the proposed \method{} on CIFAR10\&100 datasets \cite{krizhevsky2009learning} in both IID and non-IID settings. To replicate non-IID data distribution across federated clients, we employ the Dirichlet distribution \cite{Hsu2019Measuring} and the Pathological distribution \cite{Zhang2020Personalized}. More specifically, the local data of each client is partitioned using label ratios sampled from the Dirichlet distribution Dir($\alpha$). A lower value of $\alpha$ signifies a greater degree of non-IID. In our experiments, we designate $\alpha=0.3$ and $\alpha=0.6$ to denote varying levels of non-IID. Meanwhile, within the Pathological distribution, each client's local data is partitioned by sampling label ratios from the Pathological distribution Path($\alpha$), with $\alpha$ denoting the number of classes each client owns. For instance, when using the CIFAR10 dataset, we assign $\alpha = 2$ to signify that each client has solely 2 randomly chosen classes from the available 10. In our experimental arrangement, we establish $\alpha \in \{2,4,6\}$ for the CIFAR-10 dataset, and $\alpha \in \{10,20,30\}$ for the CIFAR100 dataset respectively.

\textbf{Baselines.} 
The baselines for comparison encompass various state-of-the-art (SOTA) techniques in both the CFL and DFL scenarios. More specifically, the centralized baselines consist of FedAvg \cite{mcmahan2017communication}, FedSAM \cite{caldarola2022improving,qu2022generalized}, FedProx \cite{fedprox} and SCAFFOLD \cite{scaffold2020}. In the decentralized setting, D-PSGD \cite{lian2017can}, DFedAvg, DFedAvgM \cite{Sun2022Decentralized}, and DFedSAM \cite{shi2023improving} are utilized for comparison. It is important to note that both our baseline and proposed algorithms are designed to function synchronously. It is noteworthy that, in comparison to other methods, SAM-based optimization methods such as DFedSAM and FedSAM incur twice the computational workload during the local update stage due to the need for twice parameter update computations. Similarly, SCAFFOLD requires twice the communication volume during the communication stage due to the transmission of correction variables.

\begin{table*}[ht]  
% \vspace{-0.3cm}
    \footnotesize  
    \centering 
    \caption{\small Top 1 test accuracy (\%) on two datasets in both IID and non-IID settings.}  
    \renewcommand{\arraystretch}{0.95}  
    \resizebox{1.0\textwidth}{!}{%
    \begin{tabular}{cccc!{\vrule width \lightrulewidth}ccc}  
        \toprule  
        \multirow{2}{*}{Algorithm}& \multicolumn{3}{c!{\vrule width \lightrulewidth}}{CIFAR-10}  & \multicolumn{3}{c}{CIFAR-100} \\  
        \cmidrule{2-7}  
        & \multicolumn{1}{c}{Dir 0.3} & \multicolumn{1}{c}{Dir 0.6} & \multicolumn{1}{c!{\vrule width \lightrulewidth}}{IID} &
         \multicolumn{1}{c}{Dir 0.3} & \multicolumn{1}{c}{Dir 0.6} & \multicolumn{1}{c}{IID} \\    
        \midrule  
        FedAvg  \cite{mcmahan2017communication}    & 78.10 $\pm$  0.81 & 79.00 $\pm$  1.04 & 81.16 $\pm$  0.20
        & 53.29 $\pm$  0.70 & 54.31 $\pm$  0.72   & 55.67 $\pm$  0.41 \\ 

        FedProx  \cite{li2020federated}    & 76.14 $\pm$  0.56 & 77.51 $\pm$  0.75 & 79.14 $\pm$  0.20
        & 38.34 $\pm$  0.46 & 38.60 $\pm$  0.38   & 38.40 $\pm$  0.23 \\
        
        FedSAM   \cite{qu2022generalized}   & 80.22 $\pm$  0.70 & 81.43 $\pm$  0.28 & 82.83 $\pm$  0.23
        & 56.01 $\pm$  0.36 & 55.03 $\pm$  0.51   & 57.29 $\pm$  0.19 \\ 

        SCAFFOLD   \cite{scaffold2020}   & 78.39 $\pm$  0.35 & 79.85 $\pm$  0.18 & 81.75 $\pm$  0.23
        & 55.92 $\pm$  0.30 & 57.57 $\pm$  0.52   & 60.56 $\pm$  0.31 \\ 
        \midrule        
        D-PSGD    \cite{lian2017can}   & 59.76 $\pm$  0.04 & 60.03 $\pm$  0.13 & 62.93 $\pm$  0.12
        & 47.05 $\pm$  0.20 & 48.07 $\pm$  0.10   & 49.76 $\pm$  0.12 \\    

        DFedAvg    \cite{Sun2022Decentralized}   & 77.25 $\pm$  0.12 & 77.83 $\pm$  0.11 & 79.97 $\pm$  0.08
        & 56.04 $\pm$  0.10 & 57.17 $\pm$  0.50   & 58.36 $\pm$  0.31 \\     
        
        DFedAvgM   \cite{Sun2022Decentralized}    & 79.30 $\pm$  0.24 & 80.66 $\pm$  0.07 & 82.72 $\pm$  0.20
        & 56.34 $\pm$  0.29 & 56.89 $\pm$  0.39   & 58.84 $\pm$  0.47 \\     
        
        DFedSAM   \cite{shi2023improving}    & 79.37 $\pm$  0.07 & 80.47 $\pm$  0.09 & 82.14 $\pm$  0.09
        &56.01 $\pm$  0.20  & 56.77 $\pm$  0.23   & 58.55 $\pm$  0.24 \\   
        \midrule

        \method{}   & \textbf{82.88} $\pm$ 0.19 & \textbf{83.12} $\pm$ 0.10 & \textbf{84.24} $\pm$ 0.20 & \textbf{60.71} $\pm$ 0.15 & \textbf{61.38} $\pm$ 0.12 & \textbf{63.22} $\pm$ 0.21\\ 
        
        \midrule  
        \multirow{2}{*}{Algorithm}& \multicolumn{3}{c!{\vrule width \lightrulewidth}}{CIFAR-10}  & \multicolumn{3}{c}{CIFAR-100} \\  
        \cmidrule{2-7}  
        & \multicolumn{1}{c}{Path 2} & \multicolumn{1}{c}{Path 4} & \multicolumn{1}{c!{\vrule width \lightrulewidth}}{Path 6} &
         \multicolumn{1}{c}{Path 10} & \multicolumn{1}{c}{Path 20} & \multicolumn{1}{c}{Path 30} \\    
        \midrule  
        FedAvg \cite{mcmahan2017communication} & 69.01 $\pm$ 0.96 & 75.80 $\pm$ 1.01 & 76.76 $\pm$ 1.11 & 45.92 $\pm$ 0.70 & 52.56 $\pm$ 0.63 & 53.49 $\pm$ 0.26 \\

        FedProx \cite{li2020federated} & 69.08 $\pm$ 0.45 & 76.17 $\pm$ 0.83 & 77.82 $\pm$ 0.95 & 36.17 $\pm$ 0.56 & 38.66 $\pm$ 0.24 & 39.12 $\pm$ 0.43 \\
        
        FedSAM \cite{qu2022generalized} & 69.33 $\pm$ 1.32 & 75.78 $\pm$ 1.04 & 77.55 $\pm$ 0.68 & 45.47 $\pm$ 0.69 & 51.31 $\pm$ 0.55 & 53.69 $\pm$ 0.36 \\

        SCAFFOLD \cite{scaffold2020} & 64.28 $\pm$ 2.12 & 78.08 $\pm$ 0.39 & 80.35 $\pm$ 0.19 & 46.73 $\pm$ 0.51 & 53.44 $\pm$ 0.67 & 56.54 $\pm$ 0.35 \\
          \midrule      
        D-PSGD \cite{lian2017can} & 59.53 $\pm$ 0.40 & 63.80 $\pm$ 0.22 & 64.86 $\pm$ 0.48 & 42.99 $\pm$ 0.58 & 46.79 $\pm$ 0.11 & 48.11 $\pm$ 0.35 \\

        DFedAvg \cite{Sun2022Decentralized} & 74.73 $\pm$ 0.26 & 77.50 $\pm$ 0.25 & 79.00 $\pm$ 0.25 & 51.25 $\pm$ 0.19 & 55.95 $\pm$ 0.11 & 56.62 $\pm$ 0.24 \\
        
        DFedAvgM \cite{Sun2022Decentralized} & 75.10 $\pm$ 0.27 & 79.50 $\pm$ 0.33 & 81.07 $\pm$ 0.27 & 44.10 $\pm$ 0.45 & 53.87 $\pm$ 0.64 & 54.69 $\pm$ 0.85 \\
        
        DFedSAM \cite{shi2023improving} & 75.08 $\pm$ 0.11 & 79.85 $\pm$ 0.09 & 81.36 $\pm$ 0.11 & 47.98 $\pm$ 0.17 & 54.31 $\pm$ 0.22 & 56.28 $\pm$ 0.21 \\
        \midrule

        \method{}   & \textbf{78.83} $\pm$ 0.19 & \textbf{82.37} $\pm$ 0.10 & \textbf{83.43} $\pm$ 0.20 & \textbf{54.70} $\pm$ 0.19 & \textbf{59.55} $\pm$ 0.10 & \textbf{60.92} $\pm$ 0.20\\
        
        \bottomrule  
    \end{tabular}  }
    \vspace{-1em}
    \label{ta:all_baselines}  
\end{table*}

\subsection{Performance Evaluation}\label{section:exper-evaluation}
\begin{figure*}[ht]
\vspace{-0.3cm}
\begin{center}
\subfloat[CIFAR-10-Path]{
    	\includegraphics[width=1\textwidth]{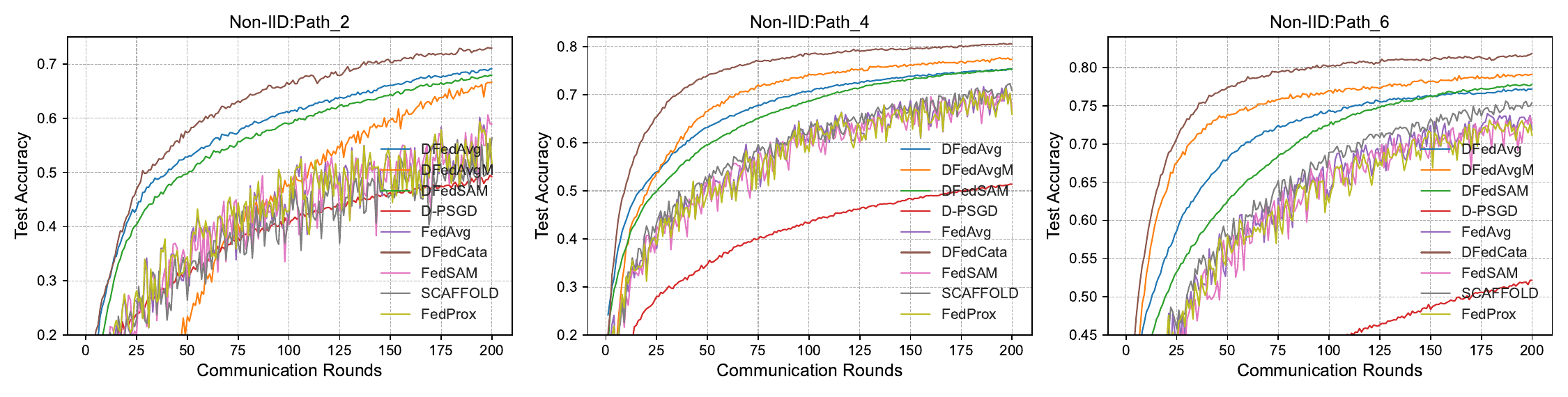}
    }
    \hfill
\subfloat[CIFAR-10-Dir]{
    	\includegraphics[width=1\textwidth]{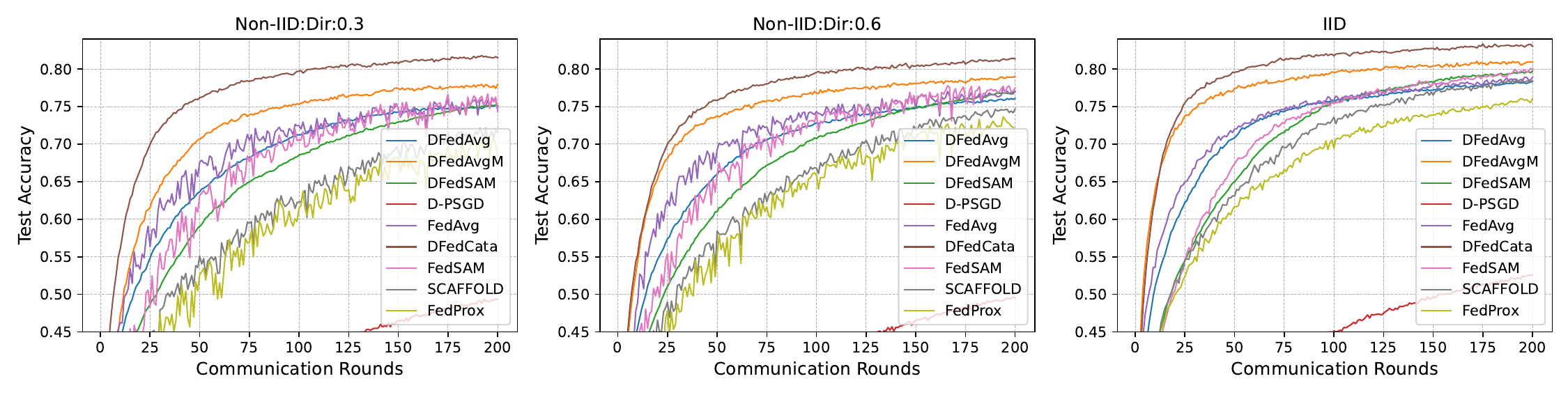}
    }
\end{center}
\vspace{-0.3cm}
\caption{ \small Test accuracy of all baselines on CIFAR-10 in both IID and different non-IID settings.}
\label{fig:Compared_baselines-cifar10}
\end{figure*}

\begin{figure*}[ht]
\vspace{-0.3cm}
\begin{center}
\subfloat[CIFAR-100-Path]{
    \includegraphics[width=1\textwidth]{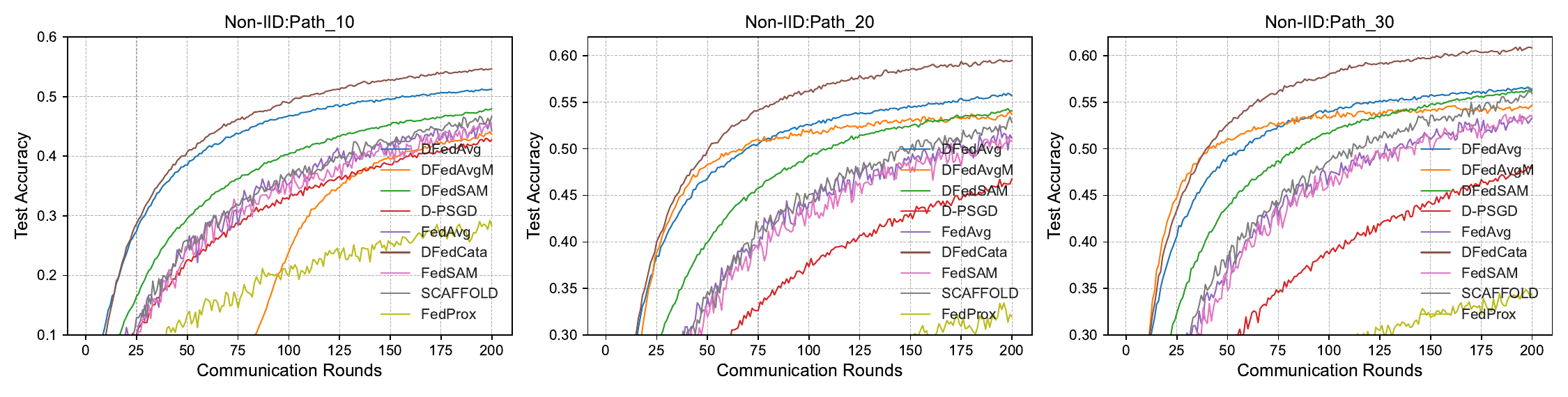}
}
    \hfill
\subfloat[CIFAR-100-Dir]{
    \includegraphics[width=1\textwidth]{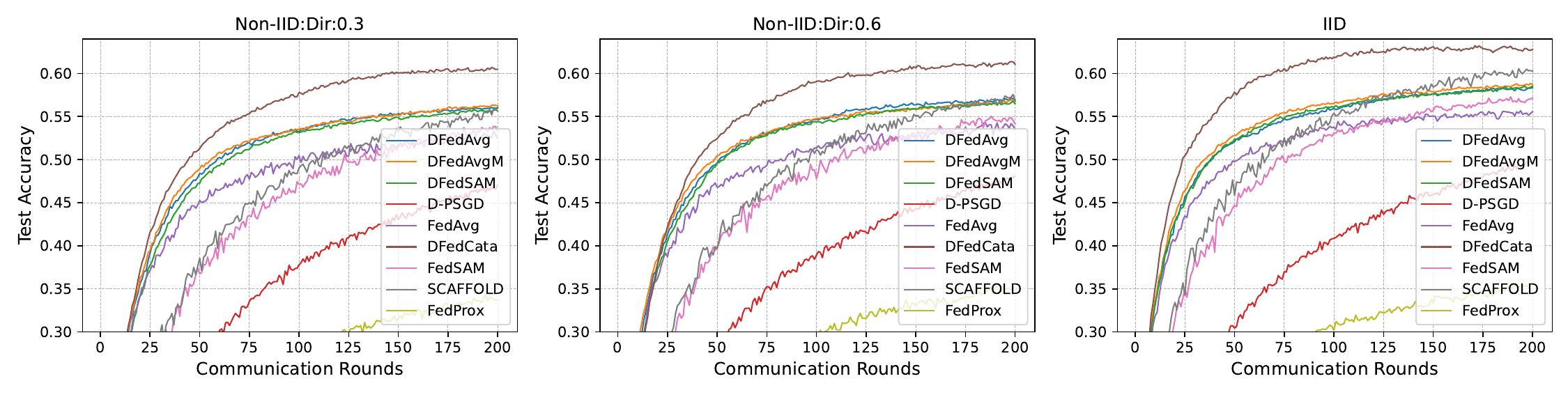}
}
\end{center}
\vspace{-0.3cm}
\caption{ \small Test accuracy of all baselines on CIFAR-100 in both IID and different non-IID settings.}
\label{fig:Compared_baselines-cifar100}
\vspace{-0.2cm}
\end{figure*}

Overall, Figures \ref{fig:Compared_baselines-cifar10} and \ref{fig:Compared_baselines-cifar100} present a comparison between the baseline approach and \method{} on the CIFAR10\&100 datasets under Dirichlet and Pathological distributions. The figures indicate that \method{} attains rapid convergence with fewer communication rounds, and outperforms all baseline methods in terms of generalization performance and convergence speed. Notably, under the Pathological distribution, CFL methods show training instability, while DFL methods are more stable.

\textbf{Performance Analysis.} In Table \ref{ta:all_baselines}, we conduct a series of experiments to compare the performance of our method with baseline methods. The results show that under various non-IID settings, our method consistently outperforms other methods, For instance, on the CIFAR10 dataset, our method surpasses the previous SOTA method DFedSAM by a minimum of 2.5\%, and up to 3.75\% in both Dirichlet and Pathological distributions. Because SAM-based approaches such as DFedSAM \cite{shi2023improving} and FedSAM \cite{qu2022generalized} require gradient ascent followed by gradient descent during updates, resulting in twice the computational workload compared to \method{}, our method not only saves computational resources but also achieves superior generalization performance. Even in the IID case, our method also outperforms DFedSAM by at least 2.1\%. Similarly, these results are also observed on the more complex CIFAR-100 dataset, where compared to DFedSAM, our method provides at least a 4.6\% improvement under Dirichlet distribution and at least a 4.7\% improvement under Pathological distribution. The results demonstrate the outstanding generalization capability of \method{}.

\textbf{Impact of non-IID levels (\(\alpha\)).} The robustness of \method{} in various non-IID settings is evident from Figure \ref{fig:Compared_baselines-cifar10}\& \ref{fig:Compared_baselines-cifar100} and Table \ref{ta:all_baselines}. A smaller value of alpha (\(\alpha\)) indicates a higher level of non-IID, which makes the task of optimizing a consensus problem more challenging. However, our algorithm consistently outperforms all baselines across different levels of non-IID.  For instance, on the CIFAR-10 dataset, under the Dirichlet distribution, the performance of our algorithm \method{} only decreases by 1.36\% from the IID to Dir 0.3, which significantly outperforms DFedSAM (2.76\%), DFedAvgM (3.42\%), DFedAvg (2.72\%), D-PSGD (3.17\%), and FedAvg (3.06\%), FedProx(3.00\%), FedSAM (2.61\%), SCAFFOLD (3.36\%) respectively. Furthermore, our algorithm's advantages become more pronounced with higher levels of non-IID. For example, on the CIFAR100 dataset using the Path 10 partition method, \method{} outperforms DFedSAM by 6.7\%, exceeding the performance improvements of 5.3\% under Path 20 and 4.64\% under Path 30. This result demonstrates our method's robustness to heterogeneous data.

\subsection{Convergence Speed}\label{sec:appendix_A_3}

The following curves depict the convergence of \method{} and other baseline methods on the CIFAR10\&100 datasets under Dirichlet and Pathological distributions. It is evident from the graphs that \method{} consistently outperforms other methods in terms of both convergence speed and generalization performance, regardless of the data distribution.

\begin{table*}[ht]
\vspace{-0.2cm}
\centering
\footnotesize
\caption{Communication rounds for each method achieving target accuracy on the CIFAR-10 dataset.}
\label{ta:CIFAR10-convergence}
\resizebox{1.0\textwidth}{!}{%
\begin{tabular}{cccc|ccc|ccc}
\toprule
\multirow{2}{*}{Methods} & \multicolumn{3}{c|}{Dir 0.3} & \multicolumn{3}{c|}{Dir 0.6} & \multicolumn{3}{c}{IID} \\  
\cmidrule{2-10}  
 & Acc@75      & Acc@77      & Acc@80      & Acc@75         & Acc@77         & Acc@80      & Acc@78         & Acc@80         & Acc@82    \\ \midrule  
FedAvg \cite{mcmahan2017communication}      & 141 (1.3$\times$) & 244 (1.7$\times$) & \textgreater 500 (1.0$\times$) & 111 (1.4$\times$) & 166 (1.7$\times$) & 485 (1.0$\times$) & 150 (1.1$\times$) & 243 (2.0$\times$) & \textgreater 500 (1.0$\times$)\\  
FedProx  \cite{fedprox}    & 334 (0.5$\times$) & \textgreater 500 (0.8$\times$) & \textgreater 500 (1.0$\times$) & 259 (0.6$\times$) & 420 (0.7$\times$) & \textgreater 500 (1.0$\times$) & 321 (0.5$\times$) & \textgreater 500 (1.0$\times$) & \textgreater 500 (1.0$\times$)\\  
FedSAM \cite{qu2022generalized}   & 141 (1.3$\times$) & 202 (2.1$\times$) & 402 (1.3$\times$) & 121 (1.3$\times$) & 165 (1.7$\times$) & 296 (1.7$\times$) & 142 (1.2$\times$) & 199 (2.4$\times$) & 363 (1.4$\times$) \\
SCAFFOLD \cite{scaffold2020}   & 264 (0.7$\times$) & 356 (1.2$\times$) & \textgreater 500 (1.0$\times$) & 202 (0.8$\times$) & 262 (1.1$\times$) & 470 (1.1$\times$) & 180 (0.9$\times$) & 273 (1.8$\times$) & \textgreater 500 (1.0$\times$) \\

\midrule
D-PSGD \cite{lian2017can}     & \textgreater 500 (0.4$\times$) & \textgreater 500 (0.8$\times$) & \textgreater 500 (1.0$\times$) & \textgreater 500 (0.3$\times$) & \textgreater 500 (0.6$\times$) & \textgreater 500 (1.0$\times$) & \textgreater 500 (0.3$\times$) & \textgreater 500 (1.0$\times$) & \textgreater 500 (1.0$\times$) \\  
DFedAvg \cite{Sun2022Decentralized}    & 179 (1.0$\times$) & 419 (1.0$\times$) & \textgreater 500 (1.0$\times$) & 152 (1.0$\times$) & 283 (1.0$\times$) & \textgreater 500 (1.0$\times$) & 176 (1.0$\times$) & 479 (1.0$\times$) & \textgreater 500 (1.0$\times$) \\  
DFedAvgM \cite{Sun2022Decentralized}   & 93 (1.9$\times$) & 141 (3.0$\times$) & \textgreater 500 (1.0$\times$) & 64 (2.4$\times$) & 99 (2.9$\times$) & 305 (1.6$\times$) & 59 (2.9$\times$) & 117 (4.1$\times$)  & 303 (1.7$\times$) \\  
DFedSAM  \cite{shi2023improving}   & 187 (1.0$\times$) & 265 (1.6$\times$)  & \textgreater 500 (1.0$\times$) & 155 (1.0$\times$) & 203 (1.4$\times$) & 414 (1.2$\times$) & 143 (1.2$\times$) & 212 (2.3$\times$) & 452 (1.1$\times$) \\  
\midrule
\method{}   & \textbf{42 (4.3$\times$)}& \textbf{56 (7.4 $\times$)}& \textbf{107 (5.0$\times$)}& \textbf{45 (3.4 $\times$)}& \textbf{58 (4.9$\times$)}& \textbf{120 (4.2$\times$)}& \textbf{36 (4.7$\times$)}& \textbf{56 (8.6$\times$)}& \textbf{96 (5.2$\times$)}\\  
\midrule
\multirow{2}{*}{Methods} & \multicolumn{3}{c|}{Path 2} & \multicolumn{3}{c|}{Path 4} & \multicolumn{3}{c}{Path 6} \\  
\cmidrule{2-10} 
 & Acc@55      & Acc@65      & Acc@70      & Acc@65         & Acc@70         & Acc@75      & Acc@70         & Acc@75         & Acc@80   \\ \midrule  
FedAvg \cite{mcmahan2017communication}     & 137 (0.5$\times$) & 283 (0.5$\times$) & \textgreater 500 (0.4$\times$) & 113 (0.5$\times$) & 180 (0.5$\times$) & 327 (0.6$\times$) & 129 (0.5$\times$) & 208 (0.5$\times$) & \textgreater 500 (1.0$\times$)\\  
FedSAM \cite{qu2022generalized}   & 156 (0.4$\times$) & 313 (0.4$\times$) & 453 (0.5$\times$) & 119 (0.5$\times$) & 172 (0.5$\times$) & 333 (0.6$\times$) & 132 (0.4$\times$) & 225 (0.5$\times$) & \textgreater 500 (1.0$\times$) \\
SCAFFOLD  \cite{scaffold2020}  & 183 (0.3$\times$) & 439 (0.3$\times$) & \textgreater 500 (0.4$\times$) & 122 (0.5$\times$) & 172 (0.5$\times$) & 287 (0.7$\times$) & 114 (0.5$\times$) & 182 (0.6$\times$) & 439 (1.1$\times$) \\

FedProx  \cite{fedprox}   & 172 (0.4$\times$) & 283 (0.5$\times$) & \textgreater 500 (0.4$\times$) & 115 (0.5$\times$) & 180 (0.5$\times$) & 355 (0.5$\times$) & 135 (0.4$\times$) & 233 (0.5$\times$) & \textgreater 500 (1.0$\times$)\\

\midrule
D-PSGD   \cite{lian2017can}   & 340 (0.18$\times$) & \textgreater 500 (0.3$\times$) & \textgreater 500 (0.4$\times$) & \textgreater 500 (0.1$\times$) & \textgreater 500 (0.2$\times$) & \textgreater 500 (0.4$\times$) & \textgreater 500 (0.1$\times$) & \textgreater 500 (0.2$\times$) & \textgreater 500 (1.0$\times$) \\  
DFedAvg  \cite{Sun2022Decentralized}   & 62 (1.0$\times$) & 139 (1.0$\times$) & 223 (1.0$\times$) & 59 (1.0$\times$) & 92 (1.0$\times$) & 187 (1.0$\times$) & 59 (1.0$\times$) & 113 (1.0$\times$) & \textgreater 500 (1.0$\times$) \\  
DFedAvgM \cite{Sun2022Decentralized}   & 126(0.5$\times$) & 179(0.8$\times$) & 244(0.9$\times$) & 46(1.3$\times$) & 64(1.4$\times$) & 119(1.6$\times$) & 32(1.8$\times$) & 65(1.7$\times$) & 257(1.9$\times$) \\  
DFedSAM  \cite{shi2023improving}   & 75(0.8$\times$) & 157(0.9$\times$) & 248(0.9$\times$) & 75 (0.8$\times$)& 111(0.8$\times$) & 187(1.0$\times$) & 84(0.7$\times$) & 128(0.9$\times$) & 328(1.5$\times$) \\  
\midrule
\method{}   & \textbf{44 (1.4$\times$)} & \textbf{87(1.6$\times$)} & \textbf{137(1.6$\times$)} & \textbf{24(2.5$\times$)} & \textbf{35(2.6$\times$)} & \textbf{58(3.2$\times$)} & \textbf{24(2.5$\times$)} & \textbf{36(3.1$\times$)} & \textbf{90(5.6$\times$)} \\  
\bottomrule
\end{tabular}}
\vspace{-0.3cm}
\end{table*}

\textbf{Convergence speed.}\label{sec:convergence_speed} In Table \ref{ta:CIFAR10-convergence}, we quantify the communication rounds required for \method{} and various baseline methods to reach a specified level of accuracy across different non-IID scenarios in the CIFAR-10 dataset. 
The results show that \method{} achieves the desired accuracy threshold in fewer rounds across diverse non-IID scenarios, outperforming competing baseline methods and excelling in convergence speed. Specifically, under the Dirichlet distribution, our method achieves at least a 4.3$\times$ acceleration ratio and up to an 8.6$\times$ acceleration ratio compared to the baseline method, DFedAvg. Under the Pathological distribution, we achieve at least a 1.4$\times$ acceleration ratio and up to a 5.4$\times$ acceleration ratio compared to the DFedAvg method. This fully demonstrates the rapid convergence speed advantage of \method{}, underscoring its considerable practical utility in real-world applications.
Furthermore, under the premise that the local computation amount of each algorithm is the same, the acceleration ratio can also represent the saving ratio of the communication and computation total amounts compared to DFedAvg. In this sense, our algorithm significantly reduces both the communication and computation total amounts compared to other baseline methods.

\subsection{Topology-aware Performance}\label{sec:topoaware}

Below, we investigate the impact of different topologies on various DFL methods on the CIFAR-10 dataset with Dirichlet $\alpha=0.3$.

\begin{figure*}[ht]
\begin{center}
\subfloat{
    	\includegraphics[width=1.0\textwidth]{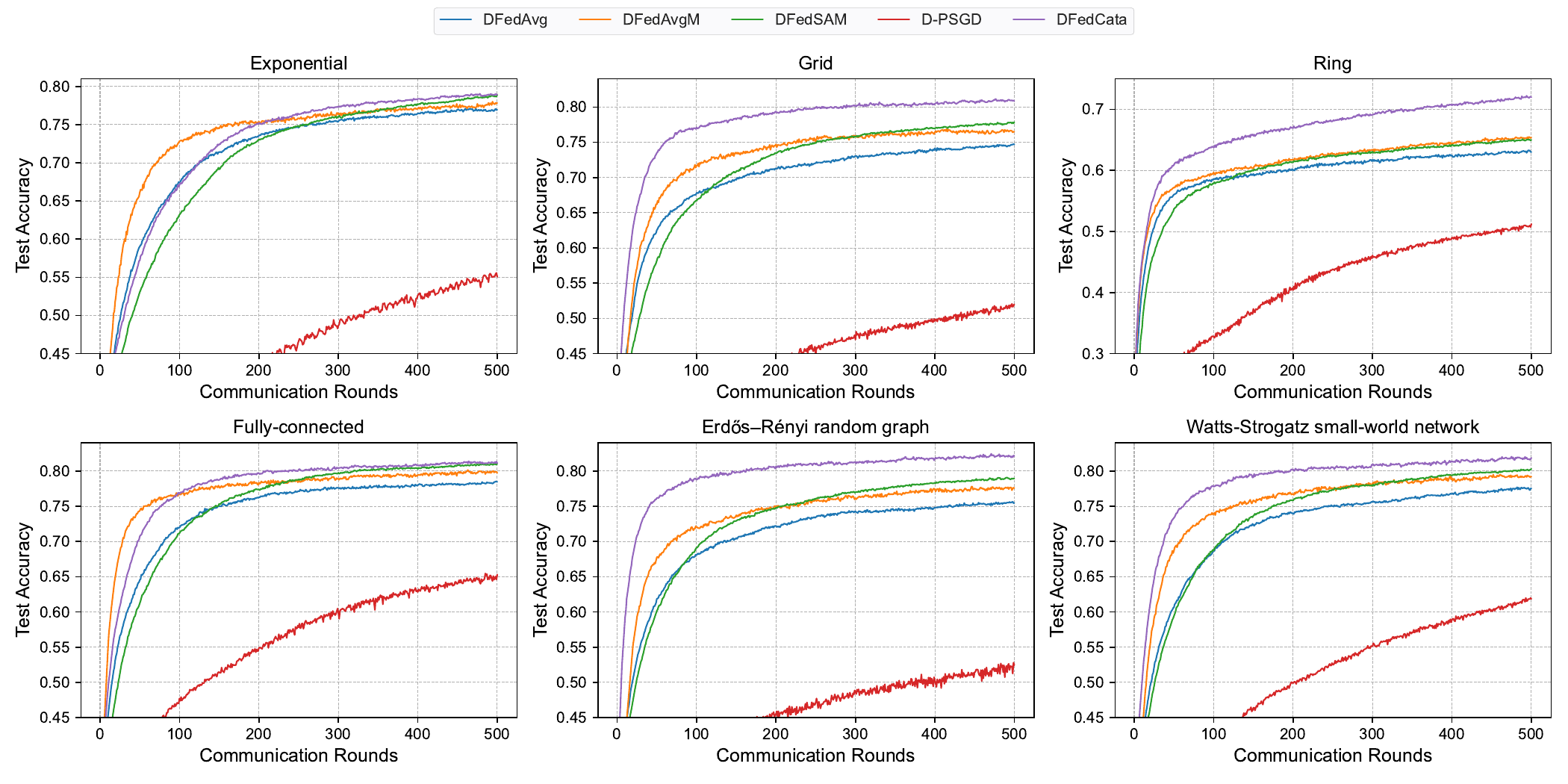}
    }
\end{center}
 \vspace{-0.4cm}
\caption{ \small Accuracy of different DFL algorithms with different decentralized topologies on the test dataset.}
\label{fig:topo}
 \vspace{-0.3cm}
\end{figure*}

\begin{table}[ht]
    \centering  
    \caption{\small Top 1 test accuracy (\%) in various network topologies compared with decentralized algorithms on CIFAR-10.}  
    \label{ta:topo}  
    \renewcommand{\arraystretch}{1}  
    \resizebox{\linewidth}{!}{%  
        \begin{tabular}{ccccccc}   
            \toprule  
            \multicolumn{1}{c}{Algorithm} & \multicolumn{1}{c}{Ring} & \multicolumn{1}{c}{Grid} & \multicolumn{1}{c}{Exp} & \multicolumn{1}{c}{Full}  & \multicolumn{1}{c}{ER} & \multicolumn{1}{c}{WS}\\   
            \midrule  
            D-PSGD        & 51.24 & 52.06 & 55.58 & 65.46 & 61.98 & 52.81            \\  
            DFedAvg       & 63.30 & 74.72 & 77.11 & 78.52  & 77.70 & 75.65            \\  
            DFedAvgM      & 65.49 & 76.89 & 78.01 & 80.14 & 79.40 & 77.80           \\  
            DFedSAM       & 65.12 & 77.86 & 78.88 & 81.04  & 80.32 & 79.07           \\  
            \midrule  
            \method{}     & \textbf{72.22} & \textbf{81.13} & \textbf{79.05} & \textbf{81.38} &\textbf{82.04} &\textbf{82.45}        \\  
            \bottomrule  
        \end{tabular}   
     }  
     \vspace{-0.3cm}
\end{table}

\textbf{Communication topology.} We test the performance of \method{} and other decentralized baseline methods on 6 common communication topologies. The exponential topology \cite{shi2023improving} establishes connections by selecting neighbors at a distance of 2 raised to the power of the client. In the scenario with 100 clients, each client is connected to other neighbors at distances $\{1,2,4,8,16,32,64\}$. The grid communication topology \cite{shi2023improving} connects a client to its four adjacent neighbors. For the Ring topology \cite{shi2023improving}, a client is connected to its two adjacent neighbors. In the fully connected topology \cite{shi2023improving}, a client is linked to all participants in the system. The Erdős–Rényi random graph \cite{gilbert1959random} randomly connects with other users in the system with a probability of $p=0.1$. Lastly, the Watts-Strogatz small-world network \cite{watts1998collective} connects a client to the eight adjacent neighbors and also has a 0.02 probability of connecting to other users.

\textbf{The performance on different topologies.} Table \ref{ta:topo} and Figure \ref{fig:topo} present the comparison between different decentralized baseline methods and \method{}, demonstrating the strong generalization performance and convergence speed of \method{} across various communication topologies. This validates the robustness of \method{} on diverse communication topologies. Particularly, the generalization performance of \method{} shines on the Watts-Strogatz small-world network, which can simulate a variety of real-world networks such as social networks \cite{watts1998collective}, further affirming the effectiveness of \method{} in practical deployment scenarios. Moreover, the performance improvement is most pronounced on the ring topology, with an impressive 7.1\% increase compared to the DFedSAM method, surpassing the performance gains observed on other communication topologies. This attests to the accelerated generalization effect of \method{}.

\subsection{Ablation Study}\label{sec:ablation_section}

We verify the influence of each component and hyperparameter in \method{}. All the ablation studies are conducted on the same topology as used in Section \ref{section:exper-evaluation}.

\begin{table}[ht]
    \centering
    \caption{Accuracy and Communication Rounds Required to Reach 75\% Accuracy for DFedCata under Different Beta}
    \begin{tabular}{lccccccc}
        \toprule
        $\beta$ Values       & 0    & 0.2  & 0.4  & 0.7  & 0.8  & 0.9  & 0.99 \\
        \midrule
        Acc (\%)        & 77.17 & 79.33 & 80.24 & 81.47 & 81.61 & 81.54 & 81.69 \\
        Acc@75  & 211   & 132   & 96    & 80    & 78    & 78    & 75   \\
        \bottomrule
    \end{tabular}
    \label{ta:dfedcata_beta}
\end{table}

\begin{figure}[ht]  
\vspace{-0.2cm}  
    \centering  
    \includegraphics[width=0.5\textwidth]{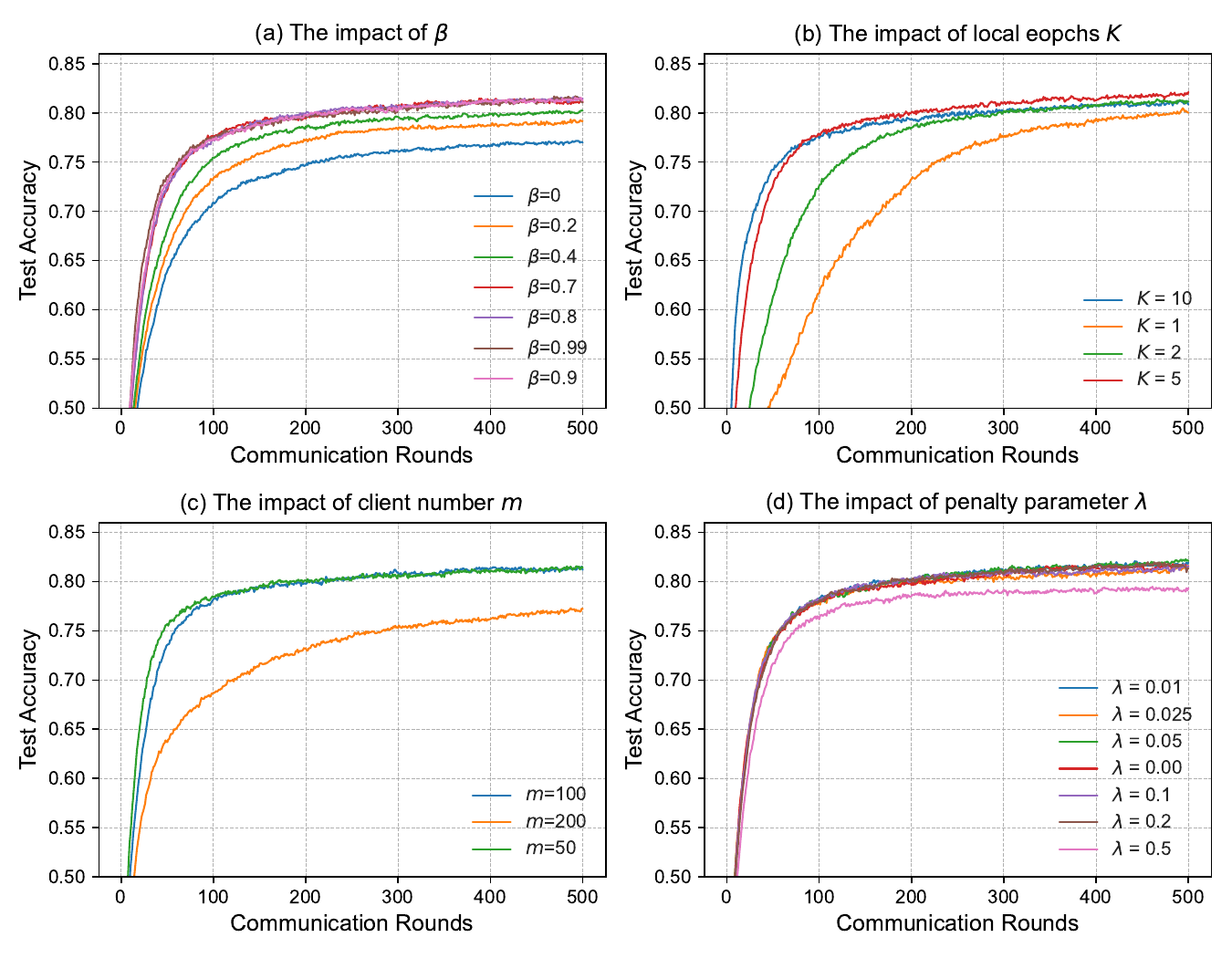}  
    \vspace{-0.2cm}  
    \caption{ \small Hyperparameter Sensitivity: local epochs $K$, Catalyst parameter $\beta$, number of participated clients $m$, penalty parameter $\lambda$.}  
    \label{fig:hyper}  
    \vspace{-1em}  
\end{figure}

 \textbf{Impact of Catalyst Parameter $\beta$.} The convergence curves under different catalyst parameters $\beta$ after 500 communication rounds are depicted in Figure \ref{fig:hyper} (a) for the CIFAR-10 dataset under the Dirichlet 0.3 distribution. A larger \(\beta\) value implies a better acceleration effect. We test the performance of the algorithm for $\beta \in \{0,0.2,0.4,0.7,0.8,0.9,0.99\}$, and the results indicate that the convergence speed and generalization performance are optimal when $\beta = 0.99$. This is consistent with the discussion about $\beta$ in Remark \ref{Remark:gener}. More specifically, As shown in the Table \ref{ta:dfedcata_beta}, the accuracy of DFedCata increases with larger $\beta$ values, stabilizing around 81.6\% when $\beta > 0.7$. Additionally, as $\beta$ increases, the convergence speed of DFedCata improves. We set the accuracy threshold at 75\%, and the number of communication rounds required for the algorithm to reach this threshold is indicated in Table \ref{ta:dfedcata_beta}. Clearly, as $\beta$ increases, the algorithm converges faster, which is consistent with our convergence rate conclusions (see Theorem \ref{the:opt_error} and Remark \ref{remark:optimization}).

\textbf{Impact of Client Number $m$.} In Figure \ref{fig:hyper} (c), we present the performance with different numbers of participants, \(m = \{50, 100, 200\}\). We observe that, with an increase in the number of local data, \method{} achieves the best performance when \(m=50\) or 100, while a decrease in performance is observed when \(m\) is relatively large. This can be attributed to the fact that a smaller number of clients has a larger number of local training samples, leading to improved training performance. This is consistent with the discussion about $m$ in Remark \ref{Remark:gener}.

\textbf{Impact of Penalty parameter $\lambda$.} The penalty parameter \(\lambda\) is an additional factor affecting the convergence of \method{}. It trades off the consistency between local update directions and the client. A larger $\lambda$ emphasizes consistency, making it difficult for the client to optimize, while a smaller $\lambda$ prioritizes individual client optimization, leading to client drift and thereby increasing inconsistency. We evaluate \method{}'s generalization performance using different values of \(\lambda\) from the set \(\{0,0.025, 0.01, 0.05, 0.1, 0.2, 0.5\}\). Figure \ref{fig:hyper} (d) illustrates the highest accuracy achieved when \(\lambda=0.05\). Additionally, we observe that a larger \(\lambda\) leads to a slower convergence speed, consistent with our previous analysis.

\textbf{Impact of local epochs $K$.} $K$ represents the number of optimization rounds performed by each client. A larger value of $K$ is more likely to lead to the "client drift" phenomenon (see Remark \ref{Remark:gener}), resulting in increased inconsistency between clients, thereby affecting the algorithm's performance. We evaluate the generalization performance of \method{} using different values of \(K\) from the set \(\{1, 2, 5, 10\}\). Figure \ref{fig:hyper} (b) illustrates the highest accuracy achieved when \(K=5\). 
% Additionally, we observe that a larger $K$ leads to faster convergence in Figure \ref{fig:hyper} (b), consistent with our theoretically proven $\mathcal{O}(\frac{1}{\sqrt{KT}})$.
This result demonstrates that when $K$ is too large, it will affect the generalization performance of the model, which is consistent with the conclusion obtained in our generalization bound (see Remark \ref{Remark:gener}).

\begin{table}[ht]
    \centering
    \vspace{-0.5em}
    \caption{Effect of Moreau Envelope and Nesterov Acceleration on Accuracy}
    \begin{tabular}{ccc}
        \toprule
        Moreau Envelope & Nesterov Acceleration & Accuracy (\%) \\
        \midrule
                        &                        & 77.25 \\
        \checkmark      &                        & 77.17 \\
                        & \checkmark            & 82.42 \\
        \checkmark      & \checkmark            & 82.88 \\
        \bottomrule
    \end{tabular}
    \label{ta:moreau_nesterov}
    \vspace{-0.5em}
\end{table}

\textbf{The impact of Moreau Envelope and Nesterov Acceleration.} In Table \ref{ta:moreau_nesterov}, we set $\lambda=0$ to evaluate DFedCata with only Nesterov acceleration and $\beta=0$ to evaluate DFedCata with only the Moreau envelope. The table shows that using the Moreau envelope alone does not significantly improve generalization performance, a scenario akin to that of FedAvg \cite{mcmahan2017communication} and FedProx \cite{fedprox}. However, its smoothing effect enhances the overall performance when combined with Nesterov acceleration. Nesterov acceleration has a significant positive impact on the model's convergence speed and final performance, substantially increasing the model's accuracy. When both the Moreau envelope and Nesterov acceleration are used together, the model achieves the best accuracy, indicating that these two techniques complement each other and enhance the overall optimization effectiveness in this scenario.

\section{disscussion}

\subsection{Further discussion on the relationship between the spectral gap $1-\lambda$ and the convergence rate and generalization bound.} 

In contrast to the convergence rate derived in previous DFL methods \cite{shi2023improving,Sun2022Decentralized,lian2017can}, which is related to the spectral gap \(1 - \lambda\), their findings suggest that a smaller spectral gap leads to a faster convergence rate. However, the convergence rate of our proposed method is independent of the spectral gap. Experimentally, we highlight that a smaller spectral gap does not necessarily lead to faster convergence. As shown in Figure \ref{fig:topo}, with a target threshold set at 75\%, DFedCata requires approximately 80 rounds on the Fully-connected topology, whereas it needs about 40 rounds on the Erdős–Rényi random graph. There is no significant relationship between the convergence rate and the spectral gap.
The fact that the convergence rate is independent of the spectral gap is an inherent property of DFedCata.
% , not due to a lack of theoretical proof. 

Regarding the relationship between the generalization bound and the spectral gap: Although the convergence rate of DFedCata is unrelated by the spectral gap, its generalization bound is related to it. As noted in Remark \ref{Remark:gener} of Theorem \ref{the:generation_err}, when $\psi \in \{0.3, 0.6\}$, the generalization bound of DFedCata becomes tighter, implying better generalization performance. This is consistent with the results in Table \ref{ta:topo}, where topologies with smaller spectral gaps, such as Fully-connected, do not exhibit higher generalization accuracy compared to topologies with larger spectral gaps, such as WS.

In summary, the independence of the spectral gap from the convergence rate is a fundamental property of DFedCata, and experiments do not support the conclusion that a smaller spectral gap leads to faster convergence. Additionally, our generalization bound derivation establishes the relationship between the spectral gap and convergence rate, and the experimental results corroborate our conclusions.

\subsection{A theoretical explanation of the effectiveness of $\beta$.} As mentioned in the introduction of the main text, the two challenges in DFL are local inconsistency and the impact of heterogeneous data. The Moreau envelope function addresses these challenges by smoothing the loss function and introducing a penalty term that enhances client consistency. When combined with Nesterov’s extrapolation step, it significantly accelerates convergence within the smooth space created by the Moreau envelope. Theoretical analysis also shows that increasing $\beta$ can significantly reduce the term $\frac{KL^2(1-\beta)}{T}G^2$ and $ \frac{\sqrt{K}L^2(1-\beta)}{\sqrt{T}}\sigma^2$ in Theorem \ref{the:opt_error} caused by heterogeneous data and client inconsistency. In summary, the Moreau envelope enhances consistency among clients, while Nesterov’s extrapolation step mitigates the effects of heterogeneous data, which is why DFedCata demonstrates better generalization in experiments.
\section{Conclusion}\label{sec: conclusion}
In this work, we address the slow convergence and poor generalization of existing DFL optimizers by proposing the \method{} method. This method mainly consists of two components:  Moreau envelope and Nesterov acceleration. The Moreau envelope is primarily used to address parameter inconsistencies in client models caused by data heterogeneity. It optimizes the Moreau envelope function $h_i(\mathbf{x}, \mathbf{x}_{i,0}^t)$ during the local update phase to enhance consistency. Nesterov acceleration is the core step that enables acceleration during the client aggregation phase. It effectively enhances convergence speed and improves generalization performance. Theoretically, we not only analyze the convergence of \method{} but also examine its generalization performance, making us the first in the DFL domain to analyze generalization based on stability,  addressing the previous lack of understanding regarding algorithm generalization in DFL. Furthermore, we have obtained the properties of DFedCata and guidance on hyperparameter selection through theoretical analysis, and these conclusions have all been verified experimentally, demonstrating the practical utility of theoretical guidance. Theoretical analysis and extensive experiments confirm the superior generalization performance and convergence speed of \method{}.

\bibliographystyle{IEEEtran}
\bibliography{main.bib}

\onecolumn

\clearpage
\section*{\centering APPENDIX}
\appendices

% \begin{center}
%  \rule{6.875in}{0.7pt}\\ % 4.0
%  {\Large\bf Supplementary Material for\\ `` \method{}: Unleashing the Potential of Decentralized Federated Learning \\
% via Opposite Lookahead Enhancement''}
%  \rule{6.875in}{0.7pt}
% \end{center}

In this part, we provide supplementary materials including more detail about the experiment, the proof of the optimization, and generalization error. Here’s the table of contents for the \textbf{Appendix}.

\begin{itemize}
    \item \textbf{Appendix} \ref{sec:appendix_A}
    \begin{itemize}
        \item \textbf{Appendix} \ref{sec:appendix_A_1} Experiment Setup about Implementation Details.
        \item \textbf{Appendix} \ref{sec:appendix_A_2} Experiment Setup about Communication Configurations.
        % \item \textbf{Appendix} \ref{sec:appendix_A_3} Convergence Speed Curve.
        \item \textbf{Appendix} \ref{sec:appendix_A_4} Convergence curves under various communication topologies.      
    \end{itemize}
    \item \textbf{Appendix} \ref{proof}: Proof of the theoretical analysis.
    \begin{itemize}
        \item \textbf{Appendix} \ref{appendix:optimization error}: Proof for optimization error.
        \item \textbf{Appendix} \ref{sec:Proofs for the generalization Error}: Proof for generalization error.
    \end{itemize}
\end{itemize}

\section{More detail about Experiment}\label{sec:appendix_A}

\subsection{Experiment Setup about Implementation Details.}\label{sec:appendix_A_1}

The total number of clients is set to 100, with 10\% of the clients participating in the communication, creating a random bidirectional topology. For decentralized methods, all clients perform the local iteration step, while for centralized methods, only the participating clients perform the local update \cite{shi2023improving}. The local learning rate is initialized to 0.1 with a decay rate of 0.998 per communication round for all experiments. For SAM-based algorithms, such as DFedSAM and FedSAM, we set the perturbation weight as $\rho = 0.1$. For the parameters $\beta$ and local epochs $K$, we designated $\beta=0.99$ and $K=5$ for CIFAR-10, and $\beta=0.99$ and $K=2$ for CIFAR-100 in the random topology. It is noteworthy that this setting on CIFAR-100 is chosen to conserve computational resources while achieving favorable generalization performance. The maximum number of communication rounds is set to 500 for all experiments on CIFAR-10 and 200 for CIFAR-100. Additionally, all ablation studies are conducted on the CIFAR-10 dataset with a data partition method of Dir 0.3 and 500 communication rounds.

\subsection{Experiment Setup about Communication Configurations.}\label{sec:appendix_A_2}

\textbf{Communication Configurations.} To ensure a fair comparison between decentralized and centralized approaches, we have implemented a dynamic and time-varying connection topology for DFL methods. This approach ensures that the number of connections in each round does not exceed the number of connections in the centralized server, thus enabling the matching of communication volume between the decentralized and centralized methods as \cite{dai2022dispfl}. Additionally, Dai et al.\cite {dai2022dispfl} point out that the communication overhead in DFL is higher than that in CFL. This is because, in DFL, each client needs to send model parameters to its neighbors, whereas in CFL, only the clients selected by the server need to send their parameters. 
To ensure a fair comparison, we regulate the number of neighbors for each client in DFL using the client participation rate. Here, we set each client to randomly select 10 neighbors during each communication round. Other communication topologies, such as Grid, are generated according to corresponding rules.

\clearpage

% \clearpage

\subsection{Convergence curves under various communication topologies.}\label{sec:appendix_A_4}

We test the generalization performance and convergence speed of \method{} and other decentralized methods on six common communication topologies. The figures show that the optimal communication topology for \method{} appears to be random or WS small-world networks, as these types of networks can simulate real-world scenarios such as social networks. Therefore, this characteristic holds significant practical implications for deploying DFL training systems.

% \clearpage

\section{Proof of the theoretical analysis.}\label{proof}

\subsection{Proofs for the Optimization Error}\label{appendix:optimization error}

In this part, we prove the training error for our proposed method. We assume the objective $ f({\bf x}):=\frac{1}{m}\sum_{i=1}^m f_i({\bf x}) $ is $L$-smooth w.r.t $\mathbf{x}$. Then we could upper bound the training error in the FL. Some useful notations in the proof are introduced in the Table \ref{ta:notation}. Then we introduce some important lemmas used in the proof.

\begin{table}[ht!]
  \caption{Some abbreviations of the used terms in the proof of bounded training error.}
  \centering  
  \begin{tabular}{ccc}
    \toprule  
    Notation & Formulation  & Description \\  
    \midrule  
    $\mathbf{x}_{i,k}^t$ & - & parameters at $k$-th iteration in round $t$ on client $i$ \\  
    $\mathbf{x}_{i}^t$ & - & global parameters in round $t$ on client $i$\\  
    $A^t$ & $\bar{\mathbf{x}}^{t+1} - \bar{\mathbf{x}}^{t}$ & average difference between consecutive communications \\
    $B^t$ & $\sum_{k=0}^{K-1}\frac{\gamma_{k}}{\gamma}\frac{1}{m}\sum_{i=1}^m\mathbf{g}_{i,k}^{t}$ & weighted gradient in round $t$\\
    $D$ & $f(\mathbf{x}^{0})-f(\mathbf{x}^{\star})$  & bias between the initialization state and optimal \\
    \bottomrule  
  \end{tabular}  
  \label{ta:notation}
\end{table}  

\subsubsection{Important Lemmas}
\begin{lemma}\label{le:lemma1}
The double stochasticity of $\mathbf{W}=[w_{i,j}]_{m\times m}$, i.e., $\sum_{i=1}^m w_{i,j} = 1$ and $\sum_{j=1}^m w_{i,j} = 1$, leads to the following conclusion for any sequence $a_j$:
\begin{equation}
    \sum_{i=1}^m\sum_{j=1}^m w_{i,j}a_j = \sum_{i=1}^ma_i
\end{equation}
\begin{proof}
    According to the double stochasticity of $\mathbf{W}$, we obtain the following equation:
    \begin{equation*}
        \sum_{i=1}^m \sum_{j=1}^m w_{i,j}a_j = \sum_{j=1}^m \sum_{i=1}^m w_{i,j}a_j = \sum_{j=1}^ma_j = \sum_{i=1}^ma_i
    \end{equation*}
    This completes the proof.
\end{proof}
\end{lemma}

\begin{lemma}\label{le:local_update}
    \textbf{(Local updates)} For $\forall \ \mathbf{x}_{i,k}^{t} \in \mathbb{R}^{d}$, we denote $\delta_{i,k}^{t}=\mathbf{x}_{i,k}^{t}-\mathbf{x}_{i,k-1}^{t}$ with setting $\delta_{i,0}^{t}=0$, and $\Delta_{i,K}^{t}=\sum_{k=0}^{K}\delta_{i,k}^{t} = \mathbf{x}_{i,K}^{t}-\mathbf{x}_{i,0}^{t}$, under the update rule in Algorithm \ref{alg:DFedCata}, we have:
    \begin{equation}
        \mathbf{x}_{i,K}^{t}-\mathbf{x}_{i,0}^{t} =-\frac{\gamma}{\lambda}\sum_{k=0}^{K-1}\frac{\gamma_{k}}{\gamma}\mathbf{g}_{i,k}^{t} 
    \end{equation}
    where $\sum_{k=0}^{K-1}\gamma_{k}=\sum_{k=0}^{K-1}\eta\lambda\bigl(1-{\eta}{\lambda}\bigr)^{K-1-k}=\gamma=1-(1-{\eta}{\lambda})^{K}$.
\begin{proof}
    According to the update rule of Line.13 in Algorithm \ref{alg:DFedCata}, we have:
    \begin{align*}
        \delta_{k} 
        &= \Delta_{i,k}^{t} - \Delta_{i,k-1}^{t}= \mathbf{x}_{i,k}^{t}-\mathbf{x}_{i,k-1}^{t}\\
        &= -\eta\bigl(\mathbf{g}_{i,k-1}^{t} + {\lambda}(\mathbf{x}_{i,k-1}^{t}-\mathbf{x}_{i,0}^{t})\bigr) = -\eta(\mathbf{g}_{i,k-1}^{t} + {\lambda}\Delta_{i,k-1}^{t}).
    \end{align*}
    Then We can formulate the iterative relationship of $\Delta_{i,k}^{t}$ as:
    \begin{align*}
        \Delta_{i,k}^{t} = \Delta_{i,k-1}^{t} -\eta(\mathbf{g}_{i,k-1}^{t}  + {\lambda}\Delta_{i,k-1}^{t})=(1-\eta{\lambda})\Delta_{i,k-1}^{t} - \eta\mathbf{g}_{i,k-1}^{t}.
    \end{align*}
    Taking the iteration on $k$ and we have:
    \begin{align*}
        \mathbf{x}_{i,K}^{t}-\mathbf{x}_{i,0}^{t} = \Delta_{i,K}^{t}
        &= (1- \eta \lambda)^{K}\Delta_{i,0}^{t} -  \eta\sum_{k=0}^{K-1}(1- \eta \lambda)^{K-1-k}\mathbf{g}_{i,k}^{t}\\
        &\overset{(a)}{=}  -  \eta\sum_{k=0}^{K-1}(1- \eta \lambda)^{K-1-k}\mathbf{g}_{i,k}^{t}\\
        &= -\frac{1}{\lambda}\sum_{k=0}^{K-1} \eta \lambda(1- \eta \lambda)^{K-1-k}\mathbf{g}_{i,k}^{t} \\
        &= -\frac{\gamma}{\lambda}\sum_{k=0}^{K-1}\frac{\gamma_{k}}{\gamma}\mathbf{g}_{i,k}^{t}.
    \end{align*}
    (a) applies $\Delta_{i,0}^{t}=\delta_{i,0}^{t}=0$.\\
\end{proof}
\end{lemma}

\begin{lemma}\label{le:bound local update}
    \textbf{(Bound local update)} In accordance with the conditions in Lemma \ref{le:local_update}, we have the following.
    \begin{equation*}
        \begin{aligned}
        \mathbb{E}\|\mathbf{x}_{i,k}^{t}-\mathbf{x}_{i,0}^{t}\|^2 \leq \frac{\gamma^2}{\lambda^2}\sum_{k=0}^{K-1}\frac{\gamma_{k}}{\gamma}\mathbb{E}\|\mathbf{g}_{i,k}^{t}\|^2 
        \end{aligned}
    \end{equation*}
    \begin{proof}
        Firstly, the conclusion of Lemma 2 holds for all $k \leq K$. To avoid confusion, we will use $j$ as the index, and the conclusion of Lemma 2 is as follows:
$$
\mathbf{x}_{i,k}^{t}-\mathbf{x}_{i,0}^{t} =-\frac{\gamma}{\lambda}\sum_{j=0}^{k-1}\frac{\gamma_{j}}{\gamma}\mathbf{g}_{i,j}^{t} = -\frac{\sum_{p=0}^{k-1}\gamma_p}{\lambda}\sum_{j=0}^{k-1}\frac{\gamma_{j}}{\sum_{p=0}^{k-1}\gamma_p}\mathbf{g}_{i,j}^{t}
$$
Let $\tilde{\gamma} = \sum_{p=0}^{k-1}\gamma_p$, then
$$
\mathbb{E}\|\mathbf{x}_{i,k}^{t}-\mathbf{x}_{i,0}^{t}\|^2 = \mathbb{E}\|\frac{\tilde{\gamma}}{\lambda}\sum_{j=0}^{k-1}\frac{\gamma_{j}}{\tilde{\gamma}}\mathbf{g}_{i,j}^{t}\|^2 = \frac{\tilde{\gamma}^2}{\lambda^2}\mathbb{E}\|\sum_{j=0}^{k-1}\frac{\gamma_{j}}{\tilde{\gamma}}\mathbf{g}_{i,j}^{t}\|^2
$$

According to Jensen’s inequality, we have

$$
\frac{\tilde{\gamma}^2}{\lambda^2}\mathbb{E}\|\sum_{j=0}^{k-1}\frac{\gamma_{j}}{\tilde{\gamma}}\mathbf{g}_{i,j}^{t}\|^2 \leq \frac{\tilde{\gamma}^2}{\lambda^2}\sum_{j=0}^{{k}-1}\frac{\gamma_{j}}{\tilde{\gamma}}\mathbb{E}\|\mathbf{g}_{i,j}^{t}\|^2 \leq \frac{\tilde{\gamma}^2}{\lambda^2}\sum_{j=0}^{{K}-1}\frac{\gamma_{j}}{\tilde{\gamma}}\mathbb{E}\|\mathbf{g}_{i,j}^{t}\|^2
$$

Since $\gamma_p = \eta\lambda\bigl(1-{\eta}{\lambda}\bigr)^{K-1-p}$ satisfies $\gamma_p>0$ under the condition $\eta\lambda<1$, it follows that for all $k \leq K$, $\tilde{\gamma} = \sum_{p=0}^{k-1}\gamma_p \leq  \sum_{p=0}^{K-1}\gamma_p = \gamma$. Therefore, we have

$$
\frac{\tilde{\gamma}^2}{\lambda^2}\sum_{j=0}^{K-1}\frac{\gamma_{j}}{\tilde{\gamma}}\mathbb{E}\|\mathbf{g}_{i,j}^{t}\|^2 = \frac{\tilde{\gamma}\gamma}{\lambda^2}\sum_{j=0}^{K-1}\frac{\gamma_{j}}{\gamma}\mathbb{E}\|\mathbf{g}_{i,j}^{t}\|^2 \leq \frac{\gamma^2}{\lambda^2}\sum_{j=0}^{K-1}\frac{\gamma_{j}}{\gamma}\mathbb{E}\|\mathbf{g}_{i,j}^{t}\|^2 
$$

Based on the above inequality, and by replacing the index $j$ with $k$, we obtain:

$$
\mathbb{E}\|\mathbf{x}_{i,k}^{t}-\mathbf{x}_{i,0}^{t}\|^2 \leq \frac{\gamma^2}{\lambda^2}\sum_{k=0}^{K-1}\frac{\gamma_{k}}{\gamma}\mathbb{E}\|\mathbf{g}_{i,k}^{t}\|^2 
$$
    \end{proof}
\end{lemma}

\begin{lemma}\label{le:average_updates}
    \textbf{(Average updates)} For the sequence $\mathbf{x}_{i,k}^{t}$ generated by Algorithm \ref{alg:DFedCata}, we define the mean sequence $\bar{\mathbf{x}}^t = \frac{1}{m}\sum_{i=1}^m\mathbf{x}_i^t$, where $\mathbf{x}_i^t  = \sum_{j=1}^m w_{i,j} \mathbf{x}_{j,K}^{t-1}$. Then the mean sequence $\bar{\mathbf{x}}^t$ satisfies:
    \begin{equation*}
        \bar{\mathbf{x}}^{t+1} = \bar{\mathbf{x}}^{t} + \beta (\bar{\mathbf{x}}^{t} - \bar{\mathbf{x}}^{t-1}) - \frac{\gamma}{\lambda}\sum_{k=0}^{K-1}\frac{\gamma_{k}}{\gamma}\left(\frac{1}{m}\sum_{i=1}^m\mathbf{g}_{i,k}^{t} \right)
    \end{equation*}
    \begin{proof}
        According to the Lemma \ref{le:local_update} and the communication step $\mathbf{x}_i^{t+1}  = \sum_{j=1}^m w_{i,j} \mathbf{x}_{j,K}^{t}$, we have the following equation:
        \begin{equation*}
        \begin{aligned}
            \mathbf{x}_i^{t+1}  &= \sum_{j=1}^m w_{i,j} \mathbf{x}_{j,K}^{t}\\
            &= \sum_{j=1}^m w_{i,j} \left( \mathbf{x}_{j,0}^{t} - \frac{\gamma}{\lambda}\sum_{k=0}^{K-1}\frac{\gamma_{k}}{\gamma}\mathbf{g}_{j,k}^{t} \right)\\
            &= \sum_{j=1}^m w_{i,j} \mathbf{x}_{j,0}^{t} - \sum_{j=1}^m w_{i,j} \left( \frac{\gamma}{\lambda}\sum_{k=0}^{K-1}\frac{\gamma_{k}}{\gamma}\mathbf{g}_{j,k}^{t} \right)\\
            &\overset{(a)}{=} \sum_{j=1}^m w_{i,j} \left(\mathbf{x}_{j}^{t} + \beta \left(\mathbf{x}_{j}^{t} - \mathbf{x}_{j}^{t-1} \right) \right)- \sum_{j=1}^m w_{i,j} \left( \frac{\gamma}{\lambda}\sum_{k=0}^{K-1}\frac{\gamma_{k}}{\gamma}\mathbf{g}_{j,k}^{t} \right)
        \end{aligned}
        \end{equation*}
        Where (a) uses the initialization condition in Algorithm \ref{alg:DFedCata} line 3. According to the Lemma \ref{le:lemma1} and the definition of mean sequence $\bar{\mathbf{x}}^{t} = \frac{1}{m}\sum_{i=1}^m\mathbf{x}_i^{t}$, we have the following equation:
        \begin{equation*}
            \begin{aligned}
                \bar{\mathbf{x}}^{t+1} = \frac{1}{m}\sum_{i=1}^m\mathbf{x}_i^{t+1} &= \frac{1}{m}\sum_{i=1}^m\sum_{j=1}^m w_{i,j} \left(\mathbf{x}_{j}^{t} + \beta \left(\mathbf{x}_{j}^{t} - \mathbf{x}_{j}^{t-1} \right) \right) - \frac{1}{m}\sum_{i=1}^m\sum_{j=1}^m w_{i,j} \left( \frac{\gamma}{\lambda}\sum_{k=0}^{K-1}\frac{\gamma_{k}}{\gamma}\mathbf{g}_{j,k}^{t} \right)\\
                &\overset{(a)}{=} \frac{1}{m}\sum_{i=1}^m\left(\mathbf{x}_{i}^{t} + \beta \left(\mathbf{x}_{i}^{t} - \mathbf{x}_{i}^{t-1} \right) \right) - \frac{1}{m}\sum_{i=1}^m \left( \frac{\gamma}{\lambda}\sum_{k=0}^{K-1}\frac{\gamma_{k}}{\gamma}\mathbf{g}_{i,k}^{t} \right)\\
                &= \bar{\mathbf{x}}^t + \beta (\bar{\mathbf{x}}^t - \bar{\mathbf{x}}^{t-1}) -   \frac{\gamma}{\lambda}\sum_{k=0}^{K-1}\frac{\gamma_{k}}{\gamma}\left(\frac{1}{m}\sum_{i=1}^m\mathbf{g}_{i,k}^{t} \right)
            \end{aligned}
        \end{equation*}
        We have thus completed the proof.
        
        % Let $\mathbf{z}^t = \bar{\mathbf{x}}^t + \frac{\beta}{1-\beta}\left(\bar{\mathbf{x}}^t - \bar{\mathbf{x}}^{t-1}\right)$, We can rewrite the above equation in the following form:
        % \begin{equation*}
        %     \begin{aligned}
        %         \mathbf{z}^{t+1} = \mathbf{z}^{t} - \frac{\gamma}{\lambda(1-\beta)}\sum_{k=0}^{K-1}\frac{\gamma_{k}}{\gamma}\left(\frac{1}{m}\sum_{i=1}^m\mathbf{g}_{i,k}^{t} \right)
        %     \end{aligned}
        % \end{equation*}
    \end{proof}
\end{lemma}

\begin{lemma}\label{le:bounded average update}
    \textbf{(Bounded average update)} The global update $\|\bar{\mathbf{x}}^t - \bar{\mathbf{x}}^{t-1}\|^2$ holds the upper bound of:
    \begin{equation*}
        \begin{aligned}
            \|\bar{\mathbf{x}}^t - \bar{\mathbf{x}}^{t-1}\|^2 \leq \frac{\|\bar{\mathbf{x}}^t - \bar{\mathbf{x}}^{t-1}\|^2 - \|\bar{\mathbf{x}}^{t+1} - \bar{\mathbf{x}}^{t}\|^2}{1-\beta} + \frac{1}{1-\beta}\frac{\gamma^2}{\lambda^2}\left\|\sum_{k=0}^{K-1}\frac{\gamma_{k}}{\gamma}\frac{1}{m}\sum_{i=1}^m\mathbf{g}_{i,k}^{t}\right\|^2
        \end{aligned}
    \end{equation*}
    \begin{proof}
        For the sake of brevity in the proof, we let $A^t = \bar{\mathbf{x}}^{t+1} - \bar{\mathbf{x}}^{t}, B^t = \sum_{k=0}^{K-1}\frac{\gamma_{k}}{\gamma}\frac{1}{m}\sum_{i=1}^m\mathbf{g}_{i,k}^{t}$, by Lemma \ref{le:average_updates} we have:
        \begin{equation*}
            \begin{aligned}
                A^{t}\leq \beta A^{t-1} - (1-\beta)\frac{\gamma}{\lambda}\frac{1}{1-\beta}B^t
            \end{aligned}
        \end{equation*}
        Take the L2-norm and we have:
        \begin{equation*}
            \begin{aligned}
                \|A^{t}\|^2 &\leq \left\|\beta A^{t-1} - (1-\beta)\frac{\gamma}{\lambda}\frac{1}{1-\beta}B^t\right\|^2\\
                &\overset{(a)}{\leq} \beta \|A^{t-1}\|^2 + \frac{1}{1-\beta}\frac{\gamma^2}{\lambda^2}\|B^t\|^2\\
                &= \left(1-(1-\beta)\right)\|A^{t-1}\|^2 + \frac{1}{1-\beta}\frac{\gamma^2}{\lambda^2}\|B^t\|^2
            \end{aligned}
        \end{equation*}
        Where (a) uses Jensen’s inequality and a simple transformation of the above equation yields:
        \begin{equation*}
            \begin{aligned}
                (1-\beta)\|A^{t-1}\|^2 &\leq \|A^{t-1}\|^2 - \|A^{t}\|^2 + \frac{1}{1-\beta}\frac{\gamma^2}{\lambda^2}\|B^t\|^2
            \end{aligned}
        \end{equation*}
        Dividing both sides of the inequality by $1-\beta$, we obtain:
        \begin{equation*}
            \|A^{t-1}\|^2 \leq \frac{\|A^{t-1}\|^2 - \|A^{t}\|^2}{1-\beta} + \frac{1}{(1-\beta)^2}\frac{\gamma^2}{\lambda^2}\|B^t\|^2
        \end{equation*}
        We have completed the proof by substituting the specific expressions of $A^t$ and $B^t$.
    \end{proof}
\end{lemma}

\begin{lemma}\label{le:auxiliary updates}
    \textbf{(Auxiliary sequence updates)} Considering the auxiliary sequence defined by the mean sequence defined in Lemma \ref{le:average_updates} as $\left\{\mathbf{z}^t = \bar{\mathbf{x}}^t + \frac{\beta}{1-\beta}\left(\bar{\mathbf{x}}^t - \bar{\mathbf{x}}^{t-1}\right)\right\}_{t>0}$, this sequence satisfies the following equation:
    \begin{equation*}
    \begin{aligned}
        \mathbf{z}^{t+1} = \mathbf{z}^{t} - \frac{\gamma}{\lambda(1-\beta)}\sum_{k=0}^{K-1}\frac{\gamma_{k}}{\gamma}\left(\frac{1}{m}\sum_{i=1}^m\mathbf{g}_{i,k}^{t} \right)
    \end{aligned}
    \end{equation*}
    \begin{proof}
        It is easy to verify the conclusion of Lemma \ref{le:average_updates}:
        \begin{equation*}
            \begin{aligned}
                \mathbf{z}^{t+1} - \mathbf{z}^{t} &= \left(1+\frac{\beta}{1-\beta}\right)\left(\bar{\mathbf{x}}^{t+1} - \bar{\mathbf{x}}^{t}\right) - \frac{\beta}{1-\beta}\left(\bar{\mathbf{x}}^t - \bar{\mathbf{x}}^{t-1}\right)\\
                &= \frac{1}{1-\beta}\left(\beta(\bar{\mathbf{x}}^t - \bar{\mathbf{x}}^{t-1}) - \frac{\gamma}{\lambda}\sum_{k=0}^{K-1}\frac{\gamma_{k}}{\gamma}\left(\frac{1}{m}\sum_{i=1}^m\mathbf{g}_{i,k}^{t} \right) \right) - \frac{\beta}{1-\beta}\left(\bar{\mathbf{x}}^t - \bar{\mathbf{x}}^{t-1}\right)\\
                &= - \frac{\gamma}{\lambda(1-\beta)}\sum_{k=0}^{K-1}\frac{\gamma_{k}}{\gamma}\left(\frac{1}{m}\sum_{i=1}^m\mathbf{g}_{i,k}^{t} \right)
            \end{aligned}
        \end{equation*}
        Thus we complete the proof.
    \end{proof}
\end{lemma}

\begin{lemma}\label{le:bounded gradient}
    \textbf{(Bounded gradient)} Let $\mathbf{g}_{i,k}^{t}$ denote the stochastic gradient at $\{\mathbf{x}_{i,k}^t\}$ generated by Algorithm \ref{alg:DFedCata}, then it satisfies the following equation:
    \begin{equation*}
        \begin{aligned}
            \left\| \sum_{k=0}^{K-1}\frac{\gamma_{k}}{\gamma}\left(\frac{1}{m}\sum_{i=1}^m\mathbf{g}_{i,k}^{t} \right) \right\|^2 \leq \frac{2\sigma^2}{m}\sum_{k=0}^{K-1}\left(\frac{\gamma_{k}}{\gamma}\right)^2 + 2\left\|\sum_{k=0}^{K-1}\frac{\gamma_{k}}{\gamma}\left(\frac{1}{m}\sum_{i=1}^m \nabla f_i(\mathbf{x}_{i,k}^t)\right) \right\|^2
        \end{aligned}
    \end{equation*}
    where $\gamma_{k}$ and ${\gamma}$ are defined in Lemma \ref{le:local_update}.
    \begin{proof}
        According to the Jensen's inequality and Assumption \ref{as:bounded_stochastic_gradient}, we can obtain:
        \begin{equation*}
            \begin{aligned}
                \left\| \sum_{k=0}^{K-1}\frac{\gamma_{k}}{\gamma}\left(\frac{1}{m}\sum_{i=1}^m\mathbf{g}_{i,k}^{t} \right) \right\|^2
                &\leq \left\|\sum_{k=0}^{K-1}\frac{\gamma_{k}}{\gamma}\left(\frac{1}{m}\sum_{i=1}^m(\mathbf{g}_{i,k}^{t} \pm \nabla f_i(\mathbf{x}_{i,k}^t)) \right)\right\|^2\\
                &\leq \frac{2\sigma^2}{m}\sum_{k=0}^{K-1}\left(\frac{\gamma_{k}}{\gamma}\right)^2 + 2\left\|\sum_{k=0}^{K-1}\frac{\gamma_{k}}{\gamma}\left(\frac{1}{m}\sum_{i=1}^m \nabla f_i(\mathbf{x}_{i,k}^t)\right) \right\|^2\\
            \end{aligned}
        \end{equation*}
    \end{proof}
\end{lemma}

\begin{lemma}\label{le:bounded error}
    \textbf{(Bounded error)} Let $\left\{\mathbf{z}^t = \bar{\mathbf{x}}^t + \frac{\beta}{1-\beta}\left(\bar{\mathbf{x}}^t - \bar{\mathbf{x}}^{t-1}\right)\right\}_{t>0}$, where $ \bar{\mathbf{x}}^t$ represents the mean sequence. The difference between them $\mathbf{z}^t - \bar{\mathbf{x}}^t$ can be bounded by the following equation:
    \begin{equation*}
        \begin{aligned}
            \|\mathbf{z}^t - \bar{\mathbf{x}}^t\|^2 \leq  \frac{\beta^2(\|A^{t-1}\|^2 - \|A^{t}\|^2)}{(1-\beta)^3} + \frac{\beta^2}{(1-\beta)^4}\frac{\gamma^2}{\lambda^2}\|B^t\|^2
        \end{aligned}
    \end{equation*}
    Where $A^t = \bar{\mathbf{x}}^{t+1} - \bar{\mathbf{x}}^{t}, B^t = \sum_{k=0}^{K-1}\frac{\gamma_{k}}{\gamma}\frac{1}{m}\sum_{i=1}^m\mathbf{g}_{i,k}^{t}$.
    \begin{proof}
        By rearranging the expression for $\mathbf{z}^t$ and taking the L2-norm on both sides, we obtain:
        \begin{equation*}
            \begin{aligned}
                \|\mathbf{z}^t - \bar{\mathbf{x}}^t\|^2 = \frac{\beta^2}{(1-\beta)^2}\|A^{t-1}\|^2
            \end{aligned}
        \end{equation*}
        By directly substituting the conclusion of Lemma  \ref{le:bounded average update} into the above equation, we get
        \begin{equation*}
            \|\mathbf{z}^t - \bar{\mathbf{x}}^t\|^2 \leq  \frac{\beta^2(\|A^{t-1}\|^2 - \|A^{t}\|^2)}{(1-\beta)^3} + \frac{\beta^2}{(1-\beta)^4}\frac{\gamma^2}{\lambda^2}\|B^t\|^2
        \end{equation*}
        Then we have completed the proof.
    \end{proof}
\end{lemma}

\subsubsection{Expanding the Smoothness Inequality for the Non-convex Objective}
For the non-convex and $L$-smooth function $f$ , we firstly expand the smoothness inequality at round $t$ as:
\allowdisplaybreaks
\begin{align}\label{eq:l_smooth_main}
    &\mathbb{E}[f(\mathbf{z}^{t+1})-f(\mathbf{z}^{t})] \\ \notag
    &\leq \mathbb{E}\langle\nabla f(\mathbf{z}^{t}),\mathbf{z}^{t+1}-\mathbf{z}^t\rangle+\frac L2\mathbb{E}\|\mathbf{z}^{t+1}-\mathbf{z}^{t}\|^2 \\ \notag
    & \overset{(a)}\leq -\frac{\gamma}{\lambda(1-\beta)}\mathbb{E}\left\langle\nabla f(\mathbf{z}^{t}),\sum_{k=0}^{K-1}\frac{\gamma_{k}}{\gamma}\left(\frac{1}{m}\sum_{i=1}^m \nabla f_i(\mathbf{x}_{i,k}^{t}) \right)\right\rangle+\frac L2 \frac{\gamma^2}{\lambda^2(1-\beta)^2}\mathbb{E}\left\|\sum_{k=0}^{K-1}\frac{\gamma_{k}}{\gamma}\left(\frac{1}{m}\sum_{i=1}^m\mathbf{g}_{i,k}^{t} \right)\right\|^2 \\ \notag 
    &\overset{(b)}\leq -\frac{\gamma}{\lambda(1-\beta)}\mathbb{E}\left\langle\nabla f(\mathbf{z}^{t}),\sum_{k=0}^{K-1}\frac{\gamma_{k}}{\gamma}\left(\frac{1}{m}\sum_{i=1}^m \nabla f_i(\mathbf{x}_{i,k}^{t}) \right)\right\rangle \\ \notag
    &\quad +  \frac{L\sigma^2\gamma^2}{\lambda^2(1-\beta)^2m} +  \frac{L\gamma^2}{\lambda^2(1-\beta)^2}\mathbb{E}\left\|\sum_{k=0}^{K-1}\frac{\gamma_{k}}{\gamma}\left(\frac{1}{m}\sum_{i=1}^m \nabla f_i(\mathbf{x}_{i,k}^{t}) \right)\right\|^2 \\ \notag
    &\overset{(c)}= \frac{\gamma}{2\lambda(1-\beta)}\mathbb{E}\left\|\sum_{k=0}^{K-1}\frac{\gamma_{k}}{\gamma}\left(\frac{1}{m}\sum_{i=1}^m (\nabla f_i(\mathbf{x}_{i,k}^{t})  - \nabla f_i(\mathbf{z}^{t}) )\right)\right\|^2 -\frac{\gamma}{2\lambda(1-\beta)}\mathbb{E}\left\|\sum_{k=0}^{K-1}\frac{\gamma_{k}}{\gamma}\left(\frac{1}{m}\sum_{i=1}^m \nabla f_i(\mathbf{x}_{i,k}^{t}) \right)\right\|^2 \\ \notag
    &\quad -\frac{\gamma}{2\lambda(1-\beta)}\mathbb{E}\|\nabla f(\mathbf{z}^{t})\|^2 +  \frac{L\sigma^2\gamma^2}{\lambda^2(1-\beta)^2m} +  \frac{L\gamma^2}{\lambda^2(1-\beta)^2}\mathbb{E}\left\|\sum_{k=0}^{K-1}\frac{\gamma_{k}}{\gamma}\left(\frac{1}{m}\sum_{i=1}^m \nabla f_i(\mathbf{x}_{i,k}^{t}) \right)\right\|^2 \\ \notag 
    &\overset{(d)} \leq  \frac{\gamma}{2\lambda(1-\beta)}\mathbb{E}\left\|\sum_{k=0}^{K-1}\frac{\gamma_{k}}{\gamma}\left(\frac{1}{m}\sum_{i=1}^m (\nabla f_i(\mathbf{x}_{i,k}^{t}) -\nabla f_i(\mathbf{z}^{t})) \right)\right\|^2 -\frac{\gamma}{2\lambda(1- \beta)}\mathbb{E}\|\nabla f(\mathbf{z}^{t})\|^2 \\ \notag
    &\quad  +  \frac{L\sigma^2\gamma^2}{\lambda^2(1-\beta)^2m} +  \left(\frac{L\gamma^2}{\lambda^2(1-\beta)^2} -\frac{\gamma}{2\lambda(1-\beta)} \right)\mathbb{E}\left\|\sum_{k=0}^{K-1}\frac{\gamma_{k}}{\gamma}\left(\frac{1}{m}\sum_{i=1}^m \nabla f_i(\mathbf{x}_{i,k}^{t}) \right)\right\|^2 \\ \notag
    &\overset{(e)}\leq \frac{\gamma L^2}{2\lambda(1-\beta)}\mathbb{E}\sum_{k=0}^{K-1}\frac{\gamma_{k}}{\gamma}\left(\frac{1}{m}\sum_{i=1}^m \left\|\mathbf{x}_{i,k}^{t} -\bar{\mathbf{x}}^t + \bar{\mathbf{x}}^t -\mathbf{z}^{t} \right\|^2 \right)-\frac{\gamma}{2\lambda(1-\beta)}\mathbb{E}\|\nabla f(\mathbf{z}^{t})\|^2 \\ \notag
    &\quad  +  \frac{L\sigma^2\gamma^2}{\lambda^2(1-\beta)^2m} +  \left(\frac{L\gamma^2}{\lambda^2(1-\beta)^2} -\frac{\gamma}{2\lambda(1-\beta)} \right)\mathbb{E}\left\|\sum_{k=0}^{K-1}\frac{\gamma_{k}}{\gamma}\left(\frac{1}{m}\sum_{i=1}^m \nabla f_i(\mathbf{x}_{i,k}^{t}) \right)\right\|^2 \\ \notag
    &\overset{(f)}\leq \frac{\gamma L^2}{\lambda(1-\beta)}\underbrace{\mathbb{E}\sum_{k=0}^{K-1}\frac{\gamma_{k}}{\gamma}\left(\frac{1}{m}\sum_{i=1}^m \left\|\mathbf{x}_{i,k}^{t} -\bar{\mathbf{x}}^t \right\|^2 \right)}_{\mathbf{c}^t} + \frac{\gamma L^2}{\lambda(1-\beta)}\mathbb{E}\sum_{k=0}^{K-1}\frac{\gamma_{k}}{\gamma}\left(\frac{1}{m}\sum_{i=1}^m \left\| \bar{\mathbf{x}}^t -\mathbf{z}^{t} \right\|^2 \right)  \\ \notag
    &\quad -\frac{\gamma}{2\lambda(1-\beta)}\mathbb{E}\|\nabla f(\mathbf{z}^{t})\|^2 +  \frac{L\sigma^2\gamma^2}{\lambda^2(1-\beta)^2m} +  \left(\frac{L\gamma^2}{\lambda^2(1-\beta)^2} -\frac{\gamma}{2\lambda(1-\beta)} \right)\mathbb{E}\left\|\sum_{k=0}^{K-1}\frac{\gamma_{k}}{\gamma}\left(\frac{1}{m}\sum_{i=1}^m \nabla f_i(\mathbf{x}_{i,k}^{t}) \right)\right\|^2 \\ \notag
    &\overset{(g)}\leq \frac{\gamma L^2}{\lambda(1-\beta)}\mathbf{c}^t + \frac{\gamma L^2}{\lambda(1-\beta)}\mathbb{E}\sum_{k=0}^{K-1}\frac{\gamma_{k}}{\gamma}\left(\frac{1}{m}\sum_{i=1}^m \left\| \bar{\mathbf{x}}^t -\mathbf{z}^{t} \right\|^2 \right)  -\frac{\gamma}{2\lambda(1-\beta)}\mathbb{E}\|\nabla f(\mathbf{z}^{t})\|^2 \\ \notag
    &\quad +  \frac{L\sigma^2\gamma^2}{\lambda^2(1-\beta)^2m} +  \left(\frac{L\gamma^2}{\lambda^2(1-\beta)^2} -\frac{\gamma}{2\lambda(1-\beta)} \right)\mathbb{E}\left\|B\right\|^2 \\
\end{align}
Where (a) uses Lemma \ref{le:auxiliary updates}, (b) uses Lemma \ref{le:bounded gradient}, (c) uses $- \langle a,b \rangle = \frac{1}{2}\left(\|a-b\|^2 -\|a\|^2 - \|b\|^2 \right)$, (d) uses Assumtion \ref{as:bounded_heterogeneity} and Jensen’s inequality, (e) uses Jensen’s inequality and Assumption \ref{as:smoothness}, (f) uses $\|\mathbf{x}+\mathbf{y}\|^2\leq(1+a)\|\mathbf{x}\|^2+(1+\frac{1}{a})\|\mathbf{y}\|^2$, (g) uses the definition of $\mathbf{c}^t = \frac{1}{m}\sum_{i=1}^m\sum_{k=0}^{K-1}\frac{\gamma_{k}}{\gamma}\mathbb{E}\|\mathbf{x}_{i,k}^{t} -\bar{\mathbf{x}}^t \|^2$ and $B^t = \sum_{k=0}^{K-1}\frac{\gamma_{k}}{\gamma}\left(\frac{1}{m}\sum_{i=1}^m \nabla f_i(\mathbf{x}_{i,k}^{t}) \right)$.

\textbf{(a) Estimating the upper bound of $\mathbf{c}^t$.} We denote $\mathbf{c}^t = \frac{1}{m}\sum_{i=1}^m\sum_{k=0}^{K-1}\frac{\gamma_{k}}{\gamma}\mathbb{E}\|\mathbf{x}_{i,k}^{t} -\bar{\mathbf{x}}^t \|^2$ term as the local offset after $k$ iterations updates, we firstly consider the $\mathbf{c}_k^t = \frac{1}{m}\sum_{i=1}^m \mathbb{E}\|\mathbf{x}_{i,k}^{t} -\bar{\mathbf{x}}^t \|^2$ and it can be bounded as:
\begin{equation*}
    \begin{aligned}
        \mathbf{c}_k^t &= \frac{1}{m}\sum_{i=1}^m \mathbb{E}\|\mathbf{x}_{i,k}^{t} -\bar{\mathbf{x}}^t \|^2  = \frac{1}{m}\sum_{i=1}^m \mathbb{E}\|\mathbf{x}_{i,k}^{t} - \mathbf{x}_{i,0}^t + \mathbf{x}_{i,0}^t - \bar{\mathbf{x}}^t \|^2\\
        &= \frac{1}{m}\sum_{i=1}^m \mathbb{E}\|\mathbf{x}_{i,k}^{t} - \mathbf{x}_{i,0}^t + \beta(\bar{\mathbf{x}}^t - \bar{\mathbf{x}}^{t-1}) \|^2 \\
        & \overset{(a)}\leq \frac{1}{1-\beta}\frac{1}{m}\sum_{i=1}^m \mathbb{E}\|\mathbf{x}_{i,k}^{t} - \mathbf{x}_{i,0}^t\|^2 + \beta \frac{1}{m}\sum_{i=1}^m \mathbb{E}\| \bar{\mathbf{x}}^t - \bar{\mathbf{x}}^{t-1} \|^2\\
        & \overset{(b)}\leq \frac{1}{1-\beta}\frac{\gamma^2}{\lambda^2}\underbrace{\frac{1}{m}\sum_{i=1}^m \sum_{k=0}^{K-1}\frac{\gamma_{k}}{\gamma}\mathbb{E}\|\mathbf{g}_{i,k}^{t}\|^2}_{G^t} + \frac{\beta }{1-\beta}\left(\mathbb{E}\|A^{t-1}\|^2 - \mathbb{E}\|A^{t}\|^2\right) + \frac{\beta}{(1-\beta)^2}\frac{\gamma^2}{\lambda^2}\mathbb{E}\|B^t\|^2\\
    \end{aligned}
\end{equation*}
Where (a) uses Jensen’s inequality, (b) uses Lemma \ref{le:bounded average update} and \ref{le:bound local update}.

\textbf{(a.1) Estimating the upper bound of ${G}^t$.} Next, to obtain $\mathbf{c}^t$ further, we will need to first estimate the upper bound of $G^t$.
\begin{align*}
        G^t &= \frac{1}{m}\sum_{i=1}^m \sum_{k=0}^{K-1}\frac{\gamma_{k}}{\gamma}\mathbb{E}\|\mathbf{g}_{i,k}^{t}\|^2 \\
        & = \frac{1}{m}\sum_{i=1}^m \sum_{k=0}^{K-1}\frac{\gamma_{k}}{\gamma}\mathbb{E}\|\mathbf{g}_{i,k}^{t} - \nabla f_i(\mathbf{x}_{i,k}^t) + \nabla f_i(\mathbf{x}_{i,k}^t)\|^2 \\
        &\overset{(a)}\leq \sigma^2 + \frac{1}{m}\sum_{i=1}^m \sum_{k=0}^{K-1}\frac{\gamma_{k}}{\gamma}\mathbb{E}\| \nabla f_i(\mathbf{x}_{i,k}^t) - \nabla f_i(\mathbf{z}^t) + \nabla f_i(\mathbf{z}^t)\|^2 \\
        &\overset{(b)}\leq \sigma^2 + \frac{2L^2}{m}\sum_{i=1}^m \sum_{k=0}^{K-1}\frac{\gamma_{k}}{\gamma}\mathbb{E}\| \mathbf{x}_{i,k}^t - \mathbf{z}^t\|^2 + 2G^2 + 2B^2\mathbb{E}\|\nabla f(\mathbf{z}^t)\|^2\\
        & =\sigma^2 + \frac{2L^2}{m}\sum_{i=1}^m \sum_{k=0}^{K-1}\frac{\gamma_{k}}{\gamma}\mathbb{E}\| \mathbf{x}_{i,k}^t - \bar{\mathbf{x}}^t + \bar{\mathbf{x}}^t - \mathbf{z}^t\|^2 + 2G^2 + 2B^2\mathbb{E}\|\nabla f(\mathbf{z}^t)\|^2\\
        &\leq \sigma^2 + \frac{4L^2}{m}\sum_{i=1}^m \sum_{k=0}^{K-1}\frac{\gamma_{k}}{\gamma}\mathbb{E}\| \mathbf{x}_{i,k}^t - \bar{\mathbf{x}}^t\|^2  + \frac{4L^2}{m}\sum_{i=1}^m \sum_{k=0}^{K-1}\frac{\gamma_{k}}{\gamma}\mathbb{E}\|\bar{\mathbf{x}}^t - \mathbf{z}^t\|^2 + 2G^2 + 2B^2\mathbb{E}\|\nabla f(\mathbf{z}^t)\|^2\\
        &= \sigma^2 + 4L^2\mathbf{c}^t  + \frac{4L^2}{m}\sum_{i=1}^m \sum_{k=0}^{K-1}\frac{\gamma_{k}}{\gamma}\mathbb{E}\|\bar{\mathbf{x}}^t - \mathbf{z}^t\|^2 + 2G^2 + 2B^2\mathbb{E}\|\nabla f(\mathbf{z}^t)\|^2\\
\end{align*}
Where (a) uses Assumption \ref{as:bounded_stochastic_gradient}, (b) uses Jensen’s inequality and Assumption \ref{as:bounded_heterogeneity} \& \ref{as:smoothness}.

Substituting the upper bound of $G^t$ into $\mathbf{c}_k^t$, we obtain:
\begin{equation*}
    \begin{aligned}
        \mathbf{c}_k^t &\leq \frac{1}{1-\beta}\frac{\gamma^2}{\lambda^2}\sigma^2 + \frac{4L^2}{1-\beta}\frac{\gamma^2}{\lambda^2}\mathbf{c}^t  + \frac{4L^2}{1-\beta}\frac{\gamma^2}{\lambda^2}\frac{1}{m}\sum_{i=1}^m \sum_{k=0}^{K-1}\frac{\gamma_{k}}{\gamma}\mathbb{E}\|\bar{\mathbf{x}}^t - \mathbf{z}^t\|^2 + \frac{2}{1-\beta}\frac{\gamma^2}{\lambda^2}G^2 \\
        &\quad + \frac{2B^2}{1-\beta}\frac{\gamma^2}{\lambda^2}\mathbb{E}\|\nabla f(\mathbf{z}^t)\|^2  + \frac{\beta }{1-\beta}\left(\mathbb{E}\|A^{t-1}\|^2 - \mathbb{E}\|A^{t}\|^2\right) + \frac{\beta}{(1-\beta)^2}\frac{\gamma^2}{\lambda^2}\mathbb{E}\|B^t\|^2
    \end{aligned}
\end{equation*}

Using the relationship $\mathbf{c}^t = \sum_{k=0}^{K-1}\frac{\gamma_k}{\gamma}\mathbf{c}_k^t$, we can estimate the upper bound of $\mathbf{c}^t$.
\begin{equation*}
\begin{aligned}
    \mathbf{c}^t & = \sum_{k=0}^{K-1}\frac{\gamma_k}{\gamma}\mathbf{c}_k^t \leq \frac{1}{1-\beta}\frac{\gamma^2}{\lambda^2}\sigma^2 + \frac{4L^2}{1-\beta}\frac{\gamma^2}{\lambda^2}\mathbf{c}^t  + \frac{4L^2}{1-\beta}\frac{\gamma^2}{\lambda^2}\frac{1}{m}\sum_{i=1}^m \sum_{k=0}^{K-1}\frac{\gamma_{k}}{\gamma}\mathbb{E}\|\bar{\mathbf{x}}^t - \mathbf{z}^t\|^2 + \frac{2}{1-\beta}\frac{\gamma^2}{\lambda^2}G^2 \\
    &\quad + \frac{2B^2}{1-\beta}\frac{\gamma^2}{\lambda^2}\mathbb{E}\|\nabla f(\mathbf{z}^t)\|^2
     + \frac{\beta }{1-\beta}\left(\mathbb{E}\|A^{t-1}\|^2 - \mathbb{E}\|A^{t}\|^2\right) + \frac{\beta}{(1-\beta)^2}\frac{\gamma^2}{\lambda^2}\mathbb{E}\|B^t\|^2
\end{aligned}
\end{equation*}

We set $\mu = 1 - \frac{4L^2}{1-\beta}\frac{\gamma^2}{\lambda^2}$, and since $\gamma = 1-(1-\lambda \eta)^K \leq K\eta \lambda$, when $\eta < \frac{\sqrt{1-\beta}}{2KL}$, we can ensure that $\mu >0$. Thus we get:
\begin{equation*}
    \begin{aligned}
        \mathbf{c}^t &\leq \frac{1}{1-\beta}\frac{\gamma^2}{\mu\lambda^2}\sigma^2 + \frac{4L^2}{1-\beta}\frac{\gamma^2}{\mu\lambda^2}\frac{1}{m}\sum_{i=1}^m \sum_{k=0}^{K-1}\frac{\gamma_{k}}{\gamma}\mathbb{E}\|\bar{\mathbf{x}}^t - \mathbf{z}^t\|^2 + \frac{2}{1-\beta}\frac{\gamma^2}{\mu\lambda^2}G^2 \\
        &\quad + \frac{2B^2}{1-\beta}\frac{\gamma^2}{\mu\lambda^2}\mathbb{E}\|\nabla f(\mathbf{z}^t)\|^2
        + \frac{\beta }{(1-\beta)\mu}\left(\mathbb{E}\|A^{t-1}\|^2 - \mathbb{E}\|A^{t}\|^2\right) + \frac{\beta}{(1-\beta)^2}\frac{\gamma^2}{\mu\lambda^2}\mathbb{E}\|B^t\|^2
    \end{aligned}
\end{equation*}

Substituting the upper bound of $\mathbf{c}^t$ into inequality \ref{eq:l_smooth_main}, we obtain:
\begin{equation*}
    \begin{aligned}
        &\mathbb{E}[f(\mathbf{z}^{t+1})-f(\mathbf{z}^{t})] \\
        &\leq \left(\frac{\gamma L^2}{\lambda(1-\beta)} + \frac{4\gamma^3 L^3}{\lambda^3(1-\beta)^2\mu}\right)\mathbb{E}\sum_{k=0}^{K-1}\frac{\gamma_{k}}{\gamma}\left(\frac{1}{m}\sum_{i=1}^m \left\| \bar{\mathbf{x}}^t -\mathbf{z}^{t} \right\|^2 \right) + \frac{2\gamma^3 L^2}{\lambda^3(1-\beta)^2\mu}G^2\\
        &\quad + \left(\frac{2B^2\gamma^3 L^2}{\lambda^3(1-\beta)^2\mu} -\frac{\gamma}{2\lambda(1-\beta)}\right)\mathbb{E}\|\nabla f(\mathbf{z}^{t})\|^2  +  \left(\frac{\gamma^3 L^2 m }{\lambda^3(1-\beta)^2\mu}+ \frac{L\gamma^2}{\lambda^2(1-\beta)^2}\right)\frac{\sigma^2}{m}  \\
        &\quad +  \left( \frac{\gamma^3 L^2 \beta^2}{\lambda^3(1-\beta)^3\mu} + \frac{L\gamma^2}{\lambda^2(1-\beta)^2} -\frac{\gamma}{2\lambda(1-\beta)} \right)\mathbb{E}\left\|B\right\|^2  + \frac{\gamma L^2 \beta}{\lambda(1-\beta)^2\mu}\left(\mathbb{E}\|A^{t-1}\|^2 - \mathbb{E}\|A^{t}\|^2\right)\\
        &\overset{(a)}\leq \frac{2\gamma^3 L^2}{\lambda^3(1-\beta)^2\mu}G^2 + \left(\frac{2B^2\gamma^3 L^2}{\lambda^3(1-\beta)^2\mu} -\frac{\gamma}{2\lambda(1-\beta)}\right)\mathbb{E}\|\nabla f(\mathbf{z}^{t})\|^2  +  \left(\frac{\gamma^3 L^2 m }{\lambda^3(1-\beta)^2\mu}+ \frac{L\gamma^2}{\lambda^2(1-\beta)^2}\right)\frac{\sigma^2}{m}  \\
        &\quad +  \left(\frac{\gamma^3L^2\beta^2}{\lambda^3(1-\beta)^5} + \frac{4\gamma^5L^3\beta^2}{\lambda^5(1-\beta)^6\mu} +\frac{\gamma^3 L^2 \beta^2}{\lambda^3(1-\beta)^3\mu} + \frac{L\gamma^2}{\lambda^2(1-\beta)^2} -\frac{\gamma}{2\lambda(1-\beta)} \right)\mathbb{E}\left\|B\right\|^2 \\
        &\quad + \left( \frac{\gamma L^2 \beta}{\lambda(1-\beta)^2\mu} + \frac{\gamma L^2 \beta^2}{\lambda (1-\beta)^4} + \frac{4\gamma^3 L^3 \beta^2}{\lambda^3 (1-\beta)^5 \mu} \right)\left(\mathbb{E}\|A^{t-1}\|^2 - \mathbb{E}\|A^{t}\|^2\right)\\
    \end{aligned}
\end{equation*}
Where (a) uses Lemma \ref{le:bounded error} and let $\tilde{\eta} = \frac{\gamma}{\lambda(1-\beta)}$, which simplifies the above equation to:
\begin{equation}\label{eq:main_2}
    \begin{aligned}
        &\mathbb{E}[f(\mathbf{z}^{t+1})-f(\mathbf{z}^{t})] \\
        &\leq \frac{2\tilde{\eta}^3L^2(1-\beta)}{\mu}G^2 -\frac{\tilde{\eta}}{2} \left(1 - \frac{2\tilde{\eta}^2B^2 L^2(1-\beta)}{\mu} \right)\mathbb{E}\|\nabla f(\mathbf{z}^{t})\|^2  +  \tilde{\eta}^2\left(\frac{ L^2 m (1-\beta)}{\mu} + L\right)\frac{\sigma^2}{m}  \\
        &\quad - \frac{\tilde{\eta}}{2}\left(1- \frac{L^2\beta^2 }{(1-\beta)^2}\tilde{\eta}^2 - \frac{4L^3\beta^2}{(1-\beta)\mu}\tilde{\eta}^4 -\frac{ L^2 \beta^2}{\mu}\tilde{\eta}^2 - L\tilde{\eta} \right)\mathbb{E}\left\|B\right\|^2 \\
        &\quad + \tilde{\eta}\left( \frac{ L^2 \beta}{(1-\beta)\mu} + \frac{ L^2 \beta^2}{ (1-\beta)^3} + \frac{4 L^3 \beta^2}{ (1-\beta)^2 \mu}\tilde{\eta}^2 \right)\left(\mathbb{E}\|A^{t-1}\|^2 - \mathbb{E}\|A^{t}\|^2\right)\\
    \end{aligned}
\end{equation}

Next, we will discuss $\tilde{\eta} = \frac{\gamma}{\lambda(1-\beta)}$. Firstly, because $\gamma = 1-(1-\eta\lambda)^K < K\eta\lambda $ always holds, we have $\tilde{\eta} \leq \frac{K \eta}{1-\beta}$. Therefore, by controlling the learning rate $\eta$, we can control the magnitude of $\tilde{\eta}$. We define $\kappa = 1 - \frac{2\tilde{\eta}^2B^2 L^2(1-\beta)}{\mu}, \mu = 1 - \frac{4L^2}{1-\beta}\frac{\gamma^2}{\lambda^2}$. When setting the learning rate $\eta < \frac{\sqrt{1-\beta}}{2KL\sqrt{(1+B^2)}}$, we can ensure that $\kappa > \frac{1}{2}$. Then, by setting the learning rate to $\eta < \frac{(1-\beta)^{\frac{3}{2}}}{2KL\sqrt{2L}}$, we can ensure that $ \frac{4L^3\beta^2}{(1-\beta)\mu}\tilde{\eta}^4 + \frac{ L^2 \beta^2}{\mu}\tilde{\eta}^2 < \frac12$ holds. When the learning rate satisfies $\eta < \frac{(1-\beta)^2}{KL}$, we can ensure that $\frac{L^2\beta^2 }{(1-\beta)^2}\tilde{\eta}^2+L\tilde{\eta} < \frac12$ holds. Therefore, when $\eta < \min\{\frac{(1-\beta)^{\frac{3}{2}}}{2KL\sqrt{2L}}, \frac{(1-\beta)^2}{KL}\}$, we have $1- \frac{L^2\beta^2 }{(1-\beta)^2}\tilde{\eta}^2 - \frac{4L^3\beta^2}{(1-\beta)\mu}\tilde{\eta}^4 -\frac{ L^2 \beta^2}{\mu}\tilde{\eta}^2 - L\tilde{\eta} > 0$.

In conclusion, when the learning rate satisfies $\eta < \min\{\frac{(1-\beta)^{\frac{3}{2}}}{2KL\sqrt{2L}}, \frac{(1-\beta)^2}{KL}, \frac{\sqrt{1-\beta}}{2KL\sqrt{(1+B^2)}}\}$, inequality (\ref{eq:main_2}) can be simplified to the following:

\begin{equation}\label{eq:main_3}
    \begin{aligned}
        \mathbb{E}[f(\mathbf{z}^{t+1})-f(\mathbf{z}^{t})] 
        &\leq \frac{2\tilde{\eta}^3L^2(1-\beta)}{\mu}G^2 -\frac{\tilde{\eta}}{2} \kappa \mathbb{E}\|\nabla f(\mathbf{z}^{t})\|^2  +  \tilde{\eta}^2\left(\frac{ L^2 m (1-\beta)}{\mu} + L\right)\frac{\sigma^2}{m}  \\
        &\quad + \tilde{\eta}\left( \frac{ L^2 \beta}{(1-\beta)\mu} + \frac{ L^2 \beta^2}{ (1-\beta)^3} + \frac{4 L^3 \beta^2}{ (1-\beta)^2 \mu}\tilde{\eta}^2 \right)\left(\mathbb{E}\|A^{t-1}\|^2 - \mathbb{E}\|A^{t}\|^2\right)\\
    \end{aligned}
\end{equation}

Simple rearrangement of inequality (\ref{eq:main_3}) followed by division of both sides by $\frac{\tilde{\eta}}{2}\kappa$ yields:
\begin{equation*}
    \begin{aligned}
        \mathbb{E}\|\nabla f(\mathbf{z}^{t})\|^2 
        &\leq \frac{2\mathbb{E}[f(\mathbf{z}^{t})-f(\mathbf{z}^{t+1})]}{\tilde{\eta}\kappa} + \frac{4\tilde{\eta}^2L^2(1-\beta)}{\mu\kappa}G^2 +  \frac{2\tilde{\eta}}{\kappa}\left(\frac{ L^2 m (1-\beta)}{\mu} + L\right)\frac{\sigma^2}{m}  \\
        &\quad + \frac{2}{\kappa}\left( \frac{ L^2 \beta}{(1-\beta)\mu} + \frac{ L^2 \beta^2}{ (1-\beta)^3} + \frac{4 L^3 \beta^2}{ (1-\beta)^2 \mu}\tilde{\eta}^2 \right)\left(\mathbb{E}\|A^{t-1}\|^2 - \mathbb{E}\|A^{t}\|^2\right)\\
        % &\overset{(a)}\leq \frac{4\mathbb{E}[f(\mathbf{z}^{t})-f(\mathbf{z}^{t+1})]}{\tilde{\eta}} + \frac{8\tilde{\eta}^2L^2(1-\beta)}{\mu}G^2 + 4\tilde{\eta}\left(\frac{ L^2 m (1-\beta)}{\mu} + L\right)\frac{\sigma^2}{m}  \\
        % &\quad + 4\left( \frac{ L^2 \beta}{(1-\beta)\mu} + \frac{ L^2 \beta^2}{ (1-\beta)^3} + \frac{4 L^3 \beta^2}{ (1-\beta)^2 \mu}\tilde{\eta}^2 \right)\left(\mathbb{E}\|A^{t-1}\|^2 - \mathbb{E}\|A^{t}\|^2\right)\\
    \end{aligned}
\end{equation*}

Take telescope sum on the inequality above and applying the fact that $f^{*} \leq f(\mathbf{z})$ for $\mathbf{z}\in \mathbb{R}^{d}$, we have:
\begin{equation}
    \begin{aligned}
        \frac{1}{T}\sum_{t=0}^{T-1}\mathbb{E}\|\nabla f(\mathbf{z}^{t})\|^2 
        &\leq \frac{2\mathbb{E}[f(\mathbf{z}^{0})-f(\mathbf{z}^{T})]}{T\tilde{\eta}\kappa} + \frac{4\tilde{\eta}^2L^2(1-\beta)}{\mu\kappa}G^2 +  \frac{2\tilde{\eta}}{\kappa}\left(\frac{ L^2 m (1-\beta)}{\mu} + L\right)\frac{\sigma^2}{m}  \\
        &\quad + \frac{2}{T\kappa}\left( \frac{ L^2 \beta}{(1-\beta)\mu} + \frac{ L^2 \beta^2}{ (1-\beta)^3} + \frac{4 L^3 \beta^2}{ (1-\beta)^2 \mu}\tilde{\eta}^2 \right)\left(\mathbb{E}\|A^{-1}\|^2 - \mathbb{E}\|A^{T-1}\|^2\right)\\
        &\leq \frac{2\mathbb{E}[f(\mathbf{z}^{0})-f^*]}{T\tilde{\eta}\kappa} + \frac{4\tilde{\eta}^2L^2(1-\beta)}{\mu\kappa}G^2 +  \frac{2\tilde{\eta}}{\kappa}\left(\frac{ L^2 m (1-\beta)}{\mu} + L\right)\frac{\sigma^2}{m}  \\
        &\quad + \frac{2}{T\kappa}\left( \frac{ L^2 \beta}{(1-\beta)\mu} + \frac{ L^2 \beta^2}{ (1-\beta)^3} + \frac{4 L^3 \beta^2}{ (1-\beta)^2 \mu}\tilde{\eta}^2 \right)\mathbb{E}\|\frac{1}{m}\sum_{i=1}^m\mathbf{x}_i^0\|^2\\
    \end{aligned}
\end{equation}

\subsection{Proofs for the generalization Error}\label{sec:Proofs for the generalization Error}
In this section, We first present the fundamental theorem using uniform stability to prove the algorithm's generalization bound, and then we establish an important lemma, which constructs a virtual update sequence $\{\mathbf{z}_{i,k}^t\}$. This sequence takes the form of vanilla SGD, but at the mean level, it shares the same expression as the sequence $\{\mathbf{x}_{i,k}^t\}$ generated by \method{}. Similar to how we adopt $\{\mathbf{z}^t\}$ in proving convergence, the updated format of vanilla SGD is more conducive to our analysis of the algorithm's generalization bound.

\begin{definition}
    (Uniform Stability \cite{Hardt2016train}) For these two models $\mathbf{z}^T$ and $\widetilde{\mathbf{z}}^T$ generated as introduced above, a general method satisfies $\epsilon$-uniformly stability if:
    \begin{equation}
        \operatorname*{sup}_{z_{j}\sim\{\mathcal{D}_{i}\}}\mathbb{E}[f(\mathbf{z}^T;z_{j})-f(\widetilde{\mathbf{z}}^T;z_{j})]\leq\epsilon.
    \end{equation}
    Moreover, if a method satisfies $\epsilon$-uniformly stability, its generalization error is bounded \cite{Zhang2021Stability, Hardt2016train} $$\mathcal{E}_G \leq \operatorname*{sup}_{z_{j}\sim\{\mathcal{D}_{i}\}}\mathbb{E}[f(\mathbf{z}^T;z_{j})-f(\widetilde{\mathbf{z}}^T;z_{j})]\leq\epsilon.$$
\end{definition}

\begin{lemma}\label{le:equivalent}
    (Virtual sequence) We define a virtual sequence $\{\mathbf{z}_{i,k}^t\}$, where its local updates satisfy $\mathbf{z}_{i,k+1}^t = \mathbf{z}_{i,k}^t - \frac{\gamma}{\lambda(1-\beta)}\frac{\gamma_k}{\gamma}\mathbf{g}_{i,k}^t$. The initial value is the normal mixing aggregation, $\mathbf{z}_{i,0}^t = \mathbf{z}_{i}^t = \sum_{j=1}^mw_{i,j}\mathbf{z}_{j,K}^{t-1}$, and we define $\mathbf{z}^t = \frac{1}{m}\sum_{i=1}^m\mathbf{z}_{i}^t$. Then, this virtual sequence is equivalent to the sequence $\{\mathbf{x}_{i,k}^t \}$ generated by \method{} in terms of the mean value.

    \begin{proof}
        The virtual sequence we have defined primarily stems from Lemma \ref{le:auxiliary updates}. Next, I will demonstrate that the conclusions drawn from the virtual sequence defined above are consistent with Lemma \ref{le:auxiliary updates}.
        \begin{equation*}
            \begin{aligned}
                \mathbf{z}^{t+1} &= \frac{1}{m}\sum_{i=1}^m\mathbf{z}_i^{t+1} = \frac{1}{m}\sum_{i=1}^m\sum_{j=1}^mw_{i,j}\mathbf{z}_{j,K}^{t}= \frac{1}{m}\sum_{i=1}^m\sum_{j=1}^mw_{i,j}\mathbf{z}_{j,K}^{t}\\
                & = \frac{1}{m}\sum_{i=1}^m\sum_{j=1}^mw_{i,j}(\mathbf{z}_{j,0}^{t} - \frac{\gamma}{\lambda(1-\beta)}\sum_{k=0}^{K-1}\frac{\gamma_k}{\gamma}\mathbf{g}_{j,k}^t)\\
                & = \frac{1}{m}\sum_{i=1}^m\sum_{j=1}^mw_{i,j}\mathbf{z}_{j,0}^{t} - \frac{\gamma}{\lambda(1-\beta)}\frac{1}{m}\sum_{i=1}^m\sum_{j=1}^mw_{i,j}\sum_{k=0}^{K-1}\frac{\gamma_k}{\gamma}\mathbf{g}_{j,k}^t\\
                & = \frac{1}{m}\sum_{i=1}^m\mathbf{z}_{i,0}^{t} - \frac{\gamma}{\lambda(1-\beta)}\frac{1}{m}\sum_{i=1}^m\sum_{k=0}^{K-1}\frac{\gamma_k}{\gamma}\mathbf{g}_{i,k}^t\\
                & = \frac{1}{m}\sum_{i=1}^m\mathbf{z}_{i}^{t} - \frac{\gamma}{\lambda(1-\beta)}\frac{1}{m}\sum_{i=1}^m\sum_{k=0}^{K-1}\frac{\gamma_k}{\gamma}\mathbf{g}_{i,k}^t\\
                & = \mathbf{z}^{t} - \frac{\gamma}{\lambda(1-\beta)}\frac{1}{m}\sum_{i=1}^m\sum_{k=0}^{K-1}\frac{\gamma_k}{\gamma}\mathbf{g}_{i,k}^t\\
            \end{aligned}
        \end{equation*}
        The above equation is consistent with the conclusion of Lemma \ref{le:auxiliary updates}, proving the equivalence between $\{\mathbf{z}_{i,k}^t\}$ and $\{\mathbf{x}_{i,k}^t\}$.
    \end{proof}
\end{lemma}

In the next part, we prove the generalization error for our proposed method. We assume the objective function $f$ is $L$-smooth and $L_G$-Lipschitz as defined in \cite{Hardt2016train,Zhou2021towards}. We follow the uniform stability to upper bound the generalization error in the DFL. 

We suppose there are $m$ clients participating in the training process as a set $\mathcal{C}=\{i\}_{i=1}^m$. Each client has a local dataset $\mathcal{S}_i=\{z_j\}_{j=1}^S$ with total $S$ data sampled from a specific unknown distribution $\mathcal{D}_i$. Now we define a re-sampled dataset $\widetilde{\mathcal{S}_i}$ which only differs from the dataset $\mathcal{S}_i$ on the $j^*$-th data. We replace the $\mathcal{S}_{i^*}$ with $\widetilde{\mathcal{S}}_{i^*}$ and keep other $m-1$ local dataset, which composes a new set $\widetilde{\mathcal{C}}$. $\mathcal{C}$ only differs from the $\widetilde{\mathcal{C}}$ at $j^*$-th data on the $i^*$-th client. Then, based on these two sets, our method could generate two output models, $\mathbf{z}^t$ and $\widetilde{\mathbf{z}}^t$ respectively, after $t$ training rounds. We first introduce some notations used in the proof of the generalization error.

\begin{table}[h]
  \caption{Some abbreviations of the used terms in the proof of bounded stability error.}
  \label{ta:notation_gener_2}
  \centering  
  \begin{tabular}{ccc}
    \toprule  
    Notation & Formulation  & Description \\  
    \midrule 
    $ \tau $ & $tK + k$ & index of the iterations \\
    $ \tau_0 $ & $t_0K + k_0$ & index of the observed iteration in event $\xi$ \\ 
    $(i^*,j^*)$ & $ - $ & index of the different data sample on $C$ and $\widetilde{C}$\\
    $\Delta_k^t$ & $ \sum_{i=1}^m\mathbb{E}\|\mathbf{z}_{i,k}^{t}-\widetilde{\mathbf{z}}_{i,k}^{t}\|$  & stability difference at $k$-iteration on $t$-round \\
    \bottomrule  
  \end{tabular}  
\end{table}

\subsubsection{Important lemmas}
\begin{lemma}\label{le:main_generation_error_2}
    (Lemma 3.11 in \cite{Hardt2016train}) We follow the definition in \cite{Hardt2016train,Zhou2021towards} to upper bound the uniform stability term after each communication round in DFL paradigm. Different from their vanilla calculations, DFL considers the finite-sum function on heterogeneous clients. Let non-negative objective $f$ is $L$-smooth and $L_G$-Lipschitz. After training $T$ rounds on $\mathcal{C}$ and $\widetilde{\mathcal{C}}$. our method generates two models $\mathbf{z}^{T}$ and $\widetilde{\mathbf{z}}^{T}$ respectively. For each data $z$ and every $\tau_0$, we have:
    \begin{equation*}
        \mathbb{E}\|f({\mathbf{z}}^{T+1};z)-f(\widetilde{\mathbf{z}}^{T+1};z)\|\leq \frac{L_G}{m}\sum_{i=1}^m\mathbb{E}\left[\|\mathbf{z}_{i,K}^T - \widetilde{\mathbf{z}}_{i,K}^T\||\xi\right] + \frac{U\tau_0}{S}
    \end{equation*}
    \begin{proof}
        Let $\xi = 1$ denote the event $\|\mathbf{z}^{\tau_0} - \widetilde{\mathbf{z}}^{\tau_0}\| = 0$ and $U = \sup_{\mathbf{x},z}\{ f ({\mathbf{x}}; z)\}$, we have:
        \begin{equation*}
            \begin{aligned}
                & \mathbf{E}\|f({\mathbf{z}}^{T+1};z)-f(\widetilde{\mathbf{z}}^{T+1};z)\| \\
                & = P(\{\xi\})\mathbb{E}\left[\|f({\mathbf{z}}^{T+1};z)-f(\widetilde{\mathbf{z}}^{T+1};z)\||\xi\right] + P(\{\xi^{c}\})\mathbb{E}\left[\|f({\mathbf{z}}^{T+1};z)-f(\widetilde{\mathbf{z}}^{T+1};z)\| |\xi^{c}\right]\\
                &\leq \mathbb{E}\left[\|f({\mathbf{z}}^{T+1};z)-f(\widetilde{\mathbf{z}}^{T+1};z)\||\xi\right] + P(\{\xi^{c}\})\sup_{\mathbf{x},z}\{ f ({\mathbf{x}}; z)\}\\
                &\leq L_G\mathbb{E}\left[\|{\mathbf{z}}^{T+1}-\widetilde{\mathbf{z}}^{T+1}\||\xi\right] + P(\{\xi^{c}\})U\\
                & =  L_G\mathbb{E}\left[\|\frac{1}{m}\sum_{i=1}^m(\mathbf{z}_{i,K}^T - \widetilde{\mathbf{z}}_{i,K}^T)\||\xi\right] + P(\{\xi^{c}\})U\\
                &\leq \frac{L_G}{m}\sum_{i=1}^m\mathbb{E}\left[\|\mathbf{z}_{i,K}^T - \widetilde{\mathbf{z}}_{i,K}^T\||\xi\right] + P(\{\xi^{c}\})U
            \end{aligned}
        \end{equation*}
        Before the $j^*$-th data on $i^*$-th client is sampled, the iterative states are identical on both $\mathcal{C}$ and $\widetilde{\mathcal{C}}$. Let $\widetilde{j}$ is the index of the first different sampling, if  $\widetilde{j} > \tau_0$, then $\xi = 1$ hold for $\tau_0$. Therefore, we have:
        \begin{equation*}
            P(\{\xi^{c}\}) =P(\{\xi = 0\}) =  P(\widetilde{j} \leq \tau_0) \leq \frac{\tau_0}{S}
        \end{equation*}
        We complete the proof.
    \end{proof}
\end{lemma}

\begin{lemma}\label{le:same_data_2}(Lemma 1.1 in \cite{Zhou2021towards})
    Different from their calculations, we prove similar inequalities on $f$ in the stochastic optimization. Under Assumption \ref{as:smoothness} and \ref{as:bounded_stochastic_gradient}, the local updates satisfy $\mathbf{z}_{i,k+1}^t = \mathbf{z}_{i,k}^t - \frac{\gamma}{\lambda(1-\beta)}\frac{\gamma_k}{\gamma}\mathbf{g}_{i,k}^t$ on $\mathcal{C}$ and $\widetilde{\mathbf{z}}_{i,k+1}^t = \widetilde{\mathbf{z}}_{i,k}^t - \frac{\gamma}{\lambda(1-\beta)}\frac{\gamma_k}{\gamma}\widetilde{\mathbf{g}}_{i,k}^t$ on $\widetilde{\mathcal{C}}$. If at $k$-th iteration on each round, we sample the \textbf{same} data in $\mathcal{C}$ and $\widetilde{\mathcal{C}}$, then we have:
    \begin{equation*}
        \mathbb{E}\|\mathbf{z}_{i,k+1}^{t}-\widetilde{\mathbf{z}}_{i,k+1}^{t}\| \leq (1 + \frac{\gamma L}{\lambda(1-\beta)}\frac{\gamma_k}{\gamma})\mathbb{E}\|\mathbf{z}_{i,k}^t-\widetilde{\mathbf{z}}_{i,k}^t\|
    \end{equation*}
    \begin{proof}
    In each round $t$, by the triangle inequality and omitting the same data $z$, we have:
    \begin{equation*}
        \begin{aligned}
            &\mathbb{E}\|\mathbf{z}_{i,k+1}^{t}-\widetilde{\mathbf{z}}_{i,k+1}^{t}\| \\
            &=\mathbb{E}\|\mathbf{z}_{i,k}^t - \frac{\gamma}{\lambda(1-\beta)}\frac{\gamma_k}{\gamma}\mathbf{g}_{i,k}^t -\widetilde{\mathbf{z}}_{i,k}^t
             + \frac{\gamma}{\lambda(1-\beta)}\frac{\gamma_k}{\gamma}\widetilde{\mathbf{g}}_{i,k}^t\| \\
            &\leq \mathbb{E}\|\mathbf{z}_{i,k}^t-\widetilde{\mathbf{z}}_{i,k}^t\| + \frac{\gamma}{\lambda(1-\beta)}\frac{\gamma_k}{\gamma} \mathbb{E}\|\mathbf{g}_{i,k}^t-\widetilde{\mathbf{g}}_{i,k}^t\| \\
            &\leq \mathbb{E}\|\mathbf{z}_{i,k}^t-\widetilde{\mathbf{z}}_{i,k}^t\| + \frac{\gamma}{\lambda(1-\beta)}\frac{\gamma_k}{\gamma} \mathbb{E}\|\nabla f_i(\mathbf{z}_{i,k}^t) -\nabla f_i(\widetilde{\mathbf{z}}_{i,k}^t)\| \\
            &\leq (1 + \frac{\gamma L}{\lambda(1-\beta)}\frac{\gamma_k}{\gamma})\mathbb{E}\|\mathbf{z}_{i,k}^t-\widetilde{\mathbf{z}}_{i,k}^t\| \\
        \end{aligned}
    \end{equation*}
    \end{proof}
\end{lemma}

\begin{lemma}\label{le:different_same_2}
    Different from their calculations, we prove similar inequalities on $f$ in the stochastic optimization. Under Assumption \ref{as:smoothness} and \ref{as:bounded_stochastic_gradient}, the local updates satisfy $\mathbf{z}_{i,k+1}^t = \mathbf{z}_{i,k}^t - \frac{\gamma}{\lambda(1-\beta)}\frac{\gamma_k}{\gamma}\mathbf{g}_{i,k}^t$ on $\mathcal{C}$ and $\widetilde{\mathbf{z}}_{i,k+1}^t = \widetilde{\mathbf{z}}_{i,k}^t - \frac{\gamma}{\lambda(1-\beta)}\frac{\gamma_k}{\gamma}\widetilde{\mathbf{g}}_{i,k}^t$ on $\widetilde{\mathcal{C}}$. If at $k$-th iteration on each round, we sample the \textbf{different} data in $\mathcal{C}$ and $\widetilde{\mathcal{C}}$, then we have:
    \begin{equation*}
        \mathbb{E}\|\mathbf{x}_{i,k+1}^{t}-\widetilde{\mathbf{x}}_{i,k+1}^{t}\| \leq (1+\eta(L-\lambda))\mathbb{E}\|\mathbf{x}_{i^*,k}^t-\widetilde{\mathbf{x}}_{i^*,k}^t\|+ \lambda\eta\mathbb{E}\|\mathbf{x}_{i^*,0}^t-\widetilde{\mathbf{x}}_{i^*,0}^t\| + 2\eta\sigma
    \end{equation*}
    \begin{proof}
    In each round $t$, we have:
    \begin{equation*}
        \begin{aligned}
            &\mathbb{E}\|\mathbf{z}_{i^*,k+1}^{t}-\widetilde{\mathbf{z}}_{i^*,k+1}^{t}\| \\
            &=\mathbb{E}\|\mathbf{z}_{i^*,k}^t-\frac{\gamma}{\lambda(1-\beta)}\frac{\gamma_k}{\gamma}\mathbf{g}_{i,k}^t -\widetilde{\mathbf{z}}_{i^*,k}^t
            + \frac{\gamma}{\lambda(1-\beta)}\frac{\gamma_k}{\gamma}\widetilde{\mathbf{g}}_{i,k}^t\| \\
            &\leq \mathbb{E}\|\mathbf{z}_{i^*,k}^t-\widetilde{\mathbf{z}}_{i^*,k}^t\| + \frac{\gamma}{\lambda(1-\beta)}\frac{\gamma_k}{\gamma}\mathbb{E}\|f_{i^*}(\mathbf{z}_{i^*,k}^t,z)-f_{i^*}(\widetilde{\mathbf{z}}_{i^*,k}^t,\widetilde{z})\|\\
            &\leq \mathbb{E}\|\mathbf{z}_{i^*,k}^t-\widetilde{\mathbf{z}}_{i^*,k}^t\|+ \frac{\gamma}{\lambda(1-\beta)}\frac{\gamma_k}{\gamma}\mathbb{E}\|f_{i^*}(\mathbf{z}_{i^*,k}^t,z)-f_{i^*}(\mathbf{z}_{i^*,k}^t,\widetilde{z})\| \\
            &\quad + \frac{\gamma}{\lambda(1-\beta)}\frac{\gamma_k}{\gamma}\mathbb{E}\|f_{i^*}(\mathbf{z}_{i^*,k}^t,\widetilde{z})-f_{i^*}(\widetilde{\mathbf{z}}_{i^*,k}^t,\widetilde{z})\| \\
            &\leq (1 + \frac{\gamma L}{\lambda(1-\beta)}\frac{\gamma_k}{\gamma})\mathbb{E}\|\mathbf{z}_{i^*,k}^t-\widetilde{\mathbf{z}}_{i^*,k}^t\|+ \frac{\gamma}{\lambda(1-\beta)}\frac{\gamma_k}{\gamma}\mathbb{E}\|f_{i^*}(\mathbf{z}_{i^*,k}^t,z) - f_{i^*}(\mathbf{z}_{i^*,k}^t)\| \\
            &\quad + \frac{\gamma}{\lambda(1-\beta)}\frac{\gamma_k}{\gamma}\mathbb{E}\|f_{i^*}(\mathbf{z}_{i^*,k}^t)  -f_{i^*}(\mathbf{z}_{i^*,k}^t,\widetilde{z})\|\\
            &\leq (1 + \frac{\gamma L}{\lambda(1-\beta)}\frac{\gamma_k}{\gamma})\mathbb{E}\|\mathbf{z}_{i^*,k}^t-\widetilde{\mathbf{z}}_{i^*,k}^t\| + 2 \frac{\gamma \sigma}{\lambda(1-\beta)}\frac{\gamma_k}{\gamma} \\ 
        \end{aligned}
    \end{equation*}
        The final inequality adopts assumptions of $\mathbb{E}\|\mathbf{g}_{i,k}^t-\nabla f_i(\mathbf{z}_{i,k}^t)\|\le\sqrt{\mathbb{E}\|\mathbf{g}_{i,k}^t-\nabla f_i(\mathbf{z}_{i,k}^t)\|^2}\le\sigma$. This completes the proof.
    \end{proof}
\end{lemma}

\begin{lemma}\label{le:global_stability_2}
    (Upper Bound of Aggregation Gaps) Under the assumption \ref{as:smoothness} and \ref{as:bounded_stochastic_gradient}, the aggregation of decentralized federated learning is $\mathbf{z}_{i,0}^{t+1} = \sum_{j=1}^n w_{i,j}\mathbf{z}_{j,K}^t$. On both setups, we can upper bound the aggregation gaps by:
    \begin{equation}
        \Delta_0^{t+1} \leq \Delta_{K}^t
    \end{equation}
    \begin{proof}
        According to the definition of $\Delta_k^t$ in Table \ref{ta:notation_gener_2}, we obtain:
        \begin{equation}
            \begin{aligned}
                \Delta_0^{t+1} & = \sum_{i=1}^m\mathbb{E}\|\mathbf{z}_{i,0}^{t+1} - \widetilde{\mathbf{z}}_{i,0}^{t+1}\| = \sum_{i=1}^m\mathbb{E}\|\sum_{j=1}^mw_{i,j}(\mathbf{z}_{j,K}^{t} - \widetilde{\mathbf{z}}_{j,K}^{t}) \| \leq \sum_{i=1}^m\sum_{j=1}^mw_{i,j}\mathbb{E}\|\mathbf{z}_{j,K}^{t} - \widetilde{\mathbf{z}}_{j,K}^{t}\| \\
                &= \sum_{i=1}^m\mathbb{E}\|\mathbf{z}_{i,K}^{t} - \widetilde{\mathbf{z}}_{i,K}^{t}\|  =  \Delta_K^t
            \end{aligned}
        \end{equation}
    \end{proof}
\end{lemma}

\begin{lemma}\label{le:local_stability_2}
    According to the Lemma \ref{le:same_data_2} and \ref{le:different_same_2}, we can bound the recursion in the local training:
    \begin{equation*}
        \Delta_{k+1}^t + \frac{2\sigma}{SL} \leq \left(1+\frac{\gamma L}{\lambda(1-\beta)}\frac{\gamma_k}{\gamma}\right)\left(\Delta_k^t + \frac{2\sigma}{SL}\right)
    \end{equation*}

    \begin{proof}
        In each iteration, the specific $j^*$-th data sample in the $S_i^*$ and $\widetilde{S}_i^*$ is uniformly selected with the probability of $1/S$. In other datasets $S_i$, all the data samples are the same. Thus we have:
        \begin{equation}
            \begin{aligned}
                \Delta_{k+1}^t &= \sum_{i\neq i^\star}\mathbb{E}\|\mathbf{z}_{i,k+1}^t-\widetilde{\mathbf{z}}_{i,k+1}^t\|+\mathbb{E}\|\mathbf{z}_{i^\star,k+1}^t-\widetilde{\mathbf{z}}_{i^\star,k+1}^t\|\\
                &\leq (1 +\frac{\gamma L}{\lambda(1-\beta)}\frac{\gamma_k}{\gamma})\sum_{i\neq i^\star}\mathbb{E}\|\mathbf{z}_{i,k}^t-\widetilde{\mathbf{z}}_{i,k}^t\| + \left(1-\frac1S\right)(1 +\frac{\gamma L}{\lambda(1-\beta)}\frac{\gamma_k}{\gamma})\mathbb{E}\|\mathbf{z}_{i,k}^t-\widetilde{\mathbf{z}}_{i,k}^t\| \\
                &\quad + \frac1S (1+\frac{\gamma L}{\lambda(1-\beta)}\frac{\gamma_k}{\gamma})\mathbb{E}\|\mathbf{z}_{i,k}^t-\widetilde{\mathbf{z}}_{i,k}^t\| + \frac{2 \sigma\gamma L}{\lambda(1-\beta)}\frac{\gamma_k}{S\gamma}\\
                & = (1+\frac{\gamma L}{\lambda(1-\beta)}\frac{\gamma_k}{\gamma})\Delta_k^t  + \frac{2 \sigma\gamma L}{\lambda(1-\beta)}\frac{\gamma_k}{S\gamma}
            \end{aligned}
        \end{equation}
        There we can bound the recursion formulation as
        \begin{equation*}
            \Delta_{k+1}^t + \frac{2\sigma}{SL} \leq \left(1+\frac{\gamma L}{\lambda(1-\beta)}\frac{\gamma_k}{\gamma}\right)\left(\Delta_k^t + \frac{2\sigma}{SL}\right)
        \end{equation*}
    \end{proof}
\end{lemma}

\begin{lemma}\label{le: Bounded stability term}
    (Bounded stability term) According to the Lemma \ref{le:local_stability_2} and \ref{le:global_stability_2}, it is easy to bound the local stability term. Define $\widetilde{\mu} = \mu(1-\beta)$ and let the learning rate $ \eta=\frac{\mu(1-\beta)}\tau=\frac{\mu(1-\beta)}{tK+k} = \frac{\widetilde{\mu}}{tK+k}$ is decayed as the communication round $t$ and iteration $k$ where ${\mu} \leq \frac{1}{L}$ is a specific constant, we have:
    \begin{equation*}
        \Delta_K^T + \frac{2\sigma}{SL}\leq \left(\frac{TK}{\tau_0}\right)^{\mu L}\frac{2\sigma}{SL}
    \end{equation*}
    \begin{proof}
        Firstly, due to the fact that $\gamma_k = \eta\lambda(1-\eta\lambda)^{K-1-k} \leq  \eta\lambda$, it follows that $\frac{\gamma_k}{\lambda(1-\beta)} \leq \frac{\eta}{1-\beta}$. Then, if we set $ \eta=\frac{\mu(1-\beta)}\tau=\frac{\mu(1-\beta)}{tK+k}$, we can obtain $\frac{\gamma_k}{\lambda(1-\beta)} \leq \frac{\mu}{tK+k}$.

        According to the Lemma \ref{le:local_stability_2} and \ref{le:global_stability_2}.
        \begin{equation*}
        \begin{aligned}
        \Delta_{K}^T + \frac{2\sigma}{SL} & \leq\left[\prod_{\tau=(T-1)K+1}^{TK}\left(1+\frac{\mu L}{\tau}\right)\right]\left(\Delta_{0}^{T}+\frac{2\sigma}{SL}\right)\\
        &\leq\left[\prod_{\tau=(T-1)K+1}^{TK}\left(1+\frac{\mu L}{\tau}\right)\right]\left(\Delta_{K}^{T-1}+\frac{2\sigma}{SL}\right)  \\
        &\leq\left[\prod_{\tau=t_0K+k_0+1}^{TK}\left(1+\frac{\mu L}{\tau}\right)\right]\left(\Delta_{k_0}^{t_0}+\frac{2\sigma}{SL}\right) \\
        &\leq\left[\prod_{\tau=t_0K+k_0+1}^{TK}e^{\left(\frac{\mu L}{\tau}\right)}\right]\left(\frac{2\sigma}{SL}\right)=e^{\mu L\left(\sum_{\tau=t_0K+k_0+1}^{TK}\frac{1}{\tau}\right)}\frac{2\sigma}{SL} \\
        &\leq e^{\mu L\ln\left(\frac{TK}{t_0K+k_0}\right)}\frac{2\sigma}{SL}\leq\left(\frac{TK}{\tau_0}\right)^{\mu L}\frac{2\sigma}{SL}.
        \end{aligned}
        \end{equation*}
    \end{proof}
\end{lemma}

\begin{lemma}\label{le:jifen}
    For $0 < \lambda < 1$ and $0 < \alpha < 1$, we have the following inequality:
    \begin{equation*}
        \sum_{s=0}^{t-1}\frac{\lambda^{t-s-1}}{\left(s+1\right)^\alpha}\leq\frac{\kappa_\psi}{t^\alpha}
    \end{equation*}
    where $\kappa_{\psi}=\left({\frac{\alpha}{e}}\right)^{\alpha}{\frac{1}{\lambda\left(\ln{\frac{1}{\lambda}}\right)^{\alpha}}}+{\frac{2^{\alpha}}{(1-\alpha)e\lambda\ln{\frac{1}{\lambda}}}}+{\frac{2^{\alpha}}{\lambda\ln{\frac{1}{\lambda}}}}$.
    \begin{proof}
        According to the accumulation, we have:
        \begin{equation*}
        \begin{aligned}
        \sum_{s=0}^{t-1}\frac{\lambda^{t-s-1}}{\left(s+1\right)^{\alpha}}& =\lambda^{t-1}+\sum_{s=1}^{t-1}\frac{\lambda^{t-s-1}}{\left(s+1\right)^{\alpha}}\leq\lambda^{t-1}+\int_{s=1}^{s=t}\frac{\lambda^{t-s-1}}{s^{\alpha}}ds  \\
        &=\lambda^{t-1}+\int_{s=1}^{s=\frac{t}{2}}\frac{\lambda^{t-s-1}}{s^{\alpha}}ds+\int_{s=\frac{t}{2}}^{s=t}\frac{\lambda^{t-s-1}}{s^{\alpha}}ds \\
        &\leq\lambda^{t-1}+\lambda^{\frac{t}{2}-1}\int_{s=1}^{s=\frac{t}{2}}\frac{1}{s^{\alpha}}ds+\left(\frac{2}{t}\right)^{\alpha}\int_{s=\frac{t}{2}}^{s=t}\lambda^{t-s-1}ds \\
        &\leq \lambda^{t-1}+\lambda^{\frac t2-1}\frac1{1-\alpha}\left(\frac t2\right)^{1-\alpha}+\left(\frac2t\right)^\alpha\frac{\lambda^{-1}}{\ln\frac1\lambda}
        \end{aligned}
        \end{equation*}
        Thus we have $ LHS \leq\frac{1}{t^{\alpha}}\left(\lambda^{t-1}t^{\alpha}+\lambda^{\frac{t}{2}-1}\frac{t}{(1-\alpha)2^{1-\alpha}}+\frac{2^{\alpha}}{\lambda\ln\frac{1}{\lambda}}\right)$, The first term can be bounded as $\lambda^{t-1}t^{\alpha}\leq\left(\frac{\alpha}{e}\right)^{\alpha}\frac{1}{\lambda\left(\ln\frac{1}{\lambda}\right)^{\alpha}}$ and the second term can be bounded as $\lambda^{t-1}t^{\alpha}\leq\left(\frac{\alpha}{e}\right)^{\alpha}\frac{1}{\lambda\left(\ln\frac{1}{\lambda}\right)^{\alpha}}$, which indicates the selection of the constant $\kappa_{\psi}=\left(\frac{\alpha}{e}\right)^{\alpha}\frac{1}{\lambda\left(\ln\frac{1}{\lambda}\right)^{\alpha}} + \frac{2^\alpha}{(1-\alpha)e\lambda\ln\frac1\lambda}+\frac{2^\alpha}{\lambda\ln\frac1\lambda}$. Furthermore, if $\kappa_{\psi}\leq\frac{1}{\lambda\left(\ln\frac{1}{\lambda}\right)^{\alpha}}+\frac{2\sqrt{2}}{e\lambda\ln\frac{1}{\lambda}}+\frac{\sqrt{2}}{\lambda\ln\frac{1}{\lambda}}\leq\max\left\{\frac{1}{\lambda},\frac{1}{\lambda\sqrt{\ln\frac{1}{\lambda}}}\right\} + \frac{(2+e)\sqrt{2}}{e\lambda\ln\frac{1}{\lambda}}=\mathcal{O}\left(\max\left\{\frac{1}{\lambda},\frac{1}{\lambda\sqrt{\ln\frac{1}{\lambda}}}\right\}+\frac{1}{\lambda\ln\frac{1}{\lambda}}\right) $ with respect to the constant $\lambda$.
    \end{proof}
\end{lemma}

\subsubsection{Aggregation bound with spectrum gaps}

Let $\mathbf{Z}_k^t=\left[\mathbf{z}_{1,k}^t,\mathbf{z}_{2,k}^t,\cdots,\mathbf{z}_{m,k}^t\right]^\top $ is the parameter matrix of all clients. In the stability analysis, we focus more on the parameter difference instead. Therefore, we denote the matrix of the parameter differences $\Phi_k^t=\mathbf{Z}_k^t-\widetilde{\mathbf{Z}}_k^t= \begin{bmatrix}\mathbf{z}_{1,k}^t-\widetilde{\mathbf{z}}_{1,k}^t,\mathbf{z}_{2,k}^t-\widetilde{\mathbf{z}}_{2,k}^t,\cdots,\mathbf{z}_{m,k}^t-\widetilde{\mathbf{z}}_{m,k}^t\end{bmatrix}^\top $ as the difference between the models trained on $C$ and $\widetilde{C}$ on the $k$-th iteration of $t$-th communication round. Meanwhile, consider the update rules, we have:
\begin{equation*}
    \Phi_{k+1}^t = \Phi_k^t - \frac{\gamma}{\lambda(1-\beta)}\frac{\gamma_k}{\gamma}\Gamma_k^t
\end{equation*}
Where $\Gamma_k^t=\left[\mathbf{g}_{1,k}^t-\widetilde{\mathbf{g}}_{1,k}^t,\mathbf{g}_{2,k}^t-\widetilde{\mathbf{g}}_{2,k}^t,\cdots,\mathbf{g}_{m,k}^t-\widetilde{\mathbf{g}}_{m,k}^t\right]^\top $

Based on the aggregation step in Algorithm \ref{alg:DFedCata}, which demonstrates that the initial state of each round is $\mathbf{Z}_0^t = \mathbf{W}\mathbf{Z}_K^{t-1}$, we obtain the difference $\Phi_0^t = \mathbf{W}\Phi_K^{t-1}$, Therefore, we have:
\begin{equation}\label{eq:phi_k}
    \begin{aligned}
        \Phi_{K}^t & =\Phi_0^t - \frac{\gamma}{\lambda(1-\beta)}\sum_{k=0}^{K-1}\frac{\gamma_k}{\gamma}\Gamma_{k}^t = \mathbf{W}\Phi_K^{t-1} - \frac{\gamma}{\lambda(1-\beta)}\sum_{k=0}^{K-1}\frac{\gamma_k}{\gamma}\Gamma_{k}^t
    \end{aligned}
\end{equation}
where $\sum_{k=0}^{K-1}\gamma_{k}=\sum_{k=0}^{K-1}\eta\lambda\bigl(1-{\eta}{\lambda}\bigr)^{K-1-k}=\gamma=1-(1-{\eta}{\lambda})^{K}$.

Then we prove the recurrence between adjacent rounds. Let $\mathbf{P}=\frac1m\mathbf{1}\mathbf{1}^\top\in\mathbb{R}^{m\times m}$ and $\mathbf{I} \in\mathbb{R}^{m\times m} $ is the identity matrix, due to the double stochastic property of the adjacent matrix $\mathbf{W}$, we have:
$$\mathbf{PW=WP=P}$$
Thus we have:
\begin{equation*}
    \begin{aligned}
        &(\mathbf{I-P})\Phi_{K}^t  = (\mathbf{I-P})\mathbf{W}\Phi_K^{t-1} - (\mathbf{I-P})\frac{\gamma}{\lambda(1-\beta)}\sum_{k=0}^{K-1}\frac{\gamma_k}{\gamma}\Gamma_{k}^t \\
        &= (\mathbf{W}\Phi_K^{t-1} - \frac{\gamma}{\lambda(1-\beta)}\sum_{k=0}^{K-1}\frac{\gamma_k}{\gamma}\Gamma_{k}^t) -\mathbf{P}\mathbf{W}\Phi_K^{t-1} + \mathbf{P}\mathbf{W}\Phi_K^{t-1} -\mathbf{P}(\mathbf{W}\Phi_K^{t-1} - \frac{\gamma}{\lambda(1-\beta)}\sum_{k=0}^{K-1}\frac{\gamma_k}{\gamma}\Gamma_{k}^t)
    \end{aligned}
\end{equation*}

By taking the expectation of the norm on both sides, we have:
\begin{equation*}
    \begin{aligned}
        &\mathbb{E}\|(\mathbf{I-P})\Phi_{K}^t\|\\
        &\leq \mathbb{E}\|\mathbf{W}\Phi_K^{t-1} - \frac{\gamma}{\lambda(1-\beta)}\sum_{k=0}^{K-1}\frac{\gamma_k}{\gamma}\Gamma_{k}^t -\mathbf{P}\mathbf{W}\Phi_K^{t-1}\| + \mathbb{E}\| \frac{\gamma}{\lambda(1-\beta)}\sum_{k=0}^{K-1}\frac{\gamma_k}{\gamma}\Gamma_{k}^t\| \\
        &\leq \mathbb{E}\|\mathbf{W}\Phi_K^{t-1} -\mathbf{P}\mathbf{W}\Phi_K^{t-1}\| + 2\mathbb{E}\| \frac{\gamma}{\lambda(1-\beta)}\sum_{k=0}^{K-1}\frac{\gamma_k}{\gamma}\Gamma_{k}^t\| \\
        & = \mathbb{E}\|(\mathbf{W}-\mathbf{P})(\mathbf{I}-\mathbf{P})\mathbf{W}\Phi_K^{t-1}\| + 2\mathbb{E}\| \frac{\gamma}{\lambda(1-\beta)}\sum_{k=0}^{K-1}\frac{\gamma_k}{\gamma}\Gamma_{k}^t\| \\
        & \leq \psi\mathbb{E}\|(\mathbf{I}-\mathbf{P})\Phi_K^{t-1}\| + \frac{2\gamma}{\lambda(1-\beta)}\mathbb{E}\| \sum_{k=0}^{K-1}\frac{\gamma_k}{\gamma}\Gamma_{k}^t\| \\
    \end{aligned}
\end{equation*}
The equality adopts $\left(\mathbf{W}-\mathbf{P}\right)\left(\mathbf{I}-\mathbf{P}\right)=\mathbf{W}-\mathbf{P}-\mathbf{W}\mathbf{P}+\mathbf{P}\mathbf{P}=\mathbf{W}-\mathbf{P}\mathbf{W} = \left(\mathbf{I}-\mathbf{P}\right)\mathbf{W}$. We know the fact that $\Phi_k^t=0$ where $(t,k)\in (t_0,k_0)$. Thus unwinding the above inequality we have:

\begin{equation*}
    \begin{aligned}
        \mathbb{E}\|(\mathbf{I-P})\Phi_{K}^t\| & \leq 
        \psi^{t-t_0+1}\mathbb{E}\|\left(\mathbf{I}-\mathbf{P}\right)\Phi_K^{t_0-1}\| + \frac{2\gamma}{\lambda(1-\beta)}\sum_{s=t_0}^t\psi^{t-s}\mathbb{E}\|\sum_{k=0}^{K-1}\frac{\gamma_k}{\gamma}\Gamma_k^s\|\\
        &\leq \frac{2\gamma}{\lambda(1-\beta)}\sum_{s=t_0}^t\psi^{t-s}\mathbb{E}\|\sum_{k=0}^{K-1}\frac{\gamma_k}{\gamma}\Gamma_k^s\|
    \end{aligned}
\end{equation*}

On the other hand, we can derive the following equation from equation (\ref{eq:phi_k}):
\begin{equation*}
    \begin{aligned}
    (\mathbf{W-P})\Phi_{K}^t & = (\mathbf{W-P})\mathbf{W}\Phi_K^{t-1} - (\mathbf{W-P})\frac{\gamma}{\lambda(1-\beta)}\sum_{k=0}^{K-1}\frac{\gamma_k}{\gamma}\Gamma_{k}^t \\ 
    & = (\mathbf{W-P})\mathbf{W-P}\Phi_K^{t-1} - (\mathbf{W-P})\frac{\gamma}{\lambda(1-\beta)}\sum_{k=0}^{K-1}\frac{\gamma_k}{\gamma}\Gamma_{k}^t
    \end{aligned}
\end{equation*}

The equality adopts $\left(\mathbf{W}-\mathbf{P}\right)\left(\mathbf{W}-\mathbf{P}\right)=\left(\mathbf{W}-\mathbf{P}\right)\mathbf{W}-\mathbf{W}\mathbf{P}+\mathbf{P}\mathbf{P}=\left(\mathbf{W}-\mathbf{P}\right)\mathbf{W}$. Therefore we have the following recursive formula:
\begin{equation*}
    \begin{aligned}
    \mathbb{E}\|(\mathbf{W-P})\Phi_{K}^t\| & \leq \mathbb{E}\|(\mathbf{W-P})\mathbf{W}\Phi_K^{t-1}\| + \mathbb{E}\|(\mathbf{W-P})\frac{\gamma}{\lambda(1-\beta)}\sum_{k=0}^{K-1}\frac{\gamma_k}{\gamma}\Gamma_{k}^t\| \\ 
    & \leq \mathbb{E}\|(\mathbf{W-P})(\mathbf{W-P})\Phi_K^{t-1}\| + \mathbb{E}\|(\mathbf{W-P})\|\|\frac{\gamma}{\lambda(1-\beta)}\sum_{k=0}^{K-1}\frac{\gamma_k}{\gamma}\Gamma_{k}^t\| \\ 
    & \leq \psi\mathbb{E}\|(\mathbf{W-P})\Phi_K^{t-1}\| + \psi\mathbb{E}\|\frac{\gamma}{\lambda(1-\beta)}\sum_{k=0}^{K-1}\frac{\gamma_k}{\gamma}\Gamma_{k}^t\| \\ 
    \end{aligned}
\end{equation*}

Similarly, we can unfold this recursive formulation from $t$ to $t_0$ as:

\begin{equation*}
    \begin{aligned}
    \mathbb{E}\|(\mathbf{W-P})\Phi_{K}^t\|
    & \leq \psi^{t-t_0+1}\mathbb{E}\|(\mathbf{W-P})\Phi_K^{t_0-1}\| + \frac{\gamma}{\lambda(1-\beta)}\sum_{s=t_0}^t\psi^{t-s+1}\mathbb{E}\|\sum_{k=0}^{K-1}\frac{\gamma_k}{\gamma}\Gamma_{k}^s\| \\ 
    & \leq  \frac{\gamma}{\lambda(1-\beta)}\sum_{s=t_0}^t\psi^{t-s+1}\mathbb{E}\|\sum_{k=0}^{K-1}\frac{\gamma_k}{\gamma}\Gamma_{k}^s\| \\ 
    \end{aligned}
\end{equation*}

Both terms require an upper limit on the accumulation of gradient differences. In previous works, these terms were often assumed to have a constant upper bound under the assumption of bounded gradients. However, as we have introduced in the main text, this assumption is strong and may not always hold. Therefore, we provide a new upper bound instead. When $(t,k) < (t_0, k_0)$, the sampled data is consistent across different datasets, resulting in $\Gamma_k^t = 0$. When $t=t_0$, the updates are only different for $k \geq k_0$. When $t>t_0$, all the local gradient differences during local $K$ iterations are non-zero. Therefore, we can initially investigate the upper bound of the stages with full $K$ iterations when $t>t_0$. Let the data sample $z$ be the same random data sample and $z/\widetilde{z}$ be a different sample pair for abbreviation, when $t \geq t_0$, we have:
\begin{align*}
        &\mathbb{E}\|\frac{\gamma}{\lambda(1-\beta)}\frac{\gamma_k}{\gamma}\Gamma_k^t\| = \mathbb{E}\|\frac{\gamma}{\lambda(1-\beta)}\frac{\gamma_k}{\gamma}\left[\mathbf{g}_{1,k}^t-\widetilde{\mathbf{g}}_{1,k}^t,\mathbf{g}_{2,k}^t-\widetilde{\mathbf{g}}_{2,k}^t,\cdots,\mathbf{g}_{m,k}^t -\widetilde{\mathbf{g}}_{m,k}^t\right]^\top\| \\
        &\leq\frac{\gamma}{\lambda(1-\beta)}\frac{\gamma_k}{\gamma}\sum_{i\in[m]}\mathbb{E}\|\mathbf{g}_{i,k}^t-\widetilde{\mathbf{g}}_{i,k}^t\|\\
        &\leq \frac{\gamma}{\lambda(1-\beta)}\frac{\gamma_k}{\gamma}\sum_{i\neq i^{\star}}\mathbb{E}\|\nabla f_{i}(\mathbf{z}_{i,k}^{t},z)-\nabla f_{i}(\widetilde{\mathbf{z}}_{i,k}^{t},z)\| \\
        &\quad +\frac{(S-1)}{S}\frac{\gamma}{\lambda(1-\beta)}\frac{\gamma_k}{\gamma}\mathbb{E}\|\nabla f_{i^{\star}}(\mathbf{z}_{i^{\star},k}^{t},z)-\nabla f_{i^{\star}}(\widetilde{\mathbf{z}}_{i^{\star},k}^{t},z)\|\\
        &\quad +\frac{1}{S}\frac{\gamma}{\lambda(1-\beta)}\frac{\gamma_k}{\gamma}\mathbb{E}\|\nabla f_{i^{\star}}(\mathbf{z}_{i^{\star},k}^{t},z)-\nabla f_{i^{\star}}(\widetilde{\mathbf{z}}_{i^{\star},k}^{t},z)+\nabla f_{i^{\star}}(\widetilde{\mathbf{z}}_{i^{\star},k}^{t},z)-\nabla f_{i^{\star}}(\widetilde{\mathbf{z}}_{i^{\star},k}^{t},\widetilde{z})\| \\
        &\leq \frac{\gamma L}{\lambda(1-\beta)}\frac{\gamma_k}{\gamma}\sum_{i\neq i^{\star}}\mathbb{E}\|\mathbf{z}_{i,k}^{t}-\widetilde{\mathbf{z}}_{i,k}^{t}\|+\frac{(S-1)}{S}\frac{\gamma L}{\lambda(1-\beta)}\frac{\gamma_k}{\gamma}\mathbb{E}\| \mathbf{z}_{i^{\star},k}^{t}-\widetilde{\mathbf{z}}_{i^{\star},k}\| \\
        &\quad +  \frac{1}{S}\frac{\gamma L}{\lambda(1-\beta)}\frac{\gamma_k}{\gamma}\mathbb{E}\| \mathbf{z}_{i^{\star},k}^{t}-\widetilde{\mathbf{z}}_{i^{\star},k}\| \\
        &\quad +\frac{1}{S}\frac{\gamma}{\lambda(1-\beta)}\frac{\gamma_k}{\gamma}\mathbb{E}\|(\nabla f_{i^{\star}}(\mathbf{z}_{i^{\star},k}^{t},z)-\nabla f_{i^{\star}}(\widetilde{\mathbf{z}}_{i^{\star},k}^{t}))-(\nabla f_{i^{\star}}(\widetilde{\mathbf{z}}_{i^{\star},k}^{t},\widetilde{z}) - \nabla f_{i^{\star}}(\widetilde{\mathbf{z}}_{i^{\star},k}^{t},z))\| \\
        &\leq \frac{\gamma L}{\lambda(1-\beta)}\frac{\gamma_k}{\gamma}\sum_{i\in [m]}\mathbb{E}\|\mathbf{z}_{i,k}^{t}-\widetilde{\mathbf{z}}_{i,k}^{t}\| + \frac{2\sigma}{S}\frac{\gamma}{\lambda(1-\beta)}\frac{\gamma_k}{\gamma} = \frac{\gamma L}{\lambda(1-\beta)}\frac{\gamma_k}{\gamma}(\Delta_k^t + \frac{2\sigma}{SL})
\end{align*}

According to the Lemma \ref{le:local_stability_2}, \ref{le:global_stability_2} and \ref{le: Bounded stability term}, we bound the gradient difference as:
\begin{equation*}
    \mathbb{E}\|\frac{\gamma}{\lambda(1-\beta)}\frac{\gamma_k}{\gamma}\Gamma_k^t\| \leq \frac{\gamma L}{\lambda(1-\beta)}\frac{\gamma_k}{\gamma}(\Delta_k^t + \frac{2\sigma}{SL}) \leq \left(\frac{\tau}{\tau_{0}}\right)^{\mu L}\frac{2\mu\sigma }{\tau S}
\end{equation*}

Unwinding the summation on $k$ and adopting Lemma \ref{le:jifen}, we have:
\begin{equation*}
    \begin{aligned}
        &\sum_{s=t_0}^t\psi^{t-s}\mathbb{E}\|\sum_{k=0}^{K-1}\frac{\gamma}{\lambda(1-\beta)}\frac{\gamma_k}{\gamma}\Gamma_{k}^s\|  \leq  \sum_{s=t_0}^t\psi^{t-s}\sum_{k=0}^{K-1}\mathbb{E}\|\frac{\gamma}{\lambda(1-\beta)}\frac{\gamma_k}{\gamma}\Gamma_{k}^s\| \\
        &\leq \frac{2\mu\sigma}{S\tau_0^{\mu L}}\sum_{s=t_0}^t\psi^{t-s}\sum_{k=0}^{K-1}\frac{\tau^{\mu L}}{\tau} \leq \frac{2\mu\sigma}{S\tau_0^{\mu L}}\sum_{s=t_0}^t\psi^{t-s}\sum_{k=0}^{K-1}\frac{(sK)^{\mu L}}{sK} \\
        &=\frac{2\mu\sigma}{S}\left(\frac{K}{\tau_{0}}\right)^{\mu L}\sum_{s=t_{0}}^{t}\frac{\psi^{t-s}}{s^{1-\mu L}} \\
        & \leq\frac{2\mu\sigma}{S}\left(\frac{K}{\tau_{0}}\right)^{\mu L}\sum_{{s=t_{0}-1}}^{t-1}\frac{\psi^{t-s-1}}{\left(s+1\right)^{1-\mu L}}\leq\frac{2\mu\sigma\kappa_{\psi}}{S}\left(\frac{K}{\tau_{0}}\right)^{\mu L}\frac{1}{t^{1-\mu L}}.
    \end{aligned}
\end{equation*}

Therefore, we get an upper bound on the aggregation gap which is related to the spectrum gap:

\begin{equation*}
    \begin{aligned}
        \mathbb{E}\|\left(\mathbf{I}-\mathbf{P}\right)\Phi_K^t\| \leq \frac{2\gamma}{\lambda(1-\beta)}\sum_{s=t_0}^t\psi^{t-s}\mathbb{E}\|\sum_{k=0}^{K-1}\frac{\gamma_k}{\gamma}\Gamma_k^s\| \leq \frac{4\mu\sigma\kappa_{\psi}}{S}\left(\frac{K}{\tau_{0}}\right)^{\mu L}\frac{1}{t^{1-\mu L}}
    \end{aligned}
\end{equation*}

\begin{equation*}
    \begin{aligned}
        \mathbb{E}\|\left(\mathbf{W}-\mathbf{P}\right)\Phi_K^t\| \leq \frac{\gamma}{\lambda(1-\beta)}\sum_{s=t_0}^t\psi^{t-s+1}\mathbb{E}\|\sum_{k=0}^{K-1}\frac{\gamma_k}{\gamma}\Gamma_k^s\| \leq \frac{2\mu\sigma\psi\kappa_{\psi}}{S}\left(\frac{K}{\tau_{0}}\right)^{\mu L}\frac{1}{t^{1-\mu L}}
    \end{aligned}
\end{equation*}

The first inequality provides an upper bound on the difference between the aggregated state and the vanilla state, while the second inequality provides an upper bound on the difference between the aggregated state and the average state.

\subsubsection{Stability Bound}

Now, rethinking the update rules in one round and we have:

\begin{align*}
        &\sum_{i\in[m]}\mathbb{E}\|\mathbf{z}_{i,K}^{t+1}-\widetilde{\mathbf{z}}_{i,K}^{t+1}\| \\
        & = \sum_{i\in[m]}\mathbb{E}\|\mathbf{z}_{i,0}^{t+1}-\widetilde{\mathbf{z}}_{i,0}^{t+1} - \frac{\gamma}{\lambda(1-\beta)}\sum_{k=0}^{K-1}\frac{\gamma_k}{\gamma}\left(\mathbf{g}_{i,k}^t-\widetilde{\mathbf{g}}_{i,k}^t\right)\| \\
        & = \sum_{i\in[m]}\mathbb{E}\|\mathbf{z}_{i,0}^{t+1}-\widetilde{\mathbf{z}}_{i,0}^{t+1} -(\mathbf{z}_{i,K}^{t}-\widetilde{\mathbf{z}}_{i,K}^{t}) + (\mathbf{z}_{i,K}^{t}-\widetilde{\mathbf{z}}_{i,K}^{t})- \frac{\gamma}{\lambda(1-\beta)}\sum_{k=0}^{K-1}\frac{\gamma_k}{\gamma}\left(\mathbf{g}_{i,k}^t-\widetilde{\mathbf{g}}_{i,k}^t\right)\| \\
        &\leq \sum_{i\in[m]}\mathbb{E}\left[\|\mathbf{z}_{i,0}^{t+1}-\widetilde{\mathbf{z}}_{i,0}^{t+1} -(\mathbf{z}_{i,K}^{t}-\widetilde{\mathbf{z}}_{i,K}^{t})\| + \mathbb{E}\|\mathbf{z}_{i,K}^{t}-\widetilde{\mathbf{z}}_{i,K}^{t}\| + \mathbb{E}\|\frac{\gamma}{\lambda(1-\beta)}\sum_{k=0}^{K-1}\frac{\gamma_k}{\gamma}\left(\mathbf{g}_{i,k}^t-\widetilde{\mathbf{g}}_{i,k}^t\right)\| \right]\\
        &\leq m\mathbb{E}\left[\frac{1}{m}\sum_{i\in[m]}\|\mathbf{z}_{i,0}^{t+1}-\widetilde{\mathbf{z}}_{i,0}^{t+1} -(\mathbf{z}_{i,K}^{t}-\widetilde{\mathbf{z}}_{i,K}^{t})\|\right] + \sum_{i\in[m]}\mathbb{E}\|\mathbf{z}_{i,K}^{t}-\widetilde{\mathbf{z}}_{i,K}^{t}\|\\
        &\quad + \sum_{i\in[m]}\mathbb{E}\|\frac{\gamma}{\lambda(1-\beta)}\sum_{k=0}^{K-1}\frac{\gamma_k}{\gamma}\left(\mathbf{g}_{i,k}^t-\widetilde{\mathbf{g}}_{i,k}^t\right)\| \\
        &\leq m\mathbb{E}\sqrt{\frac{1}{m}\sum_{i\in[m]}\|\mathbf{z}_{i,0}^{t+1}-\widetilde{\mathbf{z}}_{i,0}^{t+1} -(\mathbf{z}_{i,K}^{t}-\widetilde{\mathbf{z}}_{i,K}^{t})\|^2} + \sum_{i\in[m]}\mathbb{E}\|\mathbf{z}_{i,K}^{t}-\widetilde{\mathbf{z}}_{i,K}^{t}\|\\
        &\quad + \sum_{i\in[m]}\mathbb{E}\|\frac{\gamma}{\lambda(1-\beta)}\sum_{k=0}^{K-1}\frac{\gamma_k}{\gamma}\left(\mathbf{g}_{i,k}^t-\widetilde{\mathbf{g}}_{i,k}^t\right)\| \\
        &= \sqrt{m}\mathbb{E}\|\Phi_0^{t+1}-\Phi_K^t\| + \sum_{i\in[m]}\mathbb{E}\|\mathbf{z}_{i,K}^{t}-\widetilde{\mathbf{z}}_{i,K}^{t}\| + \sum_{i\in[m]}\mathbb{E}\|\frac{\gamma}{\lambda(1-\beta)}\sum_{k=0}^{K-1}\frac{\gamma_k}{\gamma}\left(\mathbf{g}_{i,k}^t-\widetilde{\mathbf{g}}_{i,k}^t\right)\| \\
        &= \sqrt{m}\mathbb{E}\|\mathbf{W}\Phi_K^{t}-\Phi_K^t\| + \sum_{i\in[m]}\mathbb{E}\|\mathbf{z}_{i,K}^{t}-\widetilde{\mathbf{z}}_{i,K}^{t}\| + \sum_{i\in[m]}\mathbb{E}\|\frac{\gamma}{\lambda(1-\beta)}\sum_{k=0}^{K-1}\frac{\gamma_k}{\gamma}\left(\mathbf{g}_{i,k}^t-\widetilde{\mathbf{g}}_{i,k}^t\right)\| \\
    &\leq \sqrt{m}\mathbb{E}\|(\mathbf{W}-\mathbf{P})\Phi_K^t\| + \sqrt{m}\mathbb{E}\|(\mathbf{P}-\mathbf{I})\Phi_K^t\| + \sum_{i\in[m]}\mathbb{E}\|\mathbf{z}_{i,K}^{t}-\widetilde{\mathbf{z}}_{i,K}^{t}\| \\
    &\quad + \sum_{i\in[m]}\mathbb{E}\|\frac{\gamma}{\lambda(1-\beta)}\sum_{k=0}^{K-1}\frac{\gamma_k}{\gamma}\left(\mathbf{g}_{i,k}^t-\widetilde{\mathbf{g}}_{i,k}^t\right)\| \\  
\end{align*}

Therefore, we can bound this by two terms in one complete communication round. One is the process of local $K$ SGD iterations, and the other is the aggregation step. For the local training process, we can continue to use Lemma \ref{le:same_data_2}, \ref{le:different_same_2}, and \ref{le:local_stability_2}. Let $\tau = tK + k$ as above, we have:

\begin{align*}
&\Delta_K^t+\frac{2\sigma}{SL} \\
&\begin{aligned}\leq\left[\prod_{k=0}^{K-1}\left(1+\frac{\gamma L}{\lambda(1-\beta)}\frac{\gamma_k}{\gamma}\right)\right]\left(\Delta_0^t+\frac{2\sigma}{SL}\right) \leq \left[\prod_{k=0}^{K-1}\left(1+\frac{\mu L}{\tau}\right)\right]\left(\Delta_0^t+\frac{2\sigma}{SL}\right)\end{aligned} \\
&\leq\left[\prod_{k=0}^{K-1}e^{\frac{\mu L}{\tau}}\right]\left(\Delta_{0}^{t}+\frac{2\sigma}{SL}\right)=e^{\mu L\sum_{k=0}^{K-1}\frac{1}{\tau}}\left(\Delta_{0}^{t}+\frac{2\sigma}{SL}\right) \\
&\leq e^{\mu L\ln\left(\frac{t+1}{t}\right)}\left(\Delta_{0}^{t}+\frac{2\sigma}{SL}\right)=\left(\frac{t+1}{t}\right)^{\mu L}\left(\Delta_{0}^{t}+\frac{2\sigma}{SL}\right) \\
&\begin{aligned}&\leq\left(\frac{t+1}{t}\right)^{\mu L}\left[\Delta_K^{t-1}+\sqrt{m}\left(\mathbb{E}\|\left(\mathbf{W}-\mathbf{P}\right)\Phi_K^t\right\|+\mathbb{E}\|\left(\mathbf{P}-\mathbf{I}\right)\Phi_K^t\|\right)+\frac{2\sigma}{SL}\end{aligned} \\
&\begin{aligned}\leq\left(\frac{t+1}{t}\right)^{\mu L}\left(\Delta_K^{t-1}+\frac{2\sigma}{SL}\right)+\sqrt{m}\left(\frac{t+1}{t}\right)^{\mu L}\left(\mathbb{E}\|\left(\mathbf{W}-\mathbf{P}\right)\Phi_K^t\|+\mathbb{E}\|\left(\mathbf{P}-\mathbf{I}\right)\Phi_K^t\|\right)\end{aligned} \\
&\leq\underbrace{\left(\frac{t+1}{t}\right)^{\mu L}\left(\Delta_K^{t-1}+\frac{2\sigma}{SL}\right)}_{\text{local updates}}+\underbrace{\frac{6\sqrt{m}\mu\sigma\kappa_\psi}{S}\left(\frac{K}{\tau_0}\right)^{\mu L}\left(\frac{t+1}{t}\right)^{\mu L}\frac{1}{t^{1-\mu L}},}_{\text{aggregation gaps}}
\end{align*}

Obviously, in the context of decentralized federated learning, the first term still arises from local training updates, while the second term comes from the aggregation gap associated with the spectral gap $\psi$. Expanding from $t_0$ to $T$, we have:

\begin{align*}
\Delta_{K}^{T}+\frac{2\sigma}{SL} &\leq\left(\frac{TK}{\tau_0}\right)^{\mu L}\frac{2\sigma}{SL}+\frac{6\sqrt{m}\mu\sigma\kappa_\psi}{S}\left(\frac{K}{\tau_0}\right)^{\mu L}\sum_{t=t_0+1}^T\left(\frac{t+1}{t}\right)^{\mu L}\frac{1}{t^{1-\mu L}} \\
&\leq\left(\frac{TK}{\tau_0}\right)^{\mu L}\frac{2\sigma}{SL}+\frac{12\sqrt{m}\mu\sigma\kappa_\psi}{S}\left(\frac{K}{\tau_0}\right)^{\mu L}\sum_{t=t_0+1}^T\frac{1}{t^{1-\mu L}} \\
&\leq\left(\frac{TK}{\tau_{0}}\right)^{\mu L}\frac{2\sigma}{SL}+\frac{12\sqrt{m}\mu\sigma\kappa_{\psi}}{S}\left(\frac{K}{\tau_{0}}\right)^{\mu L}\frac{t^{\mu L}}{\mu L}\Bigg|_{t=t_{0}+1}^{t=T} \\
& \leq \left(\frac{TK}{\tau_0}\right)^{\mu L}\frac{2\left(1+6\sqrt{m}\kappa_\psi\right)\sigma}{SL}
\end{align*}

The second inequality adopts the fact that $1 < \frac{t+1}{t} \leq 2 $ when $t > 1$ and the fact of $0< \mu <\frac{1}{L}$.

According to the Lemma \ref{le:main_generation_error_2}, the first term in the stability (conditions is omitted for abbreviation) can be bounded as:
\begin{align*}
    \frac{1}{m}\sum_{i=1}^m\mathbb{E}\|\mathbf{z}_{i,K}^T - \widetilde{\mathbf{z}}_{i,K}^T\| \leq \left(\frac{TK}{\tau_0}\right)^{\mu L}\frac{2\left(1+6\sqrt{m}\kappa_\psi\right)\sigma}{SL}
\end{align*} 

Therefore, we can upper bound the stability in decentralized federated learning as:
\begin{align*}
    \mathbb{E}\|f({\mathbf{z}}^{T+1};z)-f(\widetilde{\mathbf{z}}^{T+1};z)\| &\leq \frac{L_G}{m}\sum_{i=1}^m\mathbb{E}\left[\|\mathbf{z}_{i,K}^T - \widetilde{\mathbf{z}}_{i,K}^T\||\xi\right] + \frac{U\tau_0}{S} \\
    &\leq L_G \left(\frac{TK}{\tau_0}\right)^{\mu L}\frac{2\left(1+6\sqrt{m}\kappa_\psi\right)\sigma}{SL} + \frac{U\tau_0}{S}
\end{align*}

To minimize the error of the stability, we can select a proper event $\xi$ with a proper $\tau_0$. For $\tau \in [1,TK]$, by selecting $\tau_{0}=\left(\frac{2\sigma L_G}{UL}\frac{1+6\sqrt{m}\kappa_{\psi}}{m}\right)^{\frac{1}{1+\mu L}}(TK)^{\frac{\mu L}{1+\mu L}}$, we get the minimal:
\begin{align*}
    \mathbb{E}\|f({\mathbf{z}}^{T+1};z)-f(\widetilde{\mathbf{z}}^{T+1};z)\| &\leq \frac{2U\tau_0}{S} = \frac{2U}{S}\left(\frac{2\sigma L_G}{UL}\frac{1+6\sqrt{m}\kappa_{\psi}}{m}\right)^{\frac{1}{1+\mu L}}(TK)^{\frac{\mu L}{1+\mu L}}
\end{align*}
Substituting the definition of $\mu = \frac{\widetilde{\mu}}{1-\beta}$ from Lemma \ref{le: Bounded stability term} into the above equation yields
\begin{align*}
    \mathbb{E}\|f({\mathbf{z}}^{T+1};z)-f(\widetilde{\mathbf{z}}^{T+1};z)\| &\leq \frac{2U\tau_0}{S} = \frac{2U}{S}\left(\frac{2\sigma L_G}{UL}\frac{1+6\sqrt{m}\kappa_{\psi}}{m}\right)^{\frac{1-\beta}{1 - \beta +\widetilde{\mu} L}}(TK)^{\frac{\widetilde{\mu} L}{1 - \beta +\widetilde{\mu} L}}
\end{align*}

% \newpage

 

% \end{IEEEbiography}



% \vfill

\end{document}